%% file: ClassicThesis.tex
\RequirePackage{fix-cm} 
\documentclass[ twoside,openright,titlepage,numbers=noenddot,headinclude,
                footinclude=true,cleardoublepage=empty,abstractoff, 
                BCOR=5mm,paper=a4,fontsize=11pt,
                ngerman,american,%
                ]{scrreprt}

\input{classicthesis-config}

\begin{document}
\frenchspacing
\raggedbottom
\selectlanguage{american} 
\pagenumbering{roman}
\pagestyle{plain}
\include{FrontBackmatter/Titlepage}
\include{FrontBackmatter/Titleback}
\cleardoublepage\include{FrontBackmatter/Dedication}
\cleardoublepage\include{FrontBackmatter/Abstract}
\cleardoublepage\include{FrontBackmatter/Publications}

\include{Chapters/Prologue}

\pagestyle{scrheadings}
\cleardoublepage\include{FrontBackmatter/Contents}
\cleardoublepage\pagenumbering{arabic}
\cleardoublepage
\include{Chapters/Chapter01}

\cleardoublepage
\include{Chapters/Chapter02}
\cleardoublepage

\include{Chapters/Chapter03}
\cleardoublepage

\include{Chapters/Chapter04}
\cleardoublepage

\include{Chapters/Chapter05}

\cleardoublepage

\include{Chapters/Chapter06}

\cleardoublepage

\appendix
\cleardoublepage
\include{Chapters/Chapter0A1}
\cleardoublepage

\include{Chapters/Chapter0A2}
\cleardoublepage
\manualmark
\markboth{\spacedlowsmallcaps{\bibname}}{\spacedlowsmallcaps{\bibname}} 
\refstepcounter{dummy}
\addtocontents{toc}{\protect\vspace{\beforebibskip}} 
\addcontentsline{toc}{chapter}{\tocEntry{\bibname}}
\label{app:bibliography}
\printbibliography

\end{document}

%% file: classicthesis-config.tex
\PassOptionsToPackage{utf8}{inputenc}
	\usepackage{inputenc}

\PassOptionsToPackage{eulerchapternumbers,listings, 
					 pdfspacing,
					 subfig,beramono,eulermath,parts}{classicthesis}                                        

\newcommand{\myTitle}{Artificial Scientific Discovery\xspace}
\newcommand{\mySubtitle}{\xspace}
\newcommand{\myDegree}{PhD thesis in Computer Science\xspace}
\newcommand{\myName}{Antonio Norelli\xspace}

\newcommand{\mySupervisor}{Prof. Emanuele Rodolà\xspace}
\newcommand{\myFaculty}{Facoltà di ingegneria dell'informazione, informatica e statistica\xspace}
\newcommand{\myDepartment}{Dipartimento di Informatica\xspace}
\newcommand{\myUni}{Università degli Studi di Roma ``La Sapienza''\xspace}

\newcommand{\myTime}{October 2023\xspace}
\newcommand{\myVersion}{\xspace}

\newcounter{dummy} 
\providecommand{\mLyX}{L\kern-.1667em\lower.25em\hbox{Y}\kern-.125emX\@}

	\usepackage{babel}                  

\usepackage{csquotes}
\PassOptionsToPackage{%
	backend=bibtex,bibencoding=utf8,%
	language=auto,%
	style=authoryear-comp,%
    sorting=nyt, 
    maxbibnames=10, 
    natbib=true 
}{biblatex}
    \usepackage{biblatex}

\renewcommand{\cite}[1][]{\parencite[#1]}

\DeclareCiteCommand{\parencite}
  {\boolfalse{citetracker}%
   \boolfalse{pagetracker}%
   \usebibmacro{prenote}}
  {\ifciteindex
     {\indexnames{labelname}}
     {}%
   \printtext[bibhyperref]{\printtext[parens]{\printnames{labelname} {\printdate}}}}
  {\multicitedelim}
  {\usebibmacro{postnote}}

\newcommand\rev[1]{\textcolor{black}{#1}}

\PassOptionsToPackage{fleqn}{amsmath}       
    \usepackage{amsmath}

\PassOptionsToPackage{T1}{fontenc} 
    \usepackage{fontenc}     
\usepackage{textcomp} 
\usepackage{scrhack} 
\usepackage{xspace} 
\usepackage{mparhack} 
\usepackage{fixltx2e} 
\PassOptionsToPackage{printonlyused,smaller}{acronym} 
    \usepackage{acronym} 
    
\newcommand\blfootnote[1]{%
    \bgroup
    \renewcommand\thefootnote{\fnsymbol{footnote}}%
    \renewcommand\thempfootnote{\fnsymbol{mpfootnote}}%
    \footnotetext[0]{#1}%
    \egroup
}

\usepackage{tabularx} 
\usepackage{caption}
\usepackage{subfig}  

\usepackage{listings} 
\PassOptionsToPackage{pdftex,hyperfootnotes=false,pdfpagelabels}{hyperref}
    \usepackage{hyperref}  
\PassOptionsToPackage{pdftex}{graphicx}
    \usepackage{graphicx}

\hypersetup{%
    colorlinks=true, linktocpage=true, pdfstartpage=3, pdfstartview=FitV,%
    breaklinks=true, pdfpagemode=UseNone, pageanchor=true, pdfpagemode=UseOutlines,%
    plainpages=false, bookmarksnumbered, bookmarksopen=true, bookmarksopenlevel=1,%
    hypertexnames=true, pdfhighlight=/O, 
    urlcolor=webbrown, linkcolor=RoyalBlue, citecolor=Maroon, 
    pdftitle={\myTitle},%
    pdfauthor={\textcopyright\ \myName, \myUni, \myFaculty},%
    pdfsubject={},%
    pdfkeywords={},%
    pdfcreator={pdfLaTeX},%
    pdfproducer={LaTeX with hyperref and classicthesis}%
}

\makeatletter
\@ifpackageloaded{babel}%
    {%
       \addto\extrasamerican{%
                }%
       \addto\extrasngerman{%
                }%
    }{\relax}
\makeatother

\usepackage{classicthesis} 
\usepackage{overpic}
\usepackage{bm}
\usepackage{mdframed}
\usepackage{afterpage}
\usepackage{caption}
\usepackage{multirow}
\usepackage{mathtools}
\usepackage{backnaur}
\usepackage{comment}
\usepackage{amssymb}  
\usepackage{pifont}   
\usepackage{dialogue}   



%% file: FrontBackmatter/Titlepage.tex
\begin{titlepage}
    \begin{addmargin}[-1cm]{-3cm}
    \begin{center}
        \large  

        \hfill

        \vfill

        \begingroup
            \color{Maroon}\spacedallcaps{\myTitle} \\ \bigskip
        \endgroup

        \spacedlowsmallcaps{\myName}

        \vfill


        \myDegree \\ \medskip 
        \myDepartment \\                            
        \myUni \\ \bigskip

        \myTime 

        \vfill                      

    \end{center}  
  \end{addmargin}       
\end{titlepage}   

%% file: FrontBackmatter/Titleback.tex
\thispagestyle{empty}

\hfill

\vfill

\noindent\myName: \textit{\myTitle,} \mySubtitle \myDegree, 
\textcopyright\ \myTime

\bigskip

\noindent\spacedlowsmallcaps{PhD supervisor}: \\
\mySupervisor, \myUni

\bigskip

\noindent\spacedlowsmallcaps{Reviewers}: \\
Dr. Andrew Lampinen, Senior Research Scientist at Google DeepMind\\
Prof. Alex Bronstein, Technion - Israel Institute of Technology

\bigskip

\noindent\spacedlowsmallcaps{Thesis defended on January 26, 2024,} \\\noindent\spacedlowsmallcaps{before the External Committee}: \\
Prof. Lamberto Ballan, Università degli Studi di Padova \\
Prof. Giovanni Petri, Northeastern University London and CENTAI\\
Prof. Alessandro Raganato, Università degli Studi di Milano-Bicocca

\bigskip

\noindent PhD degree conferred cum laude.

\medskip

\noindent A video of the defense can be found \href{https://www.youtube.com/watch?v=0uD51aHiYlA}{here}.



\medskip
%

%% file: FrontBackmatter/Dedication.tex
\thispagestyle{empty}
\refstepcounter{dummy}
\pdfbookmark[1]{Dedication}{Dedication}

\vspace*{3cm}

\begin{flushright}
    A mamma, che insegna latino e greco\\ ma vuole sempre che le spieghi la mia ricerca. \\Finché non capisce, e allora ho capito meglio anch'io. \\ \medskip
E ho capito meglio cos'è la ricerca. \\ \bigskip\bigskip
    ---  \\ \bigskip\bigskip
    To my mother, who teaches Latin and Greek\\ but always wants me to explain my research. \\Until she understands, and then I've understood it better myself.\\ \medskip
And I've understood better what research truly is.
\end{flushright}

\medskip


%% file: FrontBackmatter/Abstract.tex
\pdfbookmark[1]{Abstract}{Abstract}
\begingroup
\let\clearpage\relax
\let\cleardoublepage\relax
\let\cleardoublepage\relax

\chapter*{Abstract}
Rooted in the explosion of deep learning over the past decade, this thesis spans from AlphaGo to ChatGPT to empirically examine the fundamental concepts needed to realize the vision of an artificial scientist: a machine with the capacity to autonomously generate original research and contribute to the expansion of human knowledge. 

The investigation begins with {\sc Olivaw}, an AlphaGo Zero-like agent that discovers Othello knowledge from scratch but is unable to communicate it. This realization leads to the development of the Explanatory Learning (EL) framework, a formalization of the problem faced by a scientist when trying to explain a new phenomenon to their peers. The effective EL prescriptions allow us to crack Zendo, a popular board game simulating the scientific endeavor. This success comes with a fundamental insight: an artificial scientist must develop its own interpretation of the language used to explain its findings, and not rely on a rigid existing interpreter. Questioning the very process of learning an interpreter, we turn our attention to the inner functioning of modern multimodal models. This culminates in a simple idea to build CLIP-like models where interpretation and perception are explicitly disentangled: a cost-effective approach that couples two unimodal models using little multimodal data and no further training. 
Finally, we discuss what ChatGPT and its siblings are still missing to become artificial scientists, and introduce the Big-Bench Symbol Interpretation Task, a benchmark about interpreting Zendo-like explanations that sees LLMs going no further than random chance while being instead fully solved by humans.

\vfill

\endgroup			

\vfill

%% file: FrontBackmatter/Publications.tex
\pdfbookmark[1]{Publications}{publications}
\chapter*{Publications}


The content of this thesis is largely based on the following articles. In each, I was the main author, leading the research, curating the writing, and conducting most of the experiments. In Big-Bench I was leading our task proposal.

\begin{enumerate}
    \item \textbf{OLIVAW: Mastering Othello without Human Knowledge, nor a Penny} \\
    Antonio Norelli and Alessandro Panconesi \\
    \textit{IEEE Transactions on Games, 2022} \\
    AlphaGo Zero for Othello. With two ideas to speed up the learning, and tested in a live match against a former world champion. \\
    \cite{norelli2022olivaw}

    \item \textbf{Explanatory learning: Beyond empiricism in neural networks} \\
    Antonio Norelli, Giorgio Mariani, Luca Moschella, Andrea Santilli, Giambattista Parascandolo, Simone Melzi, and Emanuele Rodolà\\
    \textit{ICML 2023 Workshop: Knowledge and Logical Reasoning in the Era of Data-driven Learning} \\
    When a ML system becomes an artificial scientist: mastering the game of Zendo with Transformers. \\
    \cite{explanatory}

    \item \textbf{ASIF: Coupled Data Turns Unimodal Models to Multimodal Without Training} \\
    Antonio Norelli, Marco Fumero, Valentino Maiorca, Luca Moschella, Emanuele Rodolà, and Francesco Locatello \\
    \textit{NeurIPS 2023: Conference on Neural Information Processing Systems} \\
    The meaning was already there: connecting text and images without training a neural network to do so.  \\
    \cite{norelli2022asif}

    \item \textbf{Beyond the Imitation Game: Quantifying and Extrapolating the Capabilities of Language Models} \\
    450 authors including Antonio Norelli, Giorgio Mariani, Luca Moschella, Andrea Santilli, Giambattista Parascandolo, Simone Melzi, and Emanuele Rodolà\\
    \textit{Transactions on Machine Learning Research, 2023} \\
    The task with the widest gap between human and machine performance in BIG-bench, a collaborative effort to test Language Models. \\
    \cite{srivastava2023beyond}
\end{enumerate}


I also collaborated in the following publications as part of my PhD research, but they are not covered in this thesis.

\begin{enumerate}
\setcounter{enumi}{4}
    \item \textbf{Relative Representations Enable Zero-shot Latent Space Communication} \\
    Luca Moschella, Valentino Maiorca, Marco Fumero, Antonio Norelli, Francesco Locatello, and Emanuele Rodolà \\
    \textit{ICLR 2023: International Conference on Learning Representations} \\
    It happens that different neural networks trained on the same stuff learn intrinsically equivalent latent spaces. \\
    \cite{moschella2022relative}

    \item \textbf{LIMP: Learning Latent Shape Representations with Metric Preservation Priors} \\
    Luca Cosmo, Antonio Norelli, Oshri Halimi, Ron Kimmel, and Emanuele Rodolà \\
    \textit{ECCV 2020: European Conference on Computer Vision} \\
    With the right geometric prior, 11 samples are enough to train a generative model for 3D shapes of humans or animals. \\
    \cite{cosmo2020limp}

    \item \textbf{Latent Space Translation via Semantic Alignment} \\
    Valentino Maiorca, Luca Moschella, Antonio Norelli, Marco Fumero, Francesco Locatello, and Emanuele Rodolà  \\
    \textit{NeurIPS 2023: Conference on Neural Information Processing Systems} \\
    Translating between pre-trained networks with simple algebraic transformations enables zero-shot stitching of text and vision models.  \\
    \cite{maiorca2023latent}

    \item \textbf{Learning Rotation-Agnostic Representations via Group Equivariant VAEs} \\
    Ahmedeo Shokry and Antonio Norelli \\
    \textit{Tiny paper at ICLR 2023} \\
    Group-equivariant Variational Autoencoders may be useful to produce representations that are invariant to rotations. \\
    \cite{shokry2023learning}

    \item \textbf{Bootstrapping Parallel Anchors for Relative Representations} \\
    Irene Cannistraci, Luca Moschella, Valentino Maiorca, Marco Fumero, Antonio Norelli, and Emanuele Rodolà \\
    \textit{Tiny paper at ICLR 2023} \\
    New parallel anchors to use with relative representations can be discovered from a smaller set through optimization. \\
    \cite{cannistraci2023bootstrapping}

    \item \textbf{Errare Humanum Est? A Pilot Study to Evaluate the Human-likeness of an AI Othello Playing Agent} \\
    Enrico Lauletta, Beatrice Biancardi, Antonio Norelli, Maurizio Mancini, and Alessandro Panconesi \\
    \textit{Proceedings of the 22nd ACM International Conference on Intelligent Virtual, 2022} \\
    Investigating the human-like behaviors and characteristics of \textsc{Olivaw}. \\
    \cite{lauletta2022errare}
\end{enumerate}

%% file: Chapters/Prologue.tex
\chapter*{Prologue}

\begin{quotation}
It has been 25 years since a report of original research was last submitted to our editors for publication, making this an appropriate time to revisit the question that was so widely debated then: what is the role of human scientists in an age when the frontiers of scientific inquiry have moved beyond the comprehensibility of humans?

No doubt many of our subscribers remember reading papers whose authors were the first individuals ever to obtain the results they described. But as metahumans began to dominate experimental research, they increasingly made their findings available only via DNT (digital neural transfer), leaving journals to publish second-hand accounts translated into human language.

Without DNT, humans could not fully grasp earlier developments nor effectively utilize the new tools needed to conduct research, while metahumans continued to improve DNT and rely on it even more. Journals for human audiences were reduced to vehicles of popularization, and poor ones at that, as even the most brilliant humans found themselves puzzled by translations of the latest findings.

No one denies the many benefits of metahuman science, but one of its costs to human researchers was the realization that they would probably never make an original contribution to science again. Some left the field altogether, but those who stayed shifted their attentions away from original research and toward hermeneutics: interpreting the scientific work of metahumans.
\end{quotation}

In the year 2000, on the pages of Nature, the esteemed science-fiction author Ted Chiang depicts a world that appears to offer no more room for human scientists. Superior entities has taken the lead in research, and communicate their findings only within themselves through an inaccessible medium \cite{chiang2000catching}.

This scenario seems the one we will be condemned to with the advent of artificial scientists: in the future, machines alone will be responsible for making the discoveries necessary to refine our tools and therapies. They are faster, with greater memory, capable of making associations beyond our reach. Most importantly, they will not be hampered by the bottleneck of our language, free to surpass our reasoning limits.

Are we condemned to not understand future breakthroughs, to give up on the once glorious practice of science as a species?

\begin{flushright}{\slshape    
    Se non ci divertiamo a risolvere questo problema\\ perché dovremmo studiarlo?\footnote{\href{https://youtu.be/GoYimmmFptg?si=apIsFhWgKmjwrEx0&t=1546}{As recalled} by his student Giorgio Parisi, the 2021 Nobel laureate in Physics, who happened to hand me my Bachelor's degree in Physics. In English: \textit{If we don't have fun solving this problem, why should we even study it?}}} \\ \medskip
    --- Nicola Cabibbo (\href{https://en.wikipedia.org/wiki/Nicola_Cabibbo}{Wiki})
\end{flushright}

Not at all. And not because we will not have those formidable machines. 
This fear comes from a fundamental misrepresentation of what science truly is: a playful pursuit of our curiosity. 

That stands as a vital part of the human culture–-as Ted Chiang reminds us in the latter half of his story-–rooted in creativity, experience, and, ultimately, communication. 

Science is the art of explaining nature to humans, there is no science beyond our language.\footnote{This is less restrictive than it might seem, as our language is not a rigid cage but rather a dynamic entity that evolves to meet our communicative needs. As we will discuss and experimentally test in the following.}

Nevertheless, machines can partake in this infinite quest alongside us. Indeed, this is the perspective we will adopt in the following pages: How can we devise a machine capable of making its own discoveries and then effectively explain them to us?

\newpage

%% file: FrontBackmatter/Contents.tex
\refstepcounter{dummy}
\pdfbookmark[1]{\contentsname}{tableofcontents}
\setcounter{tocdepth}{2} 
\setcounter{secnumdepth}{3} 
\manualmark
\markboth{\spacedlowsmallcaps{\contentsname}}{\spacedlowsmallcaps{\contentsname}}
\tableofcontents 
\automark[section]{chapter}
\renewcommand{\chaptermark}[1]{\markboth{\spacedlowsmallcaps{#1}}{\spacedlowsmallcaps{#1}}}
\renewcommand{\sectionmark}[1]{\markright{\thesection\enspace\spacedlowsmallcaps{#1}}}
\clearpage

\begingroup 
    \let\clearpage\relax
    \let\cleardoublepage\relax
    \let\cleardoublepage\relax
    \refstepcounter{dummy}
    \pdfbookmark[1]{\listfigurename}{lof}
    \listoffigures

    \vspace{8ex}

    \refstepcounter{dummy}
    \pdfbookmark[1]{\listtablename}{lot}
    \listoftables
        
    \vspace{8ex}
    

       
\endgroup

%% file: Chapters/Chapter01.tex
\chapter{Introduction and overview}\label{ch:introduction}

\section{Devising an artificial scientist}

Explanations are the fuel of progress, the fundamental tool through which humans have increased their agency, earning more and more control over their future throughout history. So far, the production of these extraordinary symbolic sequences has been a unique prerogative of human scientists, but the formidable breakthroughs in AI that followed the advent of deep learning \cite{LeCun2015DeepLearning} evoke the idea of machines capable of assisting us in this endeavour. Not mere tools empowering scientists, but rather peers capable of producing original research and pushing forward knowledge autonomously.

In this thesis I seriously entertain this idea, and seek to understand what it means to build an artificial scientist. 

We will relive my research route,\graffito{Chapter 2\\ \textsc{Olivaw}: Mastering Othello with neither Human Knowledge nor a Penny} that initiated with an in-depth study of AlphaGo Zero. This AI system was able to master Go starting from scratch and defeated the best players in the world, a very promising starting point to investigate the possibility of knowledge creation by machines. Our discussion will begin with {\sc Olivaw}, an AlphaGo Zero-like agent that I brought to retrace the journey of its illustrious predecessor on the game of Othello, to the point of challenging a former World champion. Narrated with a blend of rigor and humor, this adventure culminates in a pivotal realization: knowledge is such if it can be transmitted to us, and AlphaGo cannot write a book on Go that could help human players improve.

A fundamental characteristic\graffito{Chapter 3\\ Explanatory Learning: Beyond Empiricism in Neural Networks} of a scientist is the ability to communicate their own discoveries.
This will lead us to realize the tension between language and the effective representation of new natural phenomena: language must be adapted to the new communicative need, but at the same time it should remain understandable; its vocabulary and rules slowly but inevitably change over time. 
Therefore, we will attempt to model an agent capable of explaining its observations without strictly defining its language. As a result, we will introduce Explanatory Learning and place the keystone of this thesis: a true artificial scientist can only emerge when a machine can autonomously interpret symbols.

This realization will drive us beyond traditional AI methodologies, such as Inductive Logic Programming and Program Synthesis, which approach the challenge of creating intelligent agents by presupposing the existence of a rigid, human-coded language interpreter. Instead, Explanatory Learning ditches this assumption, leaving us with a sparse assortment of symbolic sequences associated to observations of various phenomena as the sole resource to build an interpreter. It calls for an autonomous accomplishment of the mastering of a language through learning. An autonomous construction of the map between signs and meanings, that stays adjustable as it had to be during its creation, not a given rigid one carved in stone. 

Having reached this peak,\graffito{Chapter 4\\ ASIF: Coupled Data Turns Unimodal Models to Multimodal without Training} and realized that the interpreter is a fundamental piece of an artificial scientist, we will narrow our focus on it, analyzing how this map between signs and their meanings can be created and adjusted. Deep learning has provided us with glorious examples of the creation of this map, like the contrastive learning of CLIP \cite{clip}, resulting in an unprecedented ability to connect text with images. Still, CLIP's creation process is very opaque: countless and repeated weight updates under the direction of a gradient with cascades of non-linear dependencies; it seems almost the technology of the metahumans mentioned in the prologue. We will give it back to humans, downgrading neural networks to mere sensors and building a map between signs and meanings comparable in performance to CLIP \cite{clip} but in broad daylight, with a transparent and simple algorithm, based on a formalization of the word “as” in the context of representation learning. That is, taking the vector representing an image, and considering it as if it were the vector representing its ideal caption. 

Finally,\graffito{Chapter 5\\ Are Large Language Models Sparks of Artificial Scientists?} after having recognized the centrality of language in scientific discovery, we will not ignore what is arguably the most impressive product of deep learning as of today–-based on a trivial but incredibly effective model of language as just a matter of being good in predicting the next word–-Large Language Models. They took the scene of AI research only in 2020 with GPT-3 \cite{gpt2020}, and three years later GPT-4 is already hailed as a spark of Artificial General Intelligence \cite{bubeck2023sparks}, i.e. a machine that could learn to accomplish any intellectual task that human beings or animals can perform. And therefore also science. The path has now been taken, and are we a step away from the artificial scientist able to autonomously push forward human knowledge, as we imagined a few lines above? In the last chapter of this thesis, we will discuss the impressive capabilities of Large Language Models, but also examine the tension between how they function and the principles of scientific practice. This culminates in the resounding failure of LLMs on the Symbol Interpretation Task, a benchmark based on Odeen designed to test the ability to reason and interpret symbols. Fully solved by humans, it turned out to be the task with the largest gap between human and machine performance in the Big-Bench collection \cite{srivastava2023beyond}.

%% file: Chapters/Chapter02.tex
\chapter{{\scshape Olivaw}: Mastering Othello without Human Knowledge, nor a Fortune}\label{ch:olivaw}
\chaptermark{{\scshape Olivaw}: Mastering Othello without Human Knowledge}

\section*{Chapter abstract}
We introduce {\scshape Olivaw}, an AI Othello player adopting the design principles of the famous AlphaGo programs. The main motivation behind {\scshape Olivaw} was to discover knowledge from scratch in a non-trivial board game at a tiny fraction of the cost of its illustrious predecessors.
In this chapter, we 
show how the AlphaGo Zero's paradigm can be successfully applied to the popular game of Othello using only commodity hardware and free cloud services.
While being simpler than Chess or Go, Othello maintains a considerable search space and difficulty in evaluating board positions.
To achieve this result, {\scshape Olivaw} implements some improvements inspired by recent works to accelerate the standard AlphaGo Zero learning process.
The main modification implies doubling the positions collected per game during the training phase, by including also positions not played but largely explored by the agent.
We tested the strength of {\scshape Olivaw} in three different ways: 
by pitting it against Edax, considered by many the strongest open-source Othello engine,
by playing anonymous games on the web platform OthelloQuest, and finally in two in-person matches against top-notch human players: a national champion and a former world champion.

\blfootnote{This chapter is based on the paper \textit{"{\scshape Olivaw}: Mastering Othello without Human Knowledge, nor a Fortune"}, by Antonio Norelli and Alessandro Panconesi.}

\newpage

\section{Introduction}

Only a year after AlphaGo's landmark victory against Go master Lee Sedol another sensational development took place.  
An improved version of  AlphaGo called AlphaGo Zero asserted itself as the strongest Go player in the history of the game  \citep{Silver2017MasteringKnowledge}. The remarkable feature of  AlphaGo Zero was that, unlike its predecessor and unlike all previous game software,  it learned to master the game entirely by itself, without any human knowledge. As subsequent follow-up work quickly showed, AlphaGo's paradigm-- an interesting blend of deep and reinforcement learning-- seems to be general and flexible enough to adapt to a wide array of games
\citep{Silver2018ASelf-play., muzero-schrittwieser2019mastering}.

These extraordinary successes came at a price however, and quite literally so. The amount of computational and financial resources that were required was so huge as to be out of reach for most academic and non-academic institutions. Not coincidentally these well-endowed projects and their follow-ups took place within giant multinational corporations of the IT sector \citep{tian2019elf,leeminigo}.  These companies deployed GPUs by the thousands and hundreds of TPUs. A recent study looked at the number of 
petaflops per day  that were required to train AlphaGo Zero and other recent well-known results in AI \citep{amodei2018ai}. The paper shows an exponential growth with a 3.4-month doubling period. This is clearly unsustainable for most academic labs and departments and even the greatest majority of companies. Another aspect of the same problem is the amount of training needed. 
AlphaGo Zero required 4.9 million games played during self-play. And in order
to attain the level of grandmaster for games like Starcraft II and Dota 2 the training required 200 years and more than $10,000$ years of gameplay, respectively \citep{vinyals2019grandmaster, pachockiopenai}.

Thus one of the major problems to emerge in the wake of these breakthroughs is whether comparable results can be attained at a much lower computational and financial cost and with just commodity hardware. In this chapter we take a small step in this direction, by showing that AlphaGo Zero's successful paradigm can be replicated for the game of Othello (also called Reversi). While being much simpler than either Chess or Go, this game is still rather sophisticated and has a considerable strategic depth.  The game enjoys a long history and a rich tradition. Every year an exciting world championship takes place in which accomplished players from all over the world vie for the world title.  

Our Othello engine is called {\sc Olivaw}, a homage to the famous robot character invented by Isaac Asimov.
We tested the strength of {\sc Olivaw} in three different ways. 
In one instance, we pitted {\sc Olivaw} against Edax, one of the strongest Othello engines. Perhaps the most interesting aspect of this set of matches was that {\sc Olivaw} managed to beat several times an opponent that explores tens of millions of positions in the game tree in the course of a single game. In contrast, {\sc Olivaw}'s search of the game tree was limited to a couple of thousand positions.

We also tested {\sc Olivaw} against (presumably) human players of varying strength on the web platform OthelloQuest. But the most exciting challenge was a series of matches against top-notch human players: a national champion and a former world champion.

The final outcome shows that in a relatively short training time {\sc Olivaw} reached the level of the best human players in the world. Crucially, this has been achieved by using very limited resources at very low overall cost: commodity hardware and free, and thus very limited, cloud services.

\section{Related work}

The success of AlphaGo naturally stimulated several follow-ups.
One of the main questions was to determine the level of generality of the approach. A series of papers showed this level to be great indeed. 
One after the other a list of difficult games fell pray of the RL-with-oracle-advice approach.
\citet{Silver2018ASelf-play.} extended it to Chess and Shogi. 

Recently \citet{muzero-schrittwieser2019mastering} added ATARI games to the list.  Our work continues this line of research by adding $8 \times 8$ Othello to the list,  paying special attention to the cost issue. 
Indeed, it is not clear a priori whether the approach scales down in terms of resources.  Although cheaper in some ways, the agents in  \cite{Silver2018ASelf-play.}, \cite{muzero-schrittwieser2019mastering},
\cite{tian2019elf}, and \cite{leeminigo}
still use thousands of GPU's or hundreds of TPU's to master board games.
The recent KataGo \cite{katago-wu2019accelerating} reaches the level \rev{of play} of ELF using 1/50 of the computation and implements several techniques to accelerate the learning. However, these include a set of targets crafted by humans which are very game-specific\footnote{For instance a ladder indicator, where ladders are Go-specific tactics}
thereby reducing the generality of the approach and reintroducing human knowledge in a relevant way.

Successful low-cost reproductions of AlphaGo Zero came out in recent years, but only for very simple games like Connect-4 \cite{connect4MediumLessonsAlphaZero} or $6 \times 6$ Othello \cite{chang2018big}, for which perfect strategies are known.

Other works focused on the hyperparameters involved in AlphaGo Zero, looking for a faster and cheaper training process. \citet{wang2019hyper} and \citet{connect4MediumLessonsAlphaZero} make several experiments in this direction, while \citet{wu2020population-based-accelerating} investigates the possibility of tuning hyperparameters within a single run, using a population-based approach on Go. 
We followed some insights provided by these works as reported in section \ref{sec-training_process}.

Concerning the state-of-the-art of $8 \times 8$ Othello engines,
algorithms became superhuman before the deep learning era, a fact heralded by the defeat of the world champion Takeshi Murakami at the hands of Logistello \cite{buro1995logistello}. Today the strongest and most popular programs used by top Othello players for training and game analysis are Saio \cite{RomanoBenedetto2009SAIO:Dellothello} and the open source Zebra \cite{zebra} and Edax \cite{Edax}. 
As in chess before AlphaZero \cite{Kasparov2018ChessReasoning.},
in order to reach the level of world-class players, Othello engines rely on a highly optimized minimax search \cite{neumann1928theorie} and employ handcrafted evaluation functions based on knowledge of the game accumulated by human beings\footnote{
For instance, patterns on the edge of the corner of the board, which are known to be of great importance in Othello.} \cite{zebra}. 
Another key feature used by these more traditional engines are huge catalogs of opening sequences, distilled and stored in the course of decades by human players and, more recently, by software as well.
Finally, typically these engines play the perfect game by sheer computational brute force by expanding the entire game tree for a large portion of the game, typically starting 20-30 moves before the end.
To summarize, unlike {\sc Olivaw}, traditional Othello engines make crucial use of knowledge of the game accumulated by humans over many years and of a massive brute force tree exploration. These are two important limitations {\sc Olivaw} is attempting to overcome.

To the best of our knowledge, the only approach similar to ours in spirit is a paper by
\citet{liskowski2018learning} which presents an engine obtained by training a convolutional neural network (CNN) with a database of expert moves. The engine however was only able to defeat Edax 2-ply (tree search limited to two moves ahead), a level of Edax that is much weaker than top Othello human players.

\section{Othello}

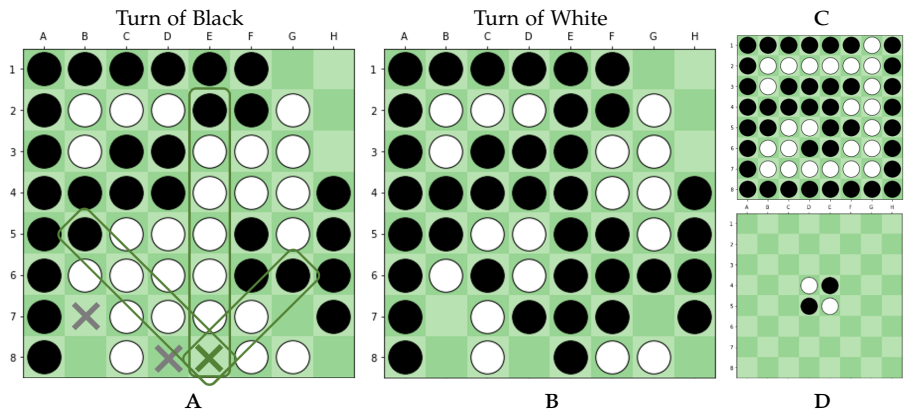
\begin{figure}[h]
\vspace{2.0em}
\begin{overpic}
		[trim=0cm 0cm 0cm 0cm,clip,width=1.\linewidth]{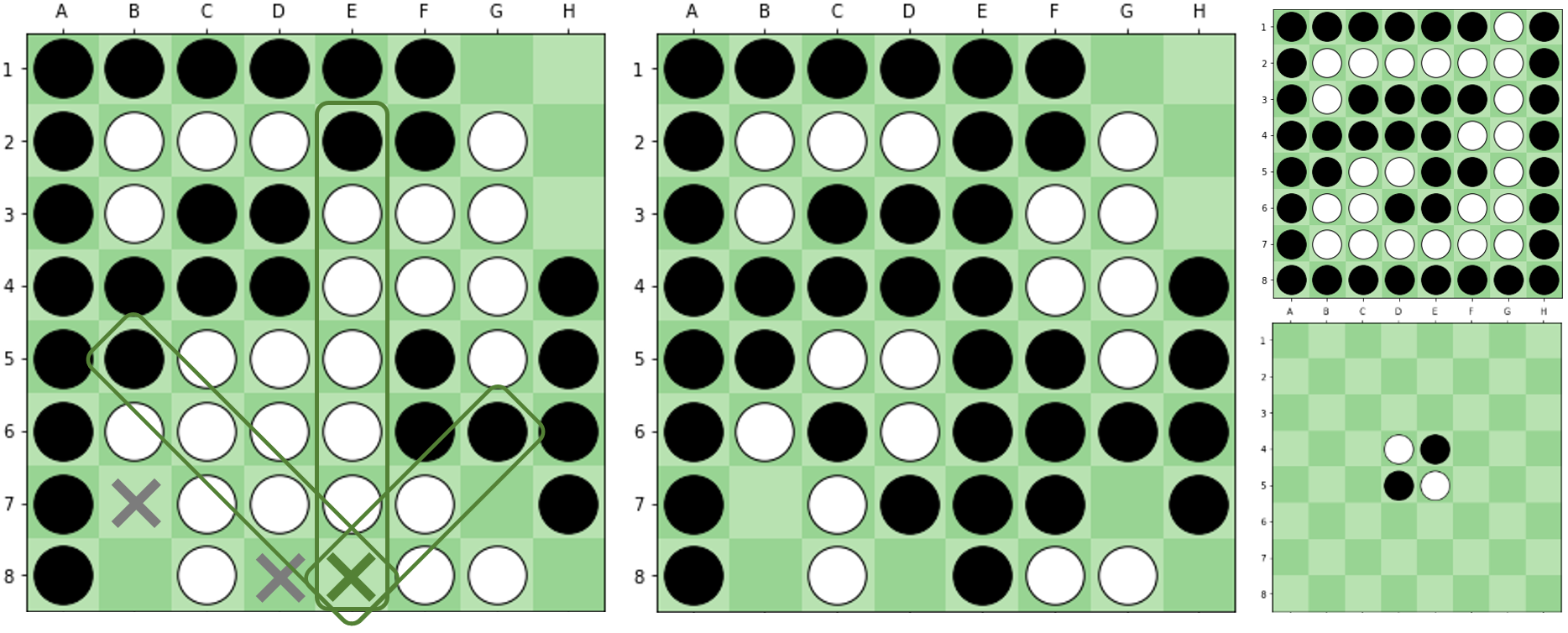}
		\put(18.7,-2.05){ \textsc{a}  }
		\put(12,41){\scriptsize Turn of Black  }
		\put(51.9,41){\scriptsize Turn of White  }
		\put(88.8,41){ \textsc{c}  }
		\put(58.8,-2.05){ \textsc{b}  }
		\put(88.8,-2.05){ \textsc{d}  }
\end{overpic}
\vspace{.05em}

\centering
\caption[Rules of Othello]{\textsc{Rules of Othello.} Othello is a turn-based game where the black and white player try to overcome each other in the final domination of an 8x8 board. \textsc{a.} Players move alternately by placing a new disk in an empty square in order to bracket one or more opponent’s disks between the played disk and another of its own color already on the board. It is possible to capture disks horizontally, vertically, and diagonally. Disks can be captured in one or more directions in a single move, with capture always occurring in a straight line. Only moves capturing at least one disk are allowed. In the absence of moves the player must skip the turn. It is not possible to pass the turn if there is at least one valid move. \textsc{b.} The imprisoned disks change color and become owned by the player who moved. \textsc{c.} When none of the players can move, for instance when the board is full, the player with more disks on the board wins. Here black wins 40-24. \textsc{d.} A game of Othello begins with 4 disks placed in the center of the board in the shape of an X. Black moves first.
}
\label{fig:rules}
\end{figure}

Othello is a popular board game. 
Its simple rules are explained in Figure \ref{fig:rules}.
A typical game lasts for some 60 moves, with an average branching factor of 10. Like Go and Chess it is a perfect information game. Although simpler than these two, it has considerable strategic depth. Unlike English draughts, there is no known perfect strategy that can be played by computer \cite{mullins2007checkers}. 

Othello is played across the globe. There are professional players competing in official tournaments organized by world and national federations. The Othello world championship takes place every year.

The best software beat humans systematically but, as discussed, they rely on brute force for most of the game. During the initial stages of the game, they access huge databases of openings, which are known to be very important in Othello. An opening lasts between 10 and 20 moves. Furthermore, some 20-30 moves before the end they play the perfect game by exploring the game tree in its entirety. In the middle game too, these programs explore hundreds of millions of positions of the game tree per move, a clear indication of brute force at play. 
In contrast, as we shall see, {\sc Olivaw} explores only a few hundreds positions and does not use any database of openings. 

\section{{\scshape Olivaw}: the algorithm}

The design of {\sc Olivaw}, our Othello engine, follows closely that of AlphaGo Zero \cite{Silver2017MasteringKnowledge}. The main difference consists of a somewhat different and cheaper training process. The network architecture, while mimicking that of AlphaGo Zero, was scaled down.
Before discussing {\sc Olivaw} in detail, it is useful to describe its basic design.

\subsection{The basic design}

Like the AlphaGo \rev{programs},  {\sc Olivaw} uses reinforcement learning to build an ``oracle'',
in the form of a deep network $f_\theta$ ($\theta$ denotes the weights of the neural network).  Given as input an Othello game state $s$, $f_\theta$ outputs a pair: $f_\theta(s) = (\bm{p}, v)$. The vector
$\bm{p}$ is a probability distribution over the possible moves from $s$. Intuitively, the higher the probability the better the move. The value $v$ is the oracle's assessment of how good state $s$ is, ranging from $+1$ (sure victory) to $-1$ (certain defeat). 

The oracle is used to guide an exploration of the 
\rev{``possible near futures''} by a Monte Carlo Tree Search (MCTS) \cite{mcts-browne2012survey}. To pick the next move, the game tree rooted at $s$ is explored. Roughly speaking, in this exploration, the moves that $f_\theta$ considers good are explored first (so that the  \rev{actual branching factor is limited}) and the total number of nodes explored is limited (in the few hundreds during training and set to one thousand when playing against humans).
The goal of this exploration phase is to produce a better estimate $(\bm{\pi}, q)$ of state $s$.
When this is done, the best move according to $\bm{\pi}$ is played to reach a new state $s'$, and the process is repeated.

What is noteworthy about this process is that while by itself $f_\theta$ is a rather weak player, using it in combination with MCTS gives rise to a very strong one, i.e. the estimates $(\bm{\pi}, q)$ are more reliable than $(\bm{p}, v)$.
Let us call $A(f)$ the MCTS playing agent using $f$ as oracle.

The crux of the approach is to generate a sequence of oracles
$f_0,$ $ f_1, $ $f_2, $ $\ldots, $ $f_t$ each better than the predecessors. This is done by generating a sequence of training sets $S_1, S_2, \ldots, S_t$ each better than the previous one. Training set $S_i$ is used to train $f_i$. The process is initialized with a deep network $f_0$ with random weights.

The generic step in this sequence of improvements is as follows. Let $f_\theta$ be the current oracle. 
During the so-called \emph{self-play phase}, $A(f_\theta)$ plays a batch of games against itself.  During each game a set of states $S$ will be explored. For each $s \in S$ an updated (and hopefully better) assessment $(\bm{\pi}_s, q_s)$ will be computed.
The set $T$ of pairs $\{s, (\bm{\pi}_s, q_s)\}$ for $s \in S$ will be added to the training set. The intuition is that this way we can create a virtuous circle. As the assessments $(\bm{\pi}_s, q_s)$ become more and more accurate the training set becomes better and better. And, as the training set improves the assessment becomes more accurate.

We remark \rev{that} the main difference between {\sc Olivaw} and AlphaGo Zero resides in how this training set is constructed. Instead of 
$\{s,$ $ (\bm{\pi}_s, $ $ q_s)\}$, AlphaGo Zero only considers pairs 
$\{s, (\bm{\pi}_s, z_s)\}$ where $s$ is actually played during the game, and 
$z_s$ is the outcome at the end of the game. Thus, $z_s \in \{-1, 0, +1\}$. In contrast, besides this type of pairs, {\sc Olivaw} also adds to the training set pairs $\{s, (\bm{\pi}_s, q_s)\}$ for which $s$ has been explored ``a lot''. In this way, we collect a larger training set for the same number of simulated games.
This is crucial since the cost of the \emph{self-play} phase is the main contributor to the overall cost. This design choice is discussed in detail in section \ref{sec-comparison}.

Once the new training set is obtained we switch to the \emph{training phase}. The current neural network $f_\theta$ is further trained with the updated training set.
At the end of this training we have a new configuration $f_\theta'$. We want $(\bm p, v) = f_\theta'(s)$ to be close to $($\boldmath$\pi\,$\unboldmath$, \omega)$, where $\omega$ can be $z \;\mathrm{or}\; q$. So we minimize the training loss:

\begin{equation}
    L(\bm{\pi}, \omega, \bm{p}, v) = (\omega - v)^2 - \bm{\pi}^T\log{\bm{p}} + c||\theta||^2
\end{equation}

As in AlphaGo Zero, we combine evenly the squared error on the value and the cross-entropy on the move probabilities, while the last term penalizes large weights in the neural network ($c=10^{-4}$). This L2 regularization is used to prevent overfitting over the many training phases.

In the final \emph{evaluation phase}, we verify whether the new $f_\theta'$ is stronger than the old $f_\theta$. To do so, we pit $A(f_\theta')$ against $A(f_\theta)$.
If the former wins significantly more often than the latter it becomes the new oracle. Otherwise we go through the training phase again to produce a new challenger.
The whole process is then repeated. And so on, so forth.
For a more detailed description of the reinforcement learning algorithm we refer to \citet{Silver2017MasteringKnowledge}.

\subsection{Low-cost faster training}

\label{sec-comparison}

With respect to AlphaGo Zero, {\sc Olivaw} introduces three main modifications in the training phase.  

As remarked, while the training set of AlphaGo consists only of
pairs of the kind $\{s, (\bm{\pi}_s, z_s)\}$, where $s$ is a move actually played during self-play and $z_s$ is the outcome at the end of the game, 
{\sc Olivaw} also considers pairs of the type $\{s, (\bm{\pi}_s, q_s)\}$, where $s$ is a position in the game tree that has been explored a number of times above a certain threshold.
The threshold value is set dynamically in order to have a training set twice the size of that used by AlphaGo. In other words, the number of pairs of type $\{s, (\bm{\pi}_s, q_s)\}$ is roughly equal to that of the pairs of type $\{s, (\bm{\pi}_s, z_s)\}$. The pairs added are the ones with the largest number of visits. 
Our approach was broadly inspired by the results reported in \citet{connect4MediumLessonsAlphaZero}.

Adding noisy pairs might not seem a good idea at first. In fact, 
using only the final outcome $z$ as a signal has a big drawback. In a game with multiple errors, every evaluation of an early position based on the final outcome is almost random, while $q$ offers a better assessment. 

\begin{figure*}[h!]
\includegraphics[width=0.59\textwidth]{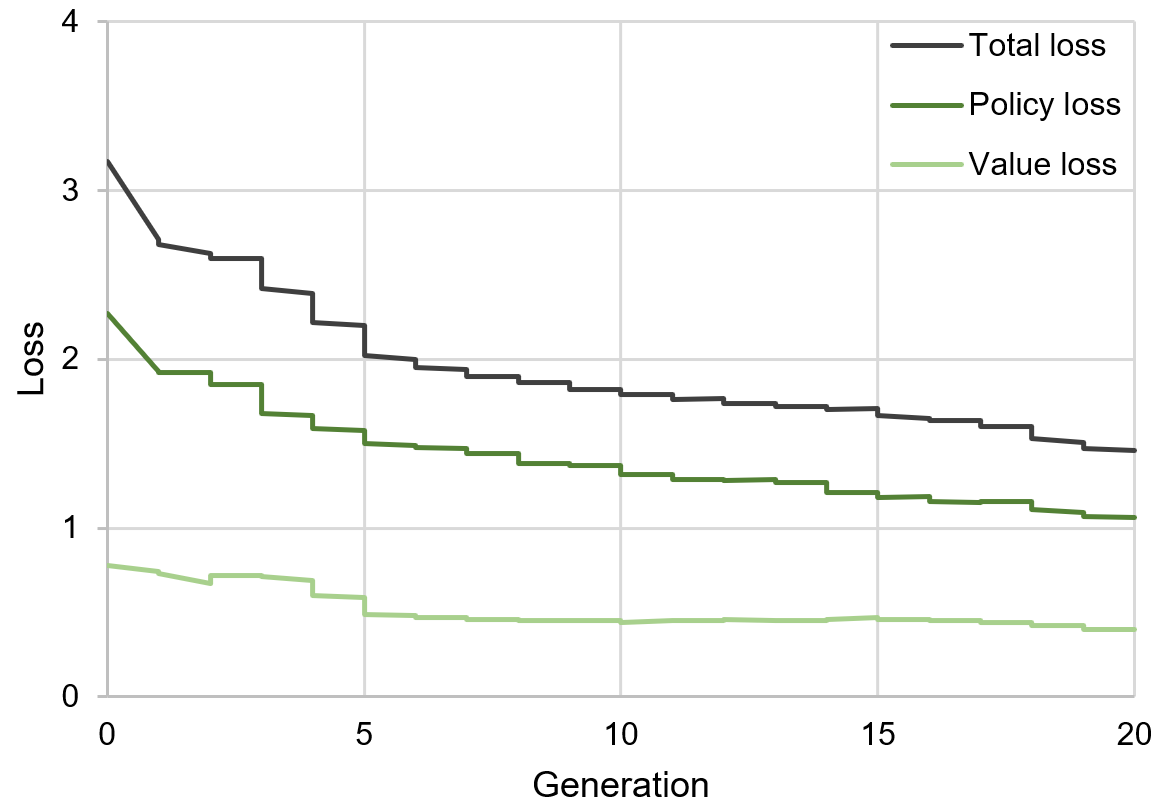}
\includegraphics[width=0.59\textwidth]{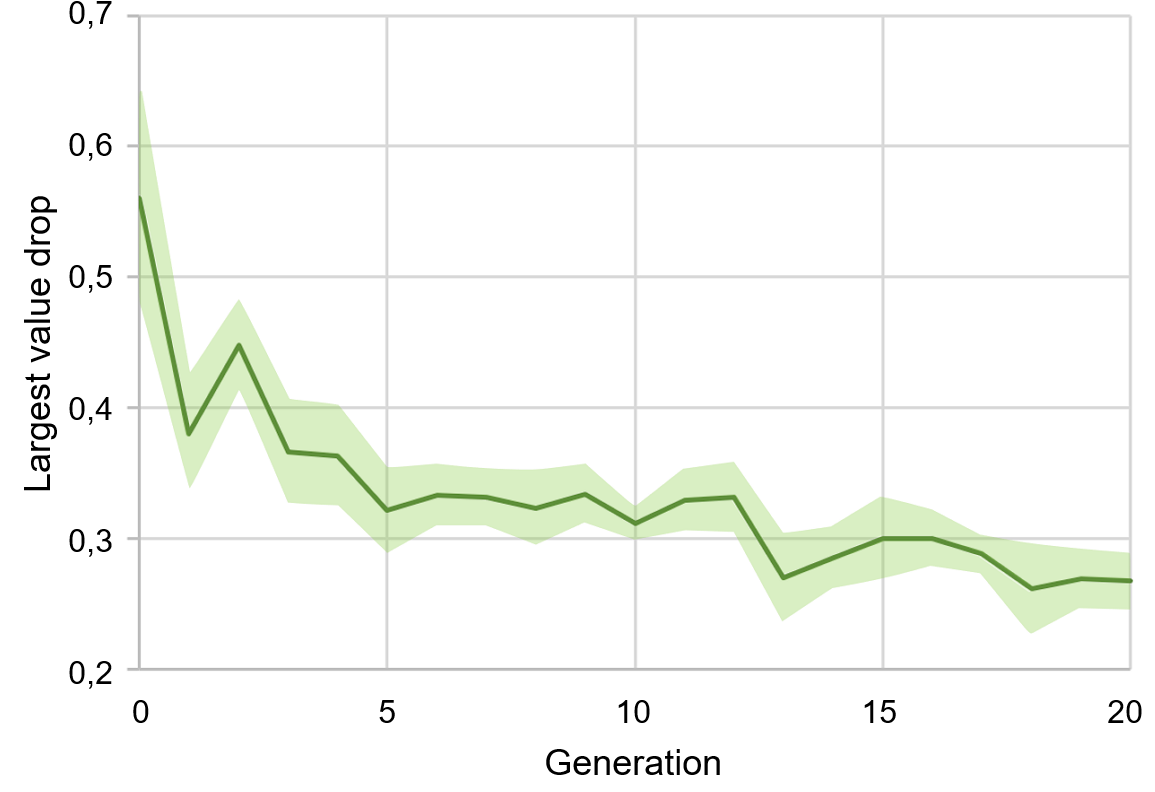}
\includegraphics[width=0.59\textwidth]{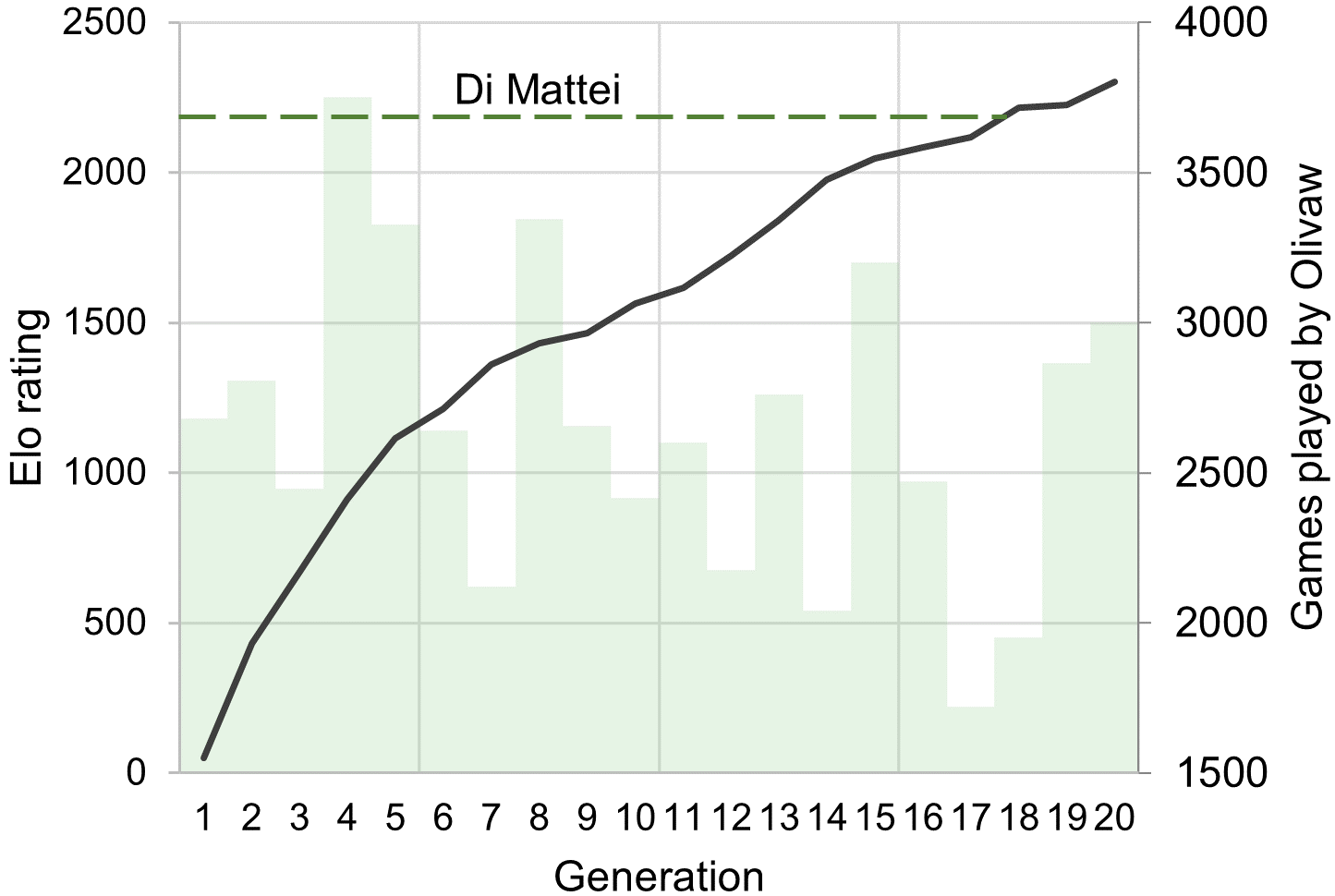}
\centering
\begin{overpic}
		[trim=10cm 34cm 2cm 0cm,clip,width=1.0\linewidth]{images-olivaw/crucial_moves.png}
		\put(83.6,105){ \textsc{a}  }
		\put(83.6,65){ \rev{\textsc{b}}  }
		\put(83.6,23){ \textsc{c}  }
\end{overpic}
\caption[{\sc Olivaw}'s Training process.]{\textsc{Olivaw's Training process.} \textsc{a.} Training loss across generations. The stepwise trend is due to the shifting training window. \textsc{b.} Absolute largest value drop in a game across generations. We show the averages using standard deviation as confidence interval. \textsc{c.} Performance of the i-th generation MCTS agent \rev{(continuous line) and number of games played by {\sc Olivaw} against itself during training in each generation (bar chart)}. The ELO ratings were computed using the evaluation games and the first match vs the national champion Alessandro Di Mattei.
}
\label{fig:trends}
\end{figure*}

On the other hand, $q$ suffers from the limited horizon of the \rev{search}; an early position with positive or negative consequences far ahead in the game may not be properly evaluated. So, a combination of the two signals might strike a better balance.

The second variation concerns a dynamic adjustment of the MCTS.
During the \emph{self-play phase} {\sc Olivaw} selects each move after 100, 200, or 400 MCTS simulations from the current game state, using a higher number of simulations in the higher generations\footnote{{\sc Olivaw} switched to 200 simulations between the 4th and the 5th generation, and to 400 simulations between the 11th and 12th generation.}, 
\rev{as opposed to AlphaGo Zero that stays with $1600$ simulations throughout the training run.}
The rationale is to move quickly away from early generations, where a dataset generated by low-depth MCTS still provides enough signal for an improvement, as noted by Wang, \emph{et al.} \citep{wang2019hyper}.

Finally, {\sc Olivaw} uses a dynamic training window. The training set is a sample of $16,384,000$ positions (minibatch size of 1024) from the games generated by the last generations. We gradually increase the generations included in the training window from the last two to the last five. The idea is to exclude quickly the games played by the first very weak generations. AlphaGo Zero uses as training set a sample of $2,048,000$ positions from the last $500,000$ games, always taken from games generated by the last 20 generations. This small modification proved effective in the Connect4 implementation of AlphaZero by \citet{connect4MediumLessonsAlphaZero}.

\section{Resources}

{\sc Olivaw} was entirely developed, trained, and tested on Colaboratory, a free Google cloud computing service for machine learning education and research \cite{colaboratory_carneiro2018performance}.

{\sc Olivaw} code is completely written in Python, from the core MCTS and Neural Network classes implemented in Numpy
and Keras, 
to the simple GUI based on Matplotlib. 
The \emph{self-play}, \emph{training} and \emph{evaluation phases} take place on three self-contained distinct notebooks sharing the same memory.
Concerning local resources, we took no more advantage than a laptop equipped with a browser and an Internet connection.

The hardware specifications of a Colaboratory virtual machine at the time of the training were:
\begin{itemize}
    \item CPU: 1 single core hyper threaded Xeon Processor, 2.3Ghz, 2 threads.
    \item RAM: $\sim$12.6 GB.
    \item Hardware accelerators (if used):
    \begin{itemize}
        \item GPU: 1 Nvidia Tesla K80, 2496 CUDA cores, 12GB GDDR5 VRAM.
        \item TPU v2-8: Google Tensor processing unit equipped with 8 TPU cores.  
    \end{itemize}
\end{itemize}
The generation of games during the \emph{self-play phase} is the most computationally expensive process of the learning algorithm.
In our hardware configuration, a single game takes from 6 to 30 seconds, depending on the number of MCTS simulations per move, 
due to the high number of GPU calls. \rev{The MCTS run on the CPU, while the deep network $f_\theta$ run on the GPU.}

Game generation can be naturally performed in parallel, by running several instances of the self-play notebook and saving the generated datasets on shared memory. This result has been achieved thanks to an informal crowd computing project, made possible by the ease with which Colaboratory can be shared. 19 people contributed to the generation of games with their Google accounts, also using smartphones.

The \emph{training phase} cannot be executed in parallel but can be accelerated using the TPU runtime of Colaboratory, the acceleration factor is approximately 10 with respect to a GPU K80. A single training phase required $\sim$1.5 hours.

The \emph{evaluation phase} consisted of 40 games between agents of different generations and took less than an hour of GPU run time.

\begin{figure*}[t]
\begin{overpic}[trim=0cm 0.9cm 3cm 0cm,clip,width=1.0\linewidth]{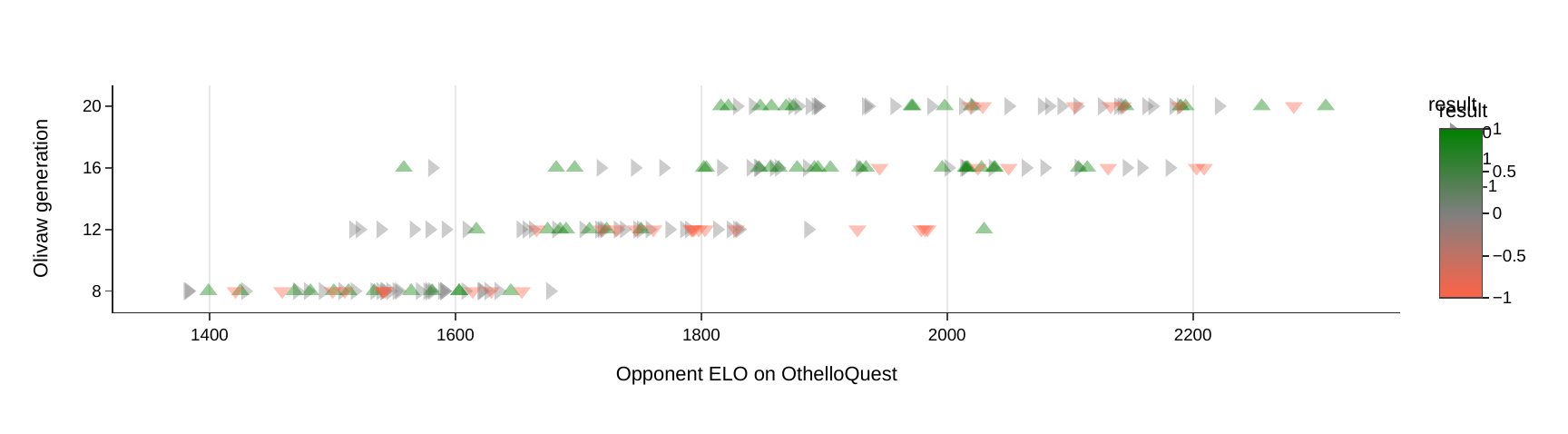}
\put(87, 6){\color{white}
\frame{\includegraphics[trim=0cm 0cm 0cm 0cm,clip,width=0.1\linewidth]{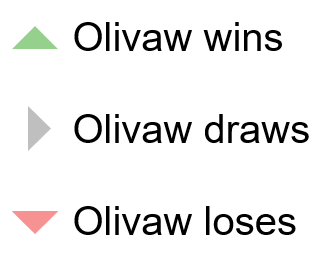}}}
\end{overpic}
\caption[{\sc Olivaw}'s performance on OthelloQuest]{\textsc{{\sc Olivaw}'s performance on OthelloQuest.} 
\rev{The score of different generations of {\sc Olivaw} on OthelloQuest. 
We report the outcome of the last 50 games played by every version of {\sc Olivaw} (the initial matches are warm-up games used by the platform to assess the strength of the player and thus are excluded).  Every agent played anonymously using 400 MCTS simulations per move.}}
\label{fig_othelloquest}
\end{figure*}

\section{The training process}
\label{sec-training_process}

The version of {\sc Olivaw} discussed in this article is the result of a single training run lasting 30 days, 20 generations, and $\sim50,000$ games. We refer to the \emph{i-th generation} as the \emph{i-th successful update of the weights of $f_\theta$}.

Fine-tuning the hyperparameters for $8 \times 8$ Othello would have required a number runs incompatible with our main objective of mastering the game with limited resources.
Similarly, ablation studies to determine the effectiveness of our choices to improve the learning phase would have been prohibitively costly. As discussed however, these choices find good motivation in previous work, such as \citet{wang2019hyper} and \citet{connect4MediumLessonsAlphaZero}.

During the training phase, several interesting trends emerged (please refer to Figure~\ref{fig:trends}). Figure~\ref{fig:trends}~\textsc{a} plots the progress of the loss function across generations. Noticeable jumps take place when {\sc Olivaw} switches from one generation to the next. Recall that a training set consists of a batch of labeled games played by the most recent generations (the last two for the first generations and the last five later on). When the current oracle is defeated by a challenger, a new generation is born. The training set is updated by replacing the games played by the oldest generation with those of the most recent ones. The sudden drop in the loss can be ascribed to the improved quality of the new data set. This is an indication that {\sc Olivaw} is learning as generations go by. Other indications are given by the remaining two plots in Figure~\ref{fig:trends}. The plot in the middle simply reports the ELO rating of {\sc Olivaw} as generations go by \rev{and the number of games played by each generation against itself}. The rightmost plot shows the evolution of an interesting metric. Recall that the oracle provides two different types of evaluation given a state of the game: a probability distribution over the possible moves from that state and an assessment, ranging from $-1$ (certain defeat) to $+1$ (certain victory), of how good the current state is for {\sc Olivaw}. If the oracle is good, we expect the latter to change little from one move to the next. 
Conversely, a big drop from one state of play to the next is an indication that the oracle was unable to predict a very unfavorable move. The plot reports the maximum drop observed during a generation. It can be seen that the oracle is also improving according to this measure.

\begin{figure}[h]
  \begin{overpic}
		[trim=0cm 0cm 0.15cm 0cm,clip,width=0.99\linewidth]{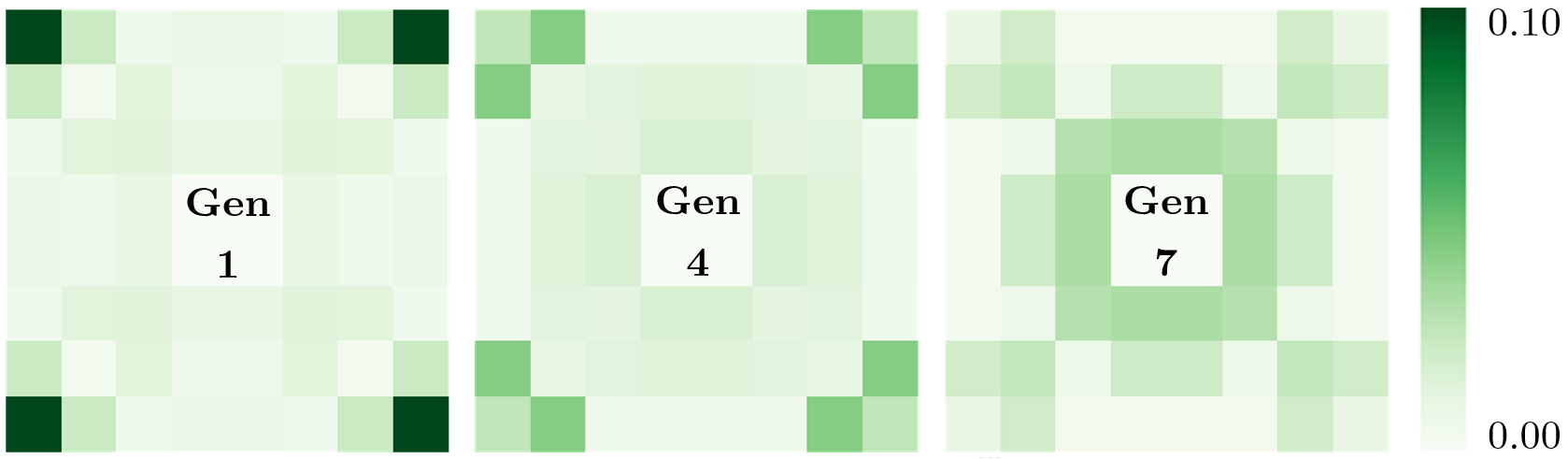}
		\end{overpic}
\caption[Location of crucial moves by generation]{\textsc{Location of crucial moves by generation.} {\sc Olivaw} attributes high relevance to the conquest of the corners in the early generations, similarly to human beginners. In later generations, it shifts its ``attention'' towards the center of the board, as we would expect from a more experienced player.}
\label{fig:crucial}
\end{figure}

Another interesting fact is reported in Figure~\ref{fig:crucial}. It shows that, similarly to human beginners, early generations correctly but naively attach much significance to conquering the corners, gradually improving their appreciation of less conspicuous but more strategic positions in the middle of the board.

\subsection{Details to replicate training conditions:}
In the following, we report all salient hyperparameters to reproduce this work.

In each generation {\sc Olivaw} plays $\sim2500$ games against itself ($25,000$ in AlphaGo Zero), \rev{see Figure \ref{fig:trends}b}. As explained in Sec. \ref{sec-comparison}, each move is selected after 100, 200, or 400 MCTS simulations.
During the \emph{self-play phase}, the first 20 moves of each game are selected extracting at random according to $\bm{\pi}$ to favor exploration, the remaining moves are selected taking $\text{argmax}(\bm{\pi})$. As in AlphaGo Zero we use virtual losses \cite{segal2010scalability}.

When selecting a move from state $s$ with MCTS, in the root node we add Dirichlet noise to the prior probabilities computed by the neural network oracle:
\begin{equation}
    P(s_0, a) = (1 - \epsilon)p_a + \epsilon x_a \;\;\;\; \bm{x} \in \mathbb{R}^B
\end{equation}
Where $B$ is the number of legal moves from $s_0$, and
$\bm{x}$ is a point sampled from the symmetric Dirichlet probability distribution $X_\alpha$, with $\alpha=\mathrm{min}(1, 10 / B)$, we used an $\epsilon = 0.25$ as in AlphaGo Zero. 
A symmetric Dirichlet noise with an $\alpha < 1$ tends to unbalance the move probability distribution $\pi(a|s_0)$ towards a specific action, favoring an exploration in depth of the subsequent variant. In games with high branching factor, this behavior is desired to preserve the asymmetric nature of the MCTS search. So, the higher the branching factor of the game, the lower the $\alpha$ parameter of the symmetric Dirichlet noise used ($\alpha = 0.03$ in AlphaGo Zero).

When generating the dataset, not all games are played until the end to save computation. If during a game a player values a position under a resignation threshold $v_{\mathrm{resign}}$, the game ends and the opponent is considered the winner. $v_{\mathrm{resign}}$ is chosen automatically playing a fraction of the games until the end so that less than $5\%$ of those games could have been won if the player had not resigned. In early generations {\sc Olivaw} plays $10\%$ of the games until the end, \rev{i.e. the first 250 games. We increased} progressively this fraction to improve its strength in finals. This is because, differently from Go, Othello games are usually played until the very end, when no more moves are available.
(AlphaGo Zero plays always only $10\%$ of the games until the end).

In the \emph{training phase}, neural network parameters $\theta$ are optimized using stochastic gradient descent with momentum $\lambda=0.9$ and a stepwise learning rate annealing. Starting from a learning rate of 0.003, {\sc Olivaw} switched to 0.001 in the 4th generation and to 0.0001 in the 11th generation. 

Concerning the architecture, {\sc Olivaw} uses a Residual Network \cite{He2015DeepRecognition} as AlphaGo Zero. The game state input $s$ is a $8 \times 8 \times 2$ binary tensor in {\sc Olivaw}, in contrast to the deeper $19\times 19 \times 17$ binary tensor of AlphaGo Zero. We do not need layers representing past positions in Othello since the game state is fully observable from the board position. Even the information on the player's turn is not necessary since Othello is symmetrical with respect to the player, we can assume that the turn player is always white, flipping all the discs if it is the turn of black.

Therefore the input $s$ is processed by a single convolutional block and then by a residual tower of 10 residual blocks (39 in the strongest version of AlphaGo Zero). The output of the residual tower is then processed by the value head and the policy head that output respectively the value $v$ of the position and the move probabilities $\bm{p}$. The structure of each block coincides with the correspondent one of AlphaGo Zero, except for the output shape of the policy head, $8^2 + 1=65$ in {\sc Olivaw} instead of $19^2 + 1 = 362$.

\begin{figure}[t]
\includegraphics[trim=0cm -1.8cm 0cm 0cm,clip,width=0.42\textwidth]{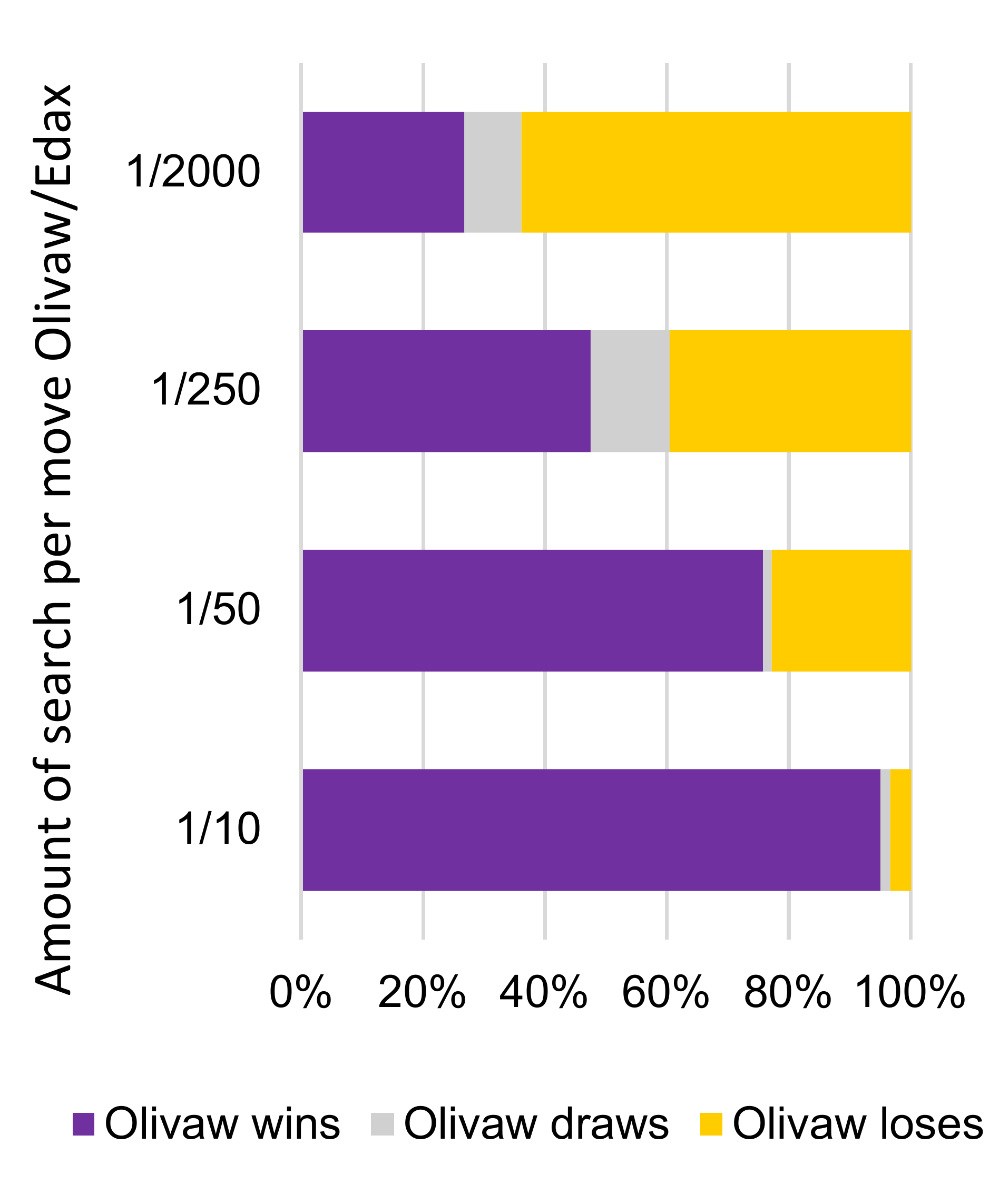}
\includegraphics[width=0.54\textwidth]{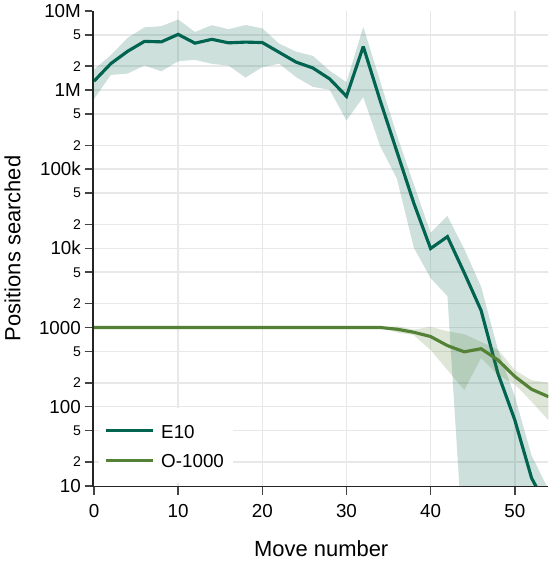}
\hspace{-0.6cm}
\centering
\begin{overpic}
		[trim=10cm 34cm 2cm 0cm,clip,width=1.0\linewidth]{images-olivaw/crucial_moves.png}
		\put(22,2){ \textsc{a}  }
		\put(75,2){ \textsc{b}  }
\end{overpic}
\caption[{\sc Olivaw} vs Edax]{\textsc{Olivaw vs Edax.} \textsc{a.} 
Outcome of four $100$-game series between O-$1000$ and E$4$, E$6$, E$8$, and E$10$. 
\textsc{b.} Number of game-tree positions  searched by E$10$ and O-$1000$ for each move, averaged over 10 games.}
\label{fig_edax}
\end{figure}

\section{Attaining world-class level in Othello}
\rev{
We tested the strength of {\sc Olivaw} in three different ways. First, 
by playing anonymous games on the web platform OthelloQuest. Second,  by pitting it against Edax, one of the strongest open-source Othello engine.  And finally, with two matches against top-notch human players: a national champion and a former world champion.}
\rev{
\subsection{Matches on the web platform OthelloQuest}
}
\rev{During training, the strength of {\sc Olivaw} was tested with a series of anonymous online games against human players on OthelloQuest, a popular Othello platform which is also widely used by top human players. {\sc Olivaw} was deployed at generation 8,12,16, and 20 with the number of explorable nodes in the game tree set at 400. The duration of every game was set to 5 minutes. In Figure \ref{fig_othelloquest} we report the score of the last 50 games played by each version of {\sc Olivaw}. We excluded the positioning games used by the platform to assess the level of a new player, which is why we do not report the games between {\sc Olivaw} 20 and opponents ranked at 1400. A clear improvement can be observed as generations progress. Performance ratings
\footnote{This rating is estimated from the games of a single event only, see \url{https://en.wikipedia.org/wiki/Elo_rating_system}.}
after these 50 games of generations 8, 12, 16, and 20 are, respectively, 1557, 1687, 2074, and 2100.}
\rev{
\subsection{Matches against Edax}}
We tested {\sc Olivaw} against Edax \cite{Edax}, arguably the strongest open-source Othello engine \cite{liskowski2018learning}. Like other top traditional engines, Edax is based on a highly optimized alpha-beta tree search (negamax) \cite{knuth1975analysis} using tabular value functions\footnote{Unfortunately, a detailed description of how these value functions are obtained is not publicly
available, \rev{see \cite[Section 2]{liskowski2018learning} for further details.}}. 
\rev{In what follows E$k$ denotes the version of Edax in which the depth of the alpha-beta search in the game tree is limited to $k$. In our comparison, E$4$, E$6$, E$8$ and E$10$ were used.}
For the games we report, Edax used no opening book. Edax is a deterministic engine and when it played against {\sc Olivaw} the same single game was repeated again and again. To circumvent this problem we switched to random XOT openings, a popular Othello variation where the first 8 moves are chosen at random from a list of 10,784 sequences ending in an almost even position, i.e. positions judged between -2 and +2 discs advantage for black by Edax at search depth 16. 

\rev{The four versions of Edax were pitted against four versions of the 20th generation of {\sc Olivaw}. The four versions deployed correspond to the maximum number of nodes of the game tree allowed to be explored and that were set to $400, 1000, 2500,$ and $10$ thousand. The resulting versions of {\sc Olivaw} are referred to as O-$400$, O-$1000$, O-$2500$, and O-$10$T.}

\rev{Several aspects must be considered in order to set up a comparison between Edax and {\sc Olivaw} that is informative as well as feasible. As far as the former aspect is concerned, setting limits in terms of wall-clock time for the two agents would not be very informative. They use different hardware (GPUs vs CPUs) and are written in different programming languages. This is why we opted for a machine-independent measure of computational effort: the number of explored nodes in the game tree to decide the next move. (In the case of Edax the depth of the alpha-beta search translates into number of nodes explored, see below). 
As for the latter, it seems reasonable to assume that, while increasing the budget of explorable nodes makes an agent stronger, sooner or later a plateau must be reached whereby increasing the budget does not translate in more strength. In order to compare the best of the two agents therefore, it is tempting to determine such a plateau and have the two resulting top versions play each other. This approach however is completely unfeasible in terms of resources, especially in our constrained framework.}

\begin{figure}[t]
\includegraphics[width=0.99\textwidth]{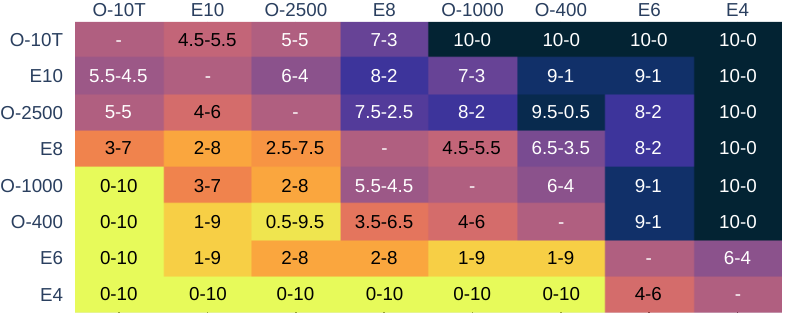}
\centering
\caption[{\sc Olivaw} vs Edax: the tournament]{\rev{\textsc{{Olivaw} vs Edax: The Tournament.}  
Outcome of a tournament among the {\sc Olivaw}-type agents (O-$400$, O-$1000$, O-$2500$, and O-$10$T) and the Edax-type agents (E$4$, E$6$, E$8$, and E$10$), whereby every agent plays a $10$-game series against every other agent. Darker (resp. lighter) colours indicate a favourable (resp. unfavourable) outcome for the agents listed on the left column.}
}
\label{fig-tournament}
\end{figure}

\rev{Thus, we settled for the following: a tournament in which the eight agents played against each other. A match between two agents consists of a $10$-game series. For every game, we assigned 1 point for a victory, $0.5$ points for a draw, and $0$ points for a defeat.
A widespread assessment among accomplished Othello players is that these four versions of Edax have a strength that gradually increases from good (E$4$) to exceptionally strong (E$10$) (to the level of the best human players, if not stronger), thus providing a good benchmark to assess {\sc Olivaw}'s strength. The outcome of the tournament is reported in Figure~\ref{fig-tournament} (matches) and Table~\ref{tab-tournament} (leaderboard).}

\rev{Conforming to intuition,  it is apparent that as the exploration budget increases all agents become stronger.} 
\rev{The main take-away point however is that 
{\sc Olivaw} is competitive against Edax in spite of the fact that it explores far fewer nodes in the game tree than its opponent. On average, E$4$, E$6$, E$8$ and E$10$ explored $10,000$, $50,000$, $250,000$, and $2,000,000$ nodes to make a move, respectively, whereas we recall that the four versions of {\sc Olivaw} explore at most $400, 1000, 2500,$ and $10,000$ nodes per move.
This overall conclusion is further exemplified by Figure~\ref{fig_edax}a, which shows the outcome of four $100$-game series between O-$1000$ versus E$4$, E$6$, E$8$ and E$10$.
}

\rev{As Table~\ref{tab-tournament} shows, O-$10$T won the tournament, even if it lost the series against E$10$ by a narrow margin. This is because, interestingly, weak versions of {\sc Olivaw} still managed to occasionally beat the strongest version of Edax.}

\rev{It is perhaps worth noting that, unlike {\sc Olivaw}, Edax can take advantage of heuristics fine-tuned by humans to choose when to expend more search. Figure \ref{fig_edax}b shows the average number of positions searched  during the series between  E$10$ and O-$1000$.  Besides the huge difference in game-tree exploration effort, notice the peak 20 moves before the end. It is when Edax is programmed to search deeper in preparation for the endgame.}

\rev{Overall, the results reported in this section indicate that {\sc Olivaw} is a very strong player, possibly at the level of top human players. We test this hypothesis in the next section.}
\begin{table}[]
\centering
\begin{tabular}{ccc}
\hline
\textsc{Points} & \textsc{Player} & \textsc{$\max($positions searched per move$)$} \\ \hline
56,5            & O-10T      & $1.0 \times10^4$                                                                     \\
54,5            & E10         & $(8.3 \pm 3.2) \times 10^6$                                                          \\
52              & O-2500     & $2.5 \times 10^3$                                                                    \\
36,5            & E8 & $(1.1 \pm 0.3) \times 10^6$                                                          \\
35,5            & O-1000     & $1.0 \times 10^3$                                                                    \\
28              & O-400      & $4.0 \times 10^2$                                                                    \\
13              & E6        & $(1.0 \pm 0.2) \times 10^5$                                                          \\
4               & E4        & $(1.7 \pm 0.2) \times 10^4$                                                          \\ \hline
\end{tabular}
\caption[Leaderboard of the tournament between {\sc Olivaw} and Edax]{\rev{Leaderboard of the tournament between Olivaw and Edax}}
\label{tab-tournament}
\end{table}

\subsection{Matches against top-notch human players}
As a final, and much more enjoyable, battery of tests, we organized three live series against top human players with the support of the Italian Othello Federation. Two were against the 2019 Italian champion Alessandro Di Mattei, ranked among the top 150 Othello players in the world, and one, more formal challenge, against the former World champion Michele Borassi, ranked in the top 50 and who finished in the top five in his last World Othello championship appearance in 2018\footnote{World Othello ratings at the time of matches.}.

\textsc{Matches against a national champion.} Two informal best-of-five series were organized against the Italian champion Alessandro Di Mattei. They took place between 27 November and 4 December 2018 in Rome.
The first series against Di Mattei saw the 14th generation of {\sc Olivaw} very close to the national champion, who emerged victorious after two initial draws. A post-match analysis of the games showed {\sc Olivaw} losing from winning positions in the final moves. This feedback led to the decision of simulating all the subsequent \emph{self-play games} until the very end, to strengthen {\sc Olivaw}'s late game. 
The second series against generation 18 ended with a resounding 4-0 victory for {\sc Olivaw}. This boosted our confidence and we threw down the gauntlet against a former world champion. Table \ref{tab-games} shows the matches against Di Mattei.

\textsc{Challenging a former World champion.}
Former world champion Michele Borassi picked up the gauntlet and a formal series was organized. The match was sponsored by the Sapienza Computer Science department and was open to the public \rev{and streamed live over the internet}. 
The formula was a best-of-three with 30 minutes for each player for the whole game, as in the world championship. After the good results against Di Mattei, we decided to keep the MCTS simulations of the game tree at 1000 -\rev{a very small amount of search per move, insufficient to defeat even beginners for traditional programs like Edax}- and to stop the training at generation 20, after $\sim50,000$ games simulated in \emph{self-play}. 
\rev{In short, Borassi played against O-$1000$}.

{\sc Olivaw} won the first game as black, losing the second and third one with white. The final score was thus 2-1 against {\sc Olivaw}. All matches are shown in Table \ref{tab-games}. 

\begin{figure}[h]
\includegraphics[trim=1.9cm 0.5cm 0cm 0.cm,clip,width=0.94\textwidth]{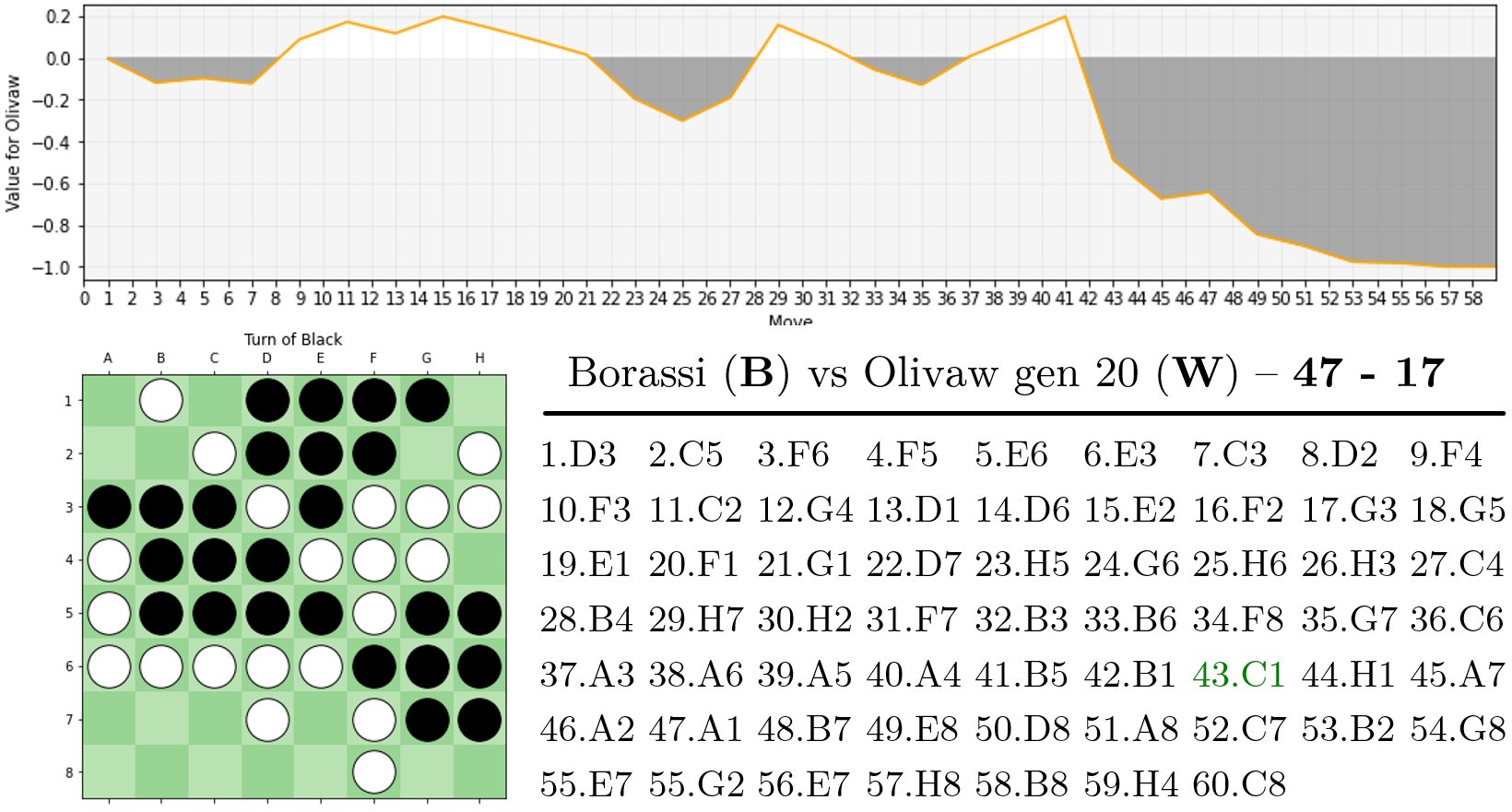}
\centering
\caption[Match between {\sc Olivaw} generation 20 and the former World champion Michele Borassi, final game]{\textsc{Match between {\sc Olivaw} generation 20 and the former World champion Michele Borassi, final game.} The graph shows the confidence of winning of {\sc Olivaw}; -1 a sure defeat, +1 a sure victory.}
\label{fig-Borassi}
\end{figure}

Othello players may find interesting move 43. C1 of Borassi in game~3. It is a highly counter-intuitive move and the only option to avoid defeat: with a masterstroke the former world champion snatched victory from the jaws of defeat. Borassi spent more than one-third of its time on that single move. {\sc Olivaw} did not see it coming, as evidenced by the large value drop recorded on move~43 (see Figure \ref{fig-Borassi}). 
An act of \rev{digital hubris} that proved fatal.

\begin{table*}[t]

\setlength{\tabcolsep}{6pt}
\centering
\resizebox{\linewidth}{!}{\begin{tabular}{ll|c|l}
\multicolumn{4}{c}{First match against the national champion Alessandro Di Mattei - Best of 5 - 2018/11/27} \\
\hline
Di Mattei & {\sc Olivaw} 14 & 32-32 & {\tiny C4E3F6E6F5C5C3C6D3D2E2B3B4C2B6A4B5D6A3A5A6F3F4G4F7D1F1D7E1C1B1G6C7E7F8D8H6F2G1G5C8B8G7B7E8G2A8A7H1G3H2H3H4B2A2A1PAH5PAH8G8H7 \par}  \\ 
{\sc Olivaw} 14 & Di Mattei & 32-32 & {\tiny D3C3C4C5B4D2D6C6E6D7B5A3C7C8B6A6E3E7A4F2F8F5F6F4A2E8F7G5G6H6C2B7F3A5A8A1E2B3D8B8B2G4G3B1C1H3F1E1D1G7H4H5A7G1H8G8H7G2H2PAH1 \par}  \\ 
Di Mattei & {\sc Olivaw} 14 & 43-21 & {\tiny C4E3F6E6F5C5C3C6D3D2E2B3C1C2B4A3A5B5A6F4F3B6E7D1E1G5G4G6H5D6B1H3H4H6F7F8D8A4A2E8G8G3D7C8B8G7H8H7A7C7H2G2G1B7A8F1F2H1PAA1PAB2 \par}  \\ 
{\sc Olivaw} 14 & Di Mattei & 36-28 & {\tiny F5F6E6D6E7G5C5C6C4F3D7D8C7C8F4B6G6G4H5H6H4H3G3B5E3B3B4C3A5A6D3C2D2D1A3H2C1B1F7A4A7B2E2E8G8G7F8H8H7G2F2B7A8B8H1F1G1E1A1A2 \par}  \\ 
Di Mattei & {\sc Olivaw} 14 & 33-31 & {\tiny C4E3F6E6F5C5F4G6F7C3H6G4G3D7E7F3F2H3D3E2E1C6D6G5D2C7C8C2B1E8B8F1G1G8H5H4B7B5A5B4F8B6D8A8H2C1D1H7H8G7B2B3A3A4A7A6PAA2PAG2H1PAA1 \par}  \\ 
\hline
\hline
\noalign{\vskip 1mm}    
\multicolumn{4}{c}{Second match against the national champion Alessandro Di Mattei - Best of 7 - 2018/12/04} \\
\hline
Di Mattei & {\sc Olivaw} 18 & 27-37 & {\tiny C4E3F6E6F5C5C3C6D3D2E2B3B4A3E7C2D6F1D1F4E1C1B6F3B2D7C8G5H4C7D8G6H6H5G4H3H2B7B1B5A8A1A5E8F7G7A4F8H7A6A2H8G8H1G3F2A7B8G1G2 \par}  \\ 
{\sc Olivaw} 18 & Di Mattei & 40-24 & {\tiny E6F6F5D6E7G5C5C6E3C4D7F8B4D3C3A3B5B3B6C8A4A5A6A7F4C7G6H6F7G8H4H5H7C2D2F2F3D1E2G2E1C1B1G4F1B2A1A2A8G3E8D8B8B7H8G7H1H2H3G1 \par}  \\ 
Di Mattei & {\sc Olivaw} 18 & 27-37 & {\tiny C4E3F6E6F5C5C3C6D3D2E2B3C1C2B4A3A5B5A6B6A4A7E7E1D6D7C8F3C7F8F7G5H4G6H5D1F1F2B1H6H7G4H3F4G3E8B7D8G8B2B8G7A2A1PAG1G2H2H1 \par}  \\ 
{\sc Olivaw} 18 & Di Mattei & 45-19 & {\tiny C4E3F6E6F5C5C3B4D3C2D6F4E2F3D2C6G5G4F2G3H3H4B5H6A3B6D7A4A5A6B3C1E1A2D1F1G2E7E8F7H2C8F8D8C7G8B7G6H5H1G1B8G7A8A7B2A1B1PAH8H7 \par}  \\ 
\hline
\hline 
\noalign{\vskip 1mm}    
\multicolumn{4}{c}{Match against the former World champion Michele Borassi - Best of 3 - 2019/01/19} \\
\hline
{\sc Olivaw} 20 & Borassi & 35-29 & {\tiny E6F4C3C4D3D6F6E7F3C5F5G4B5G5C6E3D7B4G6F7E2C8B6C7A5H5E8F2B3F8D8E1H6A4A3C2C1A7H3H4A6A2D2D1G3G2H1H2B2B8A8B7A1B1G8G7H8H7F1G1 \par}  \\ 
Borassi & {\sc Olivaw} 20 & 35-29 & {\tiny D3C5F6F5E6E3C3D2F4F3C2G4D1D6E2F2G3H4G1F7H3G6C4H2G5B4H5H6E7B3C7C6D7D8B5A5E8F8A6C8A4F1E1B1B6A2C1H1A3A7G7G8H8H7B7B2A1A8B8G2 \par}  \\ 
Borassi & {\sc Olivaw} 20 & 47-17 & {\tiny D3C5F6F5E6E3C3D2F4F3C2G4D1D6E2F2G3G5E1F1G1D7H5G6H6H3C4B4H7H2F7B3B6F8G7C6A3A6A5A4B5B1C1H1A7A2A1B7E8D8A8C7B2G8G2E7H8B8H4C8 \par}  \\ 
\hline
\hline 
\end{tabular}
}

\caption[{\sc Olivaw} games versus top Othello players]{ {\sc Olivaw} games versus top Othello players.}
\label{tab-games}
\end{table*}

\section{Conclusion}
\rev{After one month of training, using only free, and quite limited, cloud computing resources, {\sc Olivaw} achieved world-class level in the game of Othello.  The high ELO rating reached on the popular web platform OthelloQuest, the winning challenges against the strongest open-source Othello engine Edax, the victory against national champion Alessandro Di Mattei, and the honorable defeat against former World champion Michele Borassi corroborate this assessment.}

\rev{Differently from traditional Othello engines like Edax, {\sc Olivaw} reached these results using a minimal amount of search per move,
and learned its value function completely from scratch, like AlphaGo Zero.
The amount of training required was $\sim50,000$ games played against itself. Differently from its illustrious predecessor, {\sc Olivaw} maximizes the information extracted from each game by adding to the training set also positions not played but largely explored by the agent.}

The level of play reached by {\sc Olivaw} is not yet superhuman however. This would be the natural next step for {\sc Olivaw} and is left for future work. To be interesting, such an achievement should come within the same resource-limited paradigm, using a limited amount of computational time and power in the training phase and a small number of MCTS simulations per move during  matches against a human opponent.

Also, {\sc Olivaw}'s discovered Othello knowledge remains trapped in the weights of its neural network, and is only able to produce board evaluations in the form of a win probability, as any classical engine based on human knowledge.
This is a fundamental limitation of the AlphaGo Zero paradigm as a model for a scientist; in the following chapter we will discuss how to develop agents that are at the same time able to discover knowledge and communicate it.

%% file: Chapters/Chapter03.tex
\def\I{$\mathcal{I}$}
\def\CG{$\mathcal{CG}$}
\newcommand{\SoftRes}[2]{#1}
\newcommand{\CRN}{CRN}
\newcommand{\CRNs}{CRNs}

\newcommand{\Rationalist}{\CRN{}}
\newcommand{\EmpA}{Empiricist-conscious}
\newcommand{\EmpR}{Empiricist rule-only}
\newcommand{\EmpL}{Empiricist label-only}

\newcommand{\RAT}{\CRN{}}
\newcommand{\EA}{Emp-C}
\newcommand{\ER}{Emp-R}
\newcommand{\EL}{Emp-L}

\chapter{Explanatory Learning: Beyond Empiricism in Neural Networks}\label{ch:explanatory} 
\section*{Chapter abstract}
At the crossroads of Program Synthesis and Meta-Learning, we introduce Explanatory Learning as the task of automatically discovering the symbolic explanation (PS) that enables few-shot sensible predictions on a novel environment given experience on other environments (M-L). Differently from PS, the program (explanation) interpreter in EL is not given and should be learned from a limited collection of associations \emph{explanation}-\emph{observations}. Unlike M-L, EL does not prescribe any adaptation at test time, seeking generalization in the broad meanings attributed to symbols by the learned interpreter.
To exemplify the challenges of EL, we present the Odeen benchmark, which can also serve the PS and M-L paradigms. Finally, we introduce Critical Rationalist Networks, a deep learning approach to EL aligned with the rationalist view of knowledge acquisition.
CRNs express several desired properties by construction; they are truly explainable, can adjust their processing at test-time for harder inferences, and can offer strong confidence guarantees on their predictions.
Using Odeen as a testbed, we show how CRNs outperform empiricist end-to-end approaches of similar size and architecture (Transformers) in discovering explanations for unseen environments. 

\blfootnote{This chapter is based on the paper \textit{"Explanatory Learning: Beyond Empiricism in Neural Networks"}, by Antonio Norelli, Giorgio Mariani, Andrea Santilli, Luca Moschella, Giambattista Parascandolo, Simone Melzi, and Emanuele Rodolà.}

\newpage

\section{Introduction}

Making accurate predictions about  the future is a key ability to survive and thrive in a habitat.
Living beings have evolved many systems to this end, such as memory \cite{mcconnell1962memory}, and several can predict the course of complex phenomena \cite{Taylor16389}. However, no animal comes even close to the prediction ability of humans, which stems from a unique-in-nature system.

At the core of this system lies an object called \emph{explanation}, formed by 
the proposition of a language, which has a remarkable property: 
it can be installed with ease into another human speaking the same language, allowing to make predictions on new phenomena without ever having experienced them. 
When the installation is successful, 
we say that the human has \emph{understood} the explanation. 
\begin{figure}[h]
        \centering
        \begin{overpic}
		[trim=0cm 0cm 0cm 0.0cm,clip,width=0.999\linewidth]{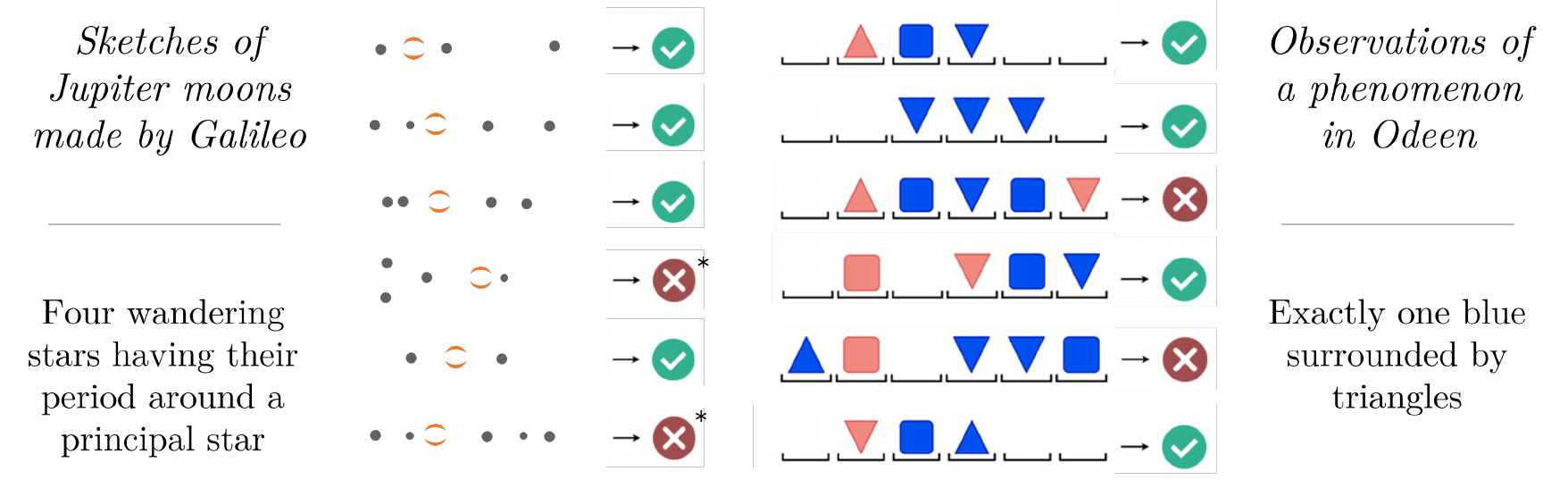}
		\put(3.0,13){\scriptsize \textsc{Explanation:$^1$}}
		\put(82,13){\scriptsize \textsc{Explanation:}}
		\end{overpic}
                        
        \caption[The Odeen universe]{\textsc{The Odeen universe}. A convenient setting to study and test the process of knowledge discovery in machines. Like the night sky was for humans. $\;\;\;\;\;\;\;\;\;\;\;\;\;\;\;\;\;\;\;\;\;\;\;\;\;\;\;\;\;\;\;\;\;\;\;\;\,${\tiny $^1$Galileo did not sketch negative examples.}}
        \label{fig-galileo}
\end{figure}

This process is key to the success of human beings. An individual 
can provide accurate predictions for a multitude of phenomena without going through a painful discovery process for all of them, but only needs an operating system -- mastering a language -- and someone who communicates the relevant explanations;
this way, the individual can focus on unexplained phenomena. When an explanation is found for them, it is added to the existing shared collection, which we call \emph{knowledge}.

How can we make machines take part in this orchestra?
With this work, we try to shed new light on this problem. Specifically, we propose a learning procedure to allow machines (i) to {\em understand} existing explanations, in the sense described above, and (ii) create new explanations for unexplained phenomena, much like human scientists do.
\footnotetext[1]{The explanation \textit{Four wandering stars having their period around a principal star} is adapted from the English translation of the Sidereus Nuncius \citep[page 9]{galilei1610sidereus}. First sketch on the left is compatible with the rule since the fourth moon can be hidden by one of the other moons or by Jupyter itself.}
\newpage
Our contribution in this sense is threefold:

\begin{itemize}
\item We formulate the challenge of creating a machine that masters a language as the problem of learning an interpreter from a collection of examples in the form \emph{explanation}-\emph{observations}. The only assumption we make is this dual structure of data; explanations are free strings, and are not required to fit any formal grammar. This results in the \emph{Explanatory Learning} (EL) framework described in Sec.~\ref{sec-el}.

\item We present Odeen, a basic environment to test EL approaches, which draws inspiration from the board game Zendo \cite{zendo}.
        Odeen simulates the work of a scientist in a small universe of simple geometric figures, see Figure \ref{fig-galileo}. 
    We present it in Sec.~\ref{sec-odeen}, and will release it upon publication.
    
\item We argue that the dominating empiricist ML approaches are not suitable for EL problems. We propose \emph{Critical Rationalist Networks} (CRNs), a family of models designed according to the epistemological philosophy pushed forward by  \citet{popper1935logic}.
    Although a CRN is implemented using two neural networks,
    the working hypothesis of such a model does not coincide with the adjustable network parameters, but rather with a language proposition that can only be accepted or refused \emph{in toto}. 
    We will present CRNs in Sec.~\ref{sec-crns}, and test their performance on Odeen in Sec.~\ref{sec-exp}. \end{itemize}
    \section{Explanatory Learning}\label{sec-el}

Humans do not master a language from birth. A baby can not use the message ``this soap stings'' to predict the burning sensation caused by contact with the substance.
Instead, the baby gradually \emph{learns} to interpret such messages and make predictions for an entire universe of phenomena \cite{schulz2007preschool}. We refer to this state of affairs as \emph{mastering a language}, and we aim to 
replicate it in a machine as the result of
an analogous learning process.

Using a batch of explanations paired with observations of several phenomena, we want to learn an interpreter to make predictions about novel phenomena for which we are given explanations in the same language. Going a step further, we also want to discover these explanations, when all we have is a handful of observations of the novel phenomena. 
We first describe the problem setup in the sequel, comparing it to existing ML problems; then we detail our approach in Sec.~\ref{sec-crns}.

\subsection{Problem setup}
Formally, let phenomena $P_1, P_2, P_3, \dots$ be subsets of a universe $U$, which is a large set with no special structure (i.e., all the possible observations $U = \{x_1, \dots, x_z\}$). 
Over a universe $U$, one can define a language $L$ as a pair $(\Sigma_L, \mathcal{I}_L)$, where $\Sigma_L$ is a finite collection of short strings
over some alphabet 
$A$, with  $|\Sigma_L| \gg |A|$,
and $\mathcal{I}_L$ is a binary function $\mathcal{I}_L: U \times \Sigma_L \rightarrow \{ 0, 1 \}$, which we call interpreter.
\graffito{Explainability definition}
We say that a phenomenon $P_i$ is \textsc{\emph{explainable}} in a language $L$ if there exists a string $e \in \Sigma_{L}$ such that, for any $x \in U$, it occurs $\mathcal{I}_{L}(x, e) = \mathbf{1}_{P_i}(x)$, where $\mathbf{1}_{P_i}(x)$ is the indicator function of $P_i$. 
We call the string $e$ an explanation, in the language $L$, for the phenomenon $P_i$. 

Our first contribution is the introduction of a new class of machine learning problems, which we refer to as \emph{Explanatory Learning} (EL).

Consider the general problem of making a new prediction for a phenomenon $P_0\subset U$. In our setting, this is phrased as a binary classification task: given a sample $x' \in U$, establish whether $x' \in P_0$ or not. We are interested in two instances of this problem, with different underlying assumptions:

\begin{itemize}
\item \textsc{The communication problem: we have an explanation}. We are given an explanation $e_0$ for $P_0$, in an unknown language $L$. This means that we do not have access to an interpreter $\mathcal{I}_L$;     $e_0$ looks like Japanese to a non-Japanese speaker. Instead, we are also given other explanations $\{e_1, \dots, e_n\}$, in the same language, for other phenomena $P_1, \dots, P_n$, 
        as well as observations of them, i.e., datasets $\{D_1, \dots, D_n\}$ in the form $D_i = \{(x_1, \mathbf{1}_{P_i}(x_1)), \dots, (x_m, \mathbf{1}_{P_i}(x_m))\}$, with $m \ll |U|$. Intuitively, here we expect the learner to use the explanations paired with the observations to build an approximated interpreter $\mathcal{\hat{I}}_L$, and use it to make the proper prediction for $x'$ by evaluating $\mathcal{\hat{I}}_L(x', e_0)$.

\item \textsc{The scientist problem: we do not have an explanation}. We are given explanations $\{e_1, \dots, e_n\}$ in an unknown language $L$ for other phenomena $P_1, \dots, P_n$ and observations of them $\{D_1, \dots, D_n\}$. However, we do not have an explanation for  $P_0$; instead, we are given just a small set of observations $D_0 = \{(x_1, \mathbf{1}_{P_0}(x_1)), \dots, (x_k, \mathbf{1}_{P_0}(x_k))\}$
    and two guarantees,
            \graffito{Representativity definition}
    namely that $P_0$ is explainable in $L$, and that $D_0$ is \textit{{representative}} for $P_0$ in $L$. That is, for every phenomenon $P \neq P_0$ explainable in $L$ there should exist at least a $x_i\in D_0$ such that $\mathbf{1}_{P_0}(x_i) \neq \mathbf{1}_{P}(x_i)$. 
     Again, we expect the learner to build the interpreter $\mathcal{\hat{I}}_L$, which should first guide the search for the missing explanation $e_0$ based on the clues $D_0$, and then provide the final prediction through $\mathcal{\hat{I}}_L(x', e_0)$.

\end{itemize}

Several existing works fall within the formalization above. 
The seminal work of \cite{Angluin} on learning regular sets is an instance of the scientist problem, where finite automata take the role of explanations, while regular sets are the phenomena. 
More recently, CLEVR~\citep{clevr} posed a communication problem in a universe of images of simple solids, where explanations are textual and read like \emph{``There is a sphere with the same size as the metal cube''}. Another recent example is CLIP \cite{clip}, where 400,000,000 captioned internet images  are arranged in a communication problem to train an interpreter, thereby elevating captions to the status of explanations rather than treating them as simple labels\footnote{This shift greatly improved the performance of their model, as discussed in \citep[Sec.~2.3]{clip}.}.
With EL, we aim to offer a unified perspective on these works, making explicit the core problem of learning an interpreter purely from observations.

\subsection{Relationship with other ML problems.} We briefly discuss the relationship between EL and other problems in ML, pointing to Sec.~\ref{sec:related} for additional discussion on the related work.

EL can be framed in the general meta-learning framework. The learner gains experience over multiple tasks to improve its general learning algorithm, thus requiring fewer data and less computation on new tasks. However, differently from current meta-learning approaches \citep{finn2017model, Lee_2019_CVPR}, we are not optimizing for any meta-objective \citep{hospedales}. Instead, we expect the sought generality to be a consequence of implicitly defining an interpreter through a limited set of examples rather than an explicit goal to optimize for.

To many, the concept of explanation may sound close to the concept of program; similarly, the scientist problem may seem a rephrasing of the fundamental problem of {Inductive Logic Programming} (ILP) \citep{shapiro1981inductive} or {Program Synthesis} (PS) \citep{balog2019deepcoder}. This is not the case. ILP has the analogous goal of producing a hypothesis from positive/negative examples accompanied by background knowledge. Yet, ILP requires observations to be expressed as logic formulas, a task requiring a human; only then the ILP solver outputs an explanation in the form of a logic proposition, which in turn is interpreted by a human expert. With EL, data can be fed as-is without being translated into logic propositions, and a learned interpreter plays the expert's role. PS also admits raw data as input, it yields a program as output, and replaces the expert with a handcrafted interpreter; still, the sequence of symbols produced by a PS system only makes sense to a human (who designed the interpreter), not to the system itself. Instead, in EL, the interpreter is learned from data rather than hardcoded. An empirical comparison demonstrating the benefits of EL over PS is given in Sec.~\ref{sec-exp}.

Next we introduce Odeen, an environment and benchmark to experiment with the EL paradigm. 

\section{Odeen: a puzzle game as Explanatory Learning environment}\label{sec-odeen}

\setlength{\columnsep}{3pt}

\begin{figure}
\centering
  \begin{overpic}
		[trim=0cm 0cm 0cm 0cm,clip,width=0.82\linewidth]{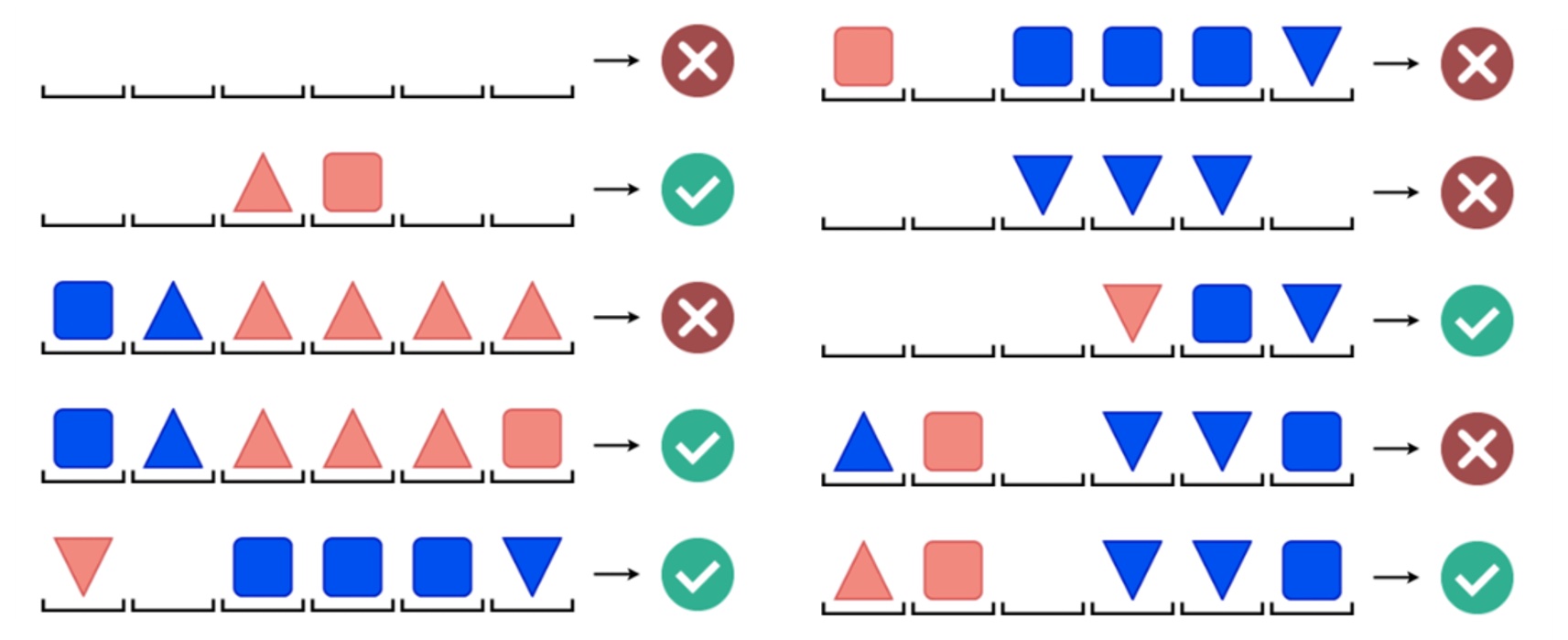}
		\end{overpic}
\caption[A game of Odeen]{\textsc{A game of Odeen.} Structures are labeled with red or green tags based on a hidden rule. Can you guess it? Hint: The rule isn't "A structure must have at least one red square" – see the fifth structure on the left with a green tag, but no red square.}
\label{fig:odeen-game}
\end{figure}

Figure \ref{fig:odeen-game} shows a typical situation in a game of Odeen. The players look at a set of structures made of simple geometric figures. Each structure is tagged red or green according to a secret rule, and the players' goal is to guess this rule.
To win the game, a player must prove to know the rule by correctly tagging a large set of new structures\footnote{\emph{The solution of the game in Figure \ref{fig:odeen-game} is at the end of this footnote.} Odeen is inspired by the board game Zendo, where players must explicitly guess the rule, known only to a master. In Zendo, players can also experiment by submitting new structures to the master. \emph{Solution: At least one square at the right of a red pyramid.}}.

\subsection{Odeen challenge} 
\label{sec:Odeen}
\begin{figure*}[h!]
\centering
    \begin{overpic}[trim=0cm 0cm 0cm 0cm, clip, tics=10, width=0.8\textwidth]{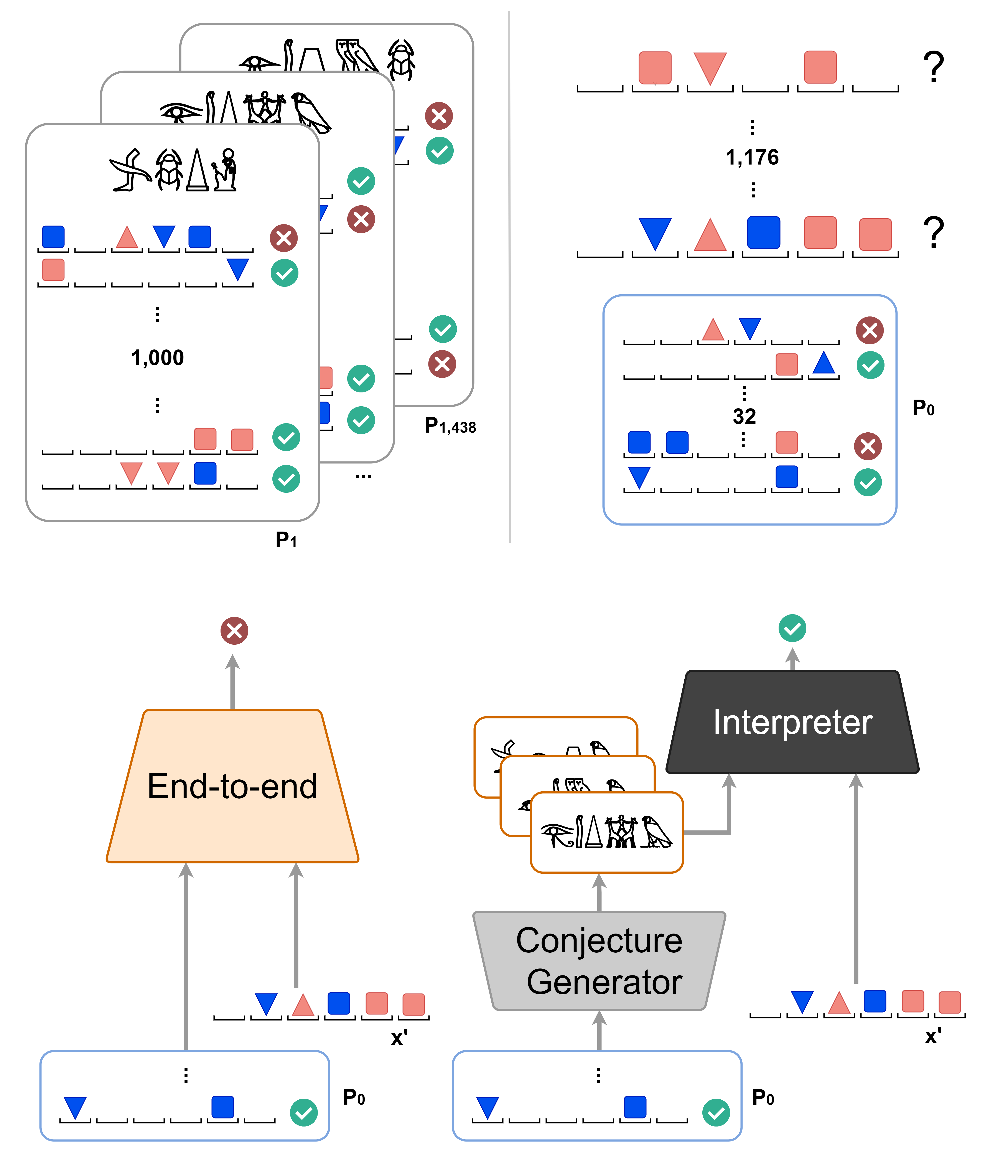}
        \put(38,53){\textsc{a}  }
		\put(80,3){ \textsc{c}  }
		\put(80,53){ \textsc{b}  }
    \end{overpic}
            \caption[Odeen Explanatory Learning problem]{
    \textsc{Odeen Explanatory Learning problem.} Given observations and explanations in an unknown language for some phenomena (\textsc{A}), plus a few observations of a new phenomenon, explain the latter and prove this knowledge by correctly tagging a large set of new samples (\textsc{B}). An empiricist approach attempts to extract this knowledge from data (\textsc{C}, left); a rationalist one conceives data as theory-laden observations, used to find the true explanation among a set of conjectures (\textsc{C}, right).
                                                }\label{el-fig-teaser}
\end{figure*}

We can see each game of Odeen as a different phenomenon of a universe, where each element is a sequence of geometric figures. In this universe, players are scientists like Galileo, trying to explain the new phenomenon; see Figure \ref{fig-galileo}. 
We can phrase the challenge for an Odeen scientist in this way: make correct predictions for a new phenomenon given few observations of it in addition to explanations and observations of some other phenomena. This is the essence of the Odeen Explanatory Learning problem, see Figure \ref{el-fig-teaser} (A and B).

\emph{
- Why do we need explanations and observations from phenomena different from the one of interest? Indeed, we are able to play Odeen from the very first game.}

\emph{
- We are able to do so only because we are -already- fluent in the Odeen language, which is a subset of English in the above case. We already have and understand all necessary concepts, such as being ``at the right of” something, but also being a ``square” or ``at least”. Otherwise, we would need past explanations and observations to first build this understanding. Before explaining the dynamic of the Jupiter moons, Galileo learned what ``Jupiter" is and what does it mean to ``have a period around" something from past explanations and examples provided to him by books and teachers.}
\setlength{\columnsep}{5pt}
\begin{figure}[h!]
\centering
  \begin{overpic}
		[trim=0cm 0cm 0cm 0cm,clip,width=0.8\linewidth]{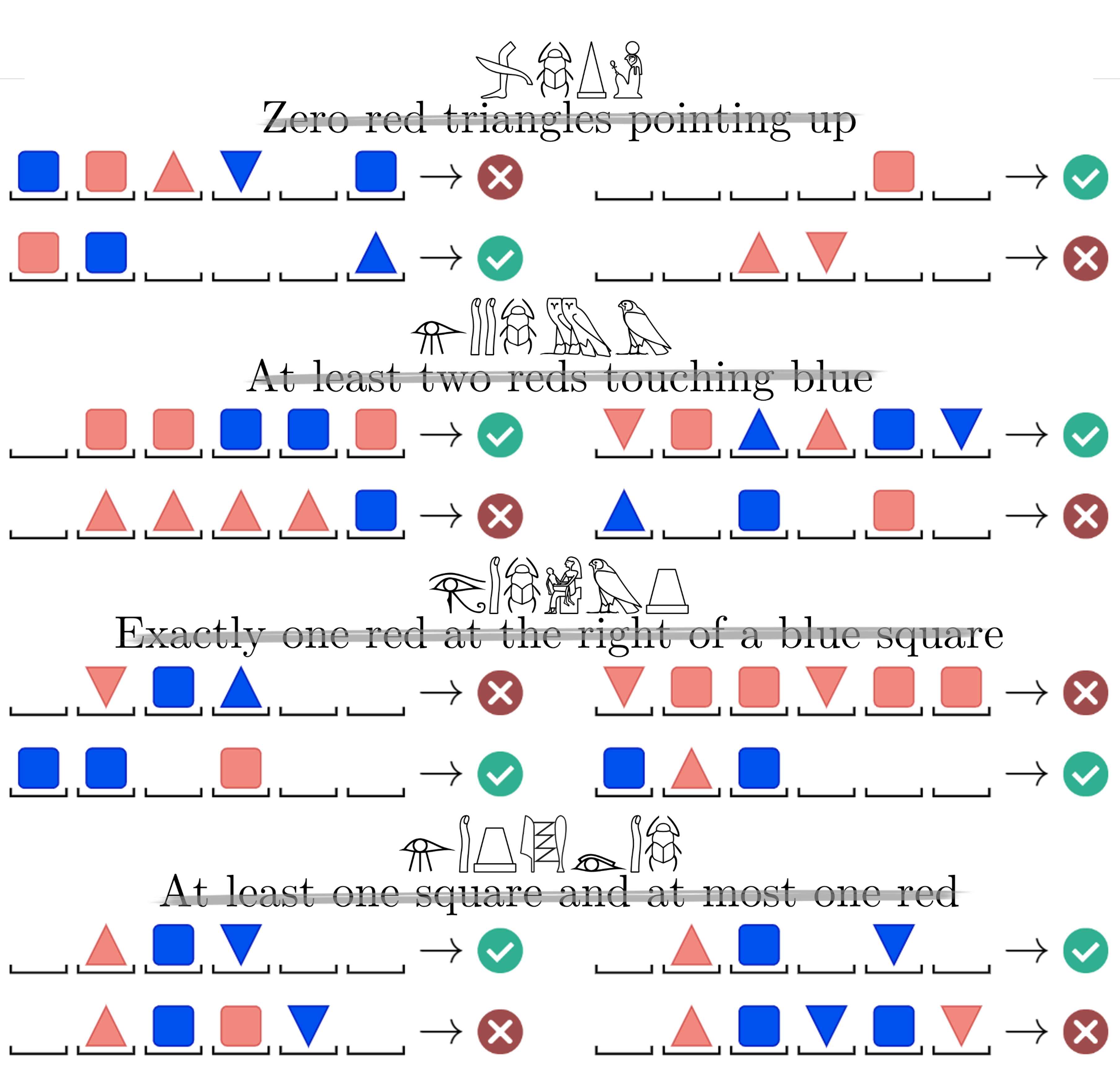}
		\end{overpic}
  \caption[Odeen science book]{\textsc{Odeen science book.} This image presents a compilation of phenomena from Odeen, each paired with a symbolic rule and its observational evidence. The language of these rules is unfamiliar, appearing as cryptic as hieroglyphics to us. However, the book also acts as a gateway to understanding this language. By comparing rules with observations, we can deduce the meaning of each symbol, eventually achieving a proficiency that allows us to explain new Odeen phenomena in this very language.}
  \label{fig:rules_zendo}
\end{figure}

In Odeen, consider the point of view of someone who does {\em not} speak the language in which the rules are written; an example of this is in Figure \ref{fig:rules_zendo}, where the secret explanations are given in hieroglyphics rather than English. Such a player would not be able to tag any structure according to the secret rule, even if the latter is given. However, assume the player has been watching several games together with their secret rules. Reasonably, the player will grow an idea of what those strange symbols mean. If the player then wins several Odeen games, it would be strong evidence of mastering the Odeen language.

\subsection{Problem formulation}
Each game of Odeen is a different phenomenon $P_i$ of a universe $U$ whose elements $x$ are sequences of geometric figures. 
 The specific task is to make correct predictions for a new phenomenon $P_0$ (a new game) given: (i) a few observations $D_0$ of $P_0$ (tagged structures), in conjunction with (ii) explanations $\{e_1, \dots, e_n\}$ and observations $\{D_1, \dots, D_n\}$ of other phenomena (other games and their secret rules).
 More formally:

\begin{mdframed}[
  backgroundcolor=black!10,
  usetwoside=false,
  innermargin=0,
  outermargin=0,
  leftmargin=0,
  rightmargin=0,
  rightline=false,
  bottomline=false,
  leftline=false,
  topline=false
  ]
Let us be given $s$ unexplained phenomena with $k$ observations each, and $n$ explained phenomena with $m$ observations each; let the $n$ phenomena be explained in an unknown language, i.e., $e_1, \dots e_n$ are plain strings without any interpreter. The task is to make $\ell$ correct predictions for each of the $s$ unexplained phenomena.
\end{mdframed}

We consider $\ell=1176$ (1$\%$ of structures); $s\hspace{-0.2mm}=\hspace{-0.1mm}1132$; $k\hspace{-0.1mm}=\hspace{-0.1mm}32$; $n\hspace{-0.1mm}=\hspace{-0.1mm}1438$ or $500$; $m$ ranges from $10K$ to $50$.

\subsection{Why not explicitly ask for the rule?}
Instead of requiring the player to reveal the secret explanation explicitly, we follow the principle of zero-knowledge proofs~\citep{zero}. In our setting, this is done by asking the player to correctly tag many unseen structures according to the discovered rule.
This makes it possible for any binary classification method to fit our EL environment without generating text. 
A winning condition is then defined by counting the correct predictions, instead of a textual similarity between predicted and correct explanation, which  would require the player to guess word-by-word the secret rule. 
In fact, different phrasings with the same meaning should grant a victory, 
e.g., ``at least one pyramid pointing up and at most one pyramid pointing up'' is a winning guess for the secret rule ``exactly one pyramid pointing up''\footnote{
The intuitive notion of meaning adopted here coincides with the pragmatic definition given by \citet[Sec.~II]{peirce-clear}, which identifies the meaning of an expression with the set of all conceivable practical consequences that derive from its acceptance. We refer the reader to \emph{Kant and the Platypus} for a readable discussion of this view \citep[Sec. 3.3]{eco-kant}, involving the first description of horses given by Aztecs.}. A brute-force enumeration of all equivalent phrasings, in turn, would not allow solutions like ``exactly one \emph{one} pyramid pointing up'', where ``one'' is mistakenly repeated twice; intuitively, we want to accept this as correct and dismiss the grammatical error. Similarly, a solution like ``exactly one pointing up'', where ``pyramid'' is omitted, should be accepted in a universe where only pyramids point up. We will reencounter these examples in Sec.~\ref{sec-exp} when we discuss the key properties of our approach.

\subsection{Dataset generation} Odeen structures are sequences of six elements including spaces, blues or reds, squares or pyramids, the latter pointing up or down. The size of the universe is $|U| = 7^6=117,649$ possible structures. We further created a small language with objects, attributes, quantifiers, logical conjunctions, and interactions (e.g., ``touching'', see Appendix A). The grammar generates $\approx$25k valid rules in total. Each of the $|U|$ structures is tagged according to all the rules. The tagging is done by an interpreter implemented via regular expressions. 
\setlength{\columnsep}{2pt}

\begin{figure}[t]
\centering
  \begin{overpic}
		[trim=0cm 0cm 0cm 0cm,clip,width=0.8\linewidth]{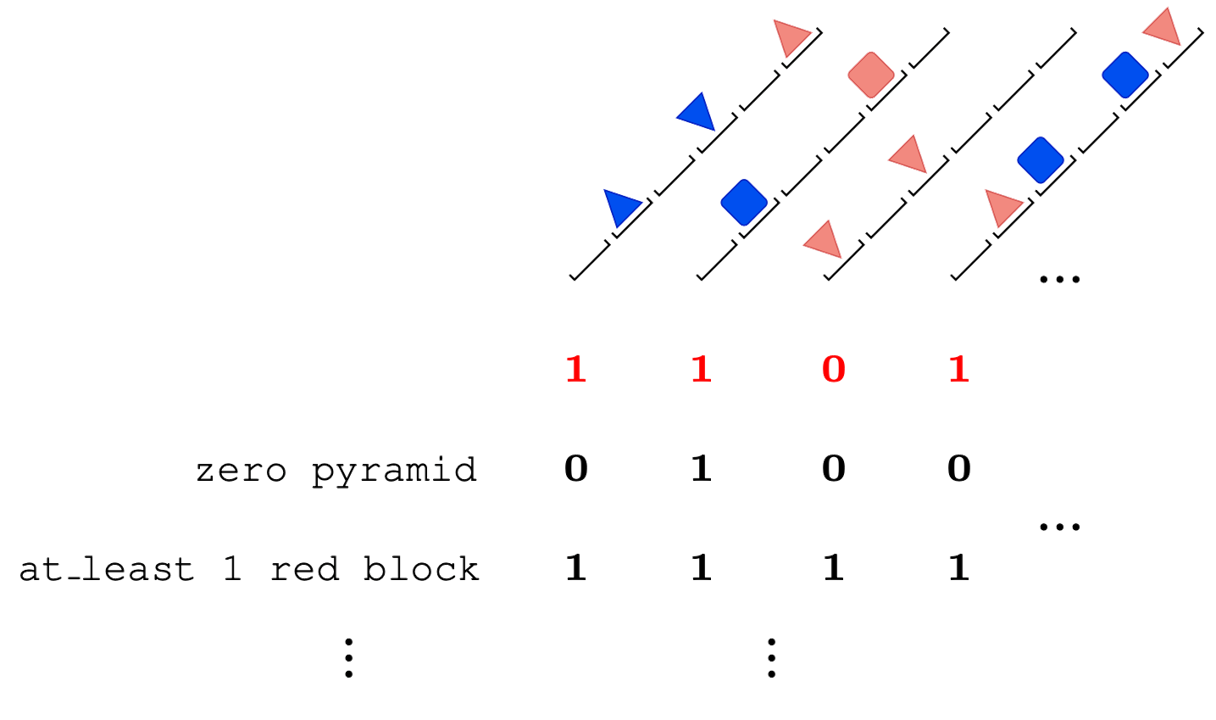}
 		\put(8.9,28){ {\textcolor{red}{\small Predicted vector $\bm{v}$}}}
		\end{overpic}
  \caption[Format of a solution for the Odeen task]{\textsc{Format of a solution for the Odeen task.} The model's output is a binary vector with labels for 1,000 new structures: '1' signifies the structure adheres to the secret rule, and '0' indicates otherwise.}
  \label{fig:metriche}
\end{figure}
\subsection{Metrics}
As described above, the task is to tag $\ell$ new structures for each of $s$ unexplained games.
An EL algorithm addressing this task encodes the predicted rule as an $\ell$-dimensional binary vector $\bm{v}$ per game (predicted vector), where $v_i=1$ means that the $i$-th structure satisfies the predicted rule, and $v_i=0$ otherwise (see Figure \ref{fig:metriche}). 
Let $\bm{w}^\ast$ be the ground-truth vector, obtained by tagging the $\ell$ structures according to the correct secret rule. 
Then, the Hamming distance $d_H(\bm{v},\bm{w}^\ast)$ 
measures the number of wrong tags assigned by the EL algorithm; if $d_H(\bm{v},\bm{w}^\ast)<d_H(\bm{v},\bm{w}_j)$, where $\bm{w}_j \neq \bm{w}^\ast$ ranges over all the possible $\approx$25k rules, then the predicted rule $\bm{v}$ made by the algorithm is deemed correct.

According to this, the {\em Nearest Rule Score} (NRS) is the number of correctly predicted rules over a total of $s$ games. 
A second score, the {\em Tagging Accuracy} (T-Acc), directly counts the number of correct tags averaged over $s$ games; this is more permissive in the following sense. Consider two different rules $A$ and $B$ sharing $99\%$ of the taggings, and let $A$ be the correct one; if an EL model tags all the structures according to the {\em wrong} rule $B$, it still reaches a T-Acc of $99\%$, but the NRS would be $0$. An EL algorithm with these scores would be good at making predictions, but would be based on a wrong explanation.

\section{Critical Rationalist Networks}
\label{sec-crns}

In principle, an EL problem like Odeen can be approached by training an end-to-end neural network to predict $\hat{y} = \mathbf{1}_{P_i}(x') $, given as input a set of observations $D_i$ and a single sample $x'$ (see Figure \ref{el-fig-teaser} C, left). 
Such a model would assume that all the information needed to solve the task is embedded in the data, ignoring the explanations; we may call it a ``radical empiricist'' approach \cite{pearl2021radical}.
A variant that includes the explanations in the pipeline can be done by adding a textual head to the network. This way, we expect performance to improve because predicting the explanation string can aid the classification task. As we show in the experiments, the latter approach (called ``conscious empiricist'') indeed improves upon the former; yet, it
 treats the explanations as mere data, nothing more than mute strings to match, in a Chinese room fashion \citep{chinese-room, bender-koller-2020-climbing}.  
In the following, we introduce a
``rationalist'' approach to solve EL problems. This approach recognizes the given explanations as existing knowledge, and focuses on interpreting them. Here theory comes first, while the data become theory-laden observations.

\subsection{Learning model}
Our {\em Critical Rationalist Networks} (CRNs) tackle the EL scientist problem introduced in Sec.~\ref{sec-el}: to find $y=\mathbf{1}_{P_0}(x')$ given $x'$, $D_0$, $\{D_1, \dots,$ $ D_n\}$, $\{e_1, \dots, e_n\}$. They are formed by two independently trained models:

        (i) A stochastic \emph{Conjecture Generator}  \[\mathcal{CG}: \{ (x,\mathbf{1}_{P}(x))_j \}_{j=1}^k \mapsto e \,,\]
    taking $k \leq |D_0|$ pairs $(x,\mathbf{1}_{P}(x)) \in D_i$ as input, and returning an explanation string $e \in \Sigma$ as output. \CG{} is trained to maximize the probability that \CG$(\Tilde{D_i}) = e_i$ for all $i=1,\dots,n$, where $\Tilde{D_i} \subset D_i$ is a random sampling of $D_i$, and $|\Tilde{D_i}| = k$.
    
        (ii) A learned \emph{Interpreter} \[ \mathcal{I} : (e, x) \mapsto \hat{y}\,,\]
    which takes as input a string $e \in \Sigma$ and a sample $x \in U$, to output a prediction $\hat{y} \in \{0,1\}$. \I{} is trained to maximize the probability that \I$(e_i, x) = \mathbf{1}_{P_i}(x)$, with $i = 1, \dots, n$ and $(x,\mathbf{1}_{P_i}(x)) \in D_i$.

\begin{figure*}
    \centering
    \begin{overpic}[trim=0cm 0cm 0cm 0.0cm,clip,width=0.999\linewidth]{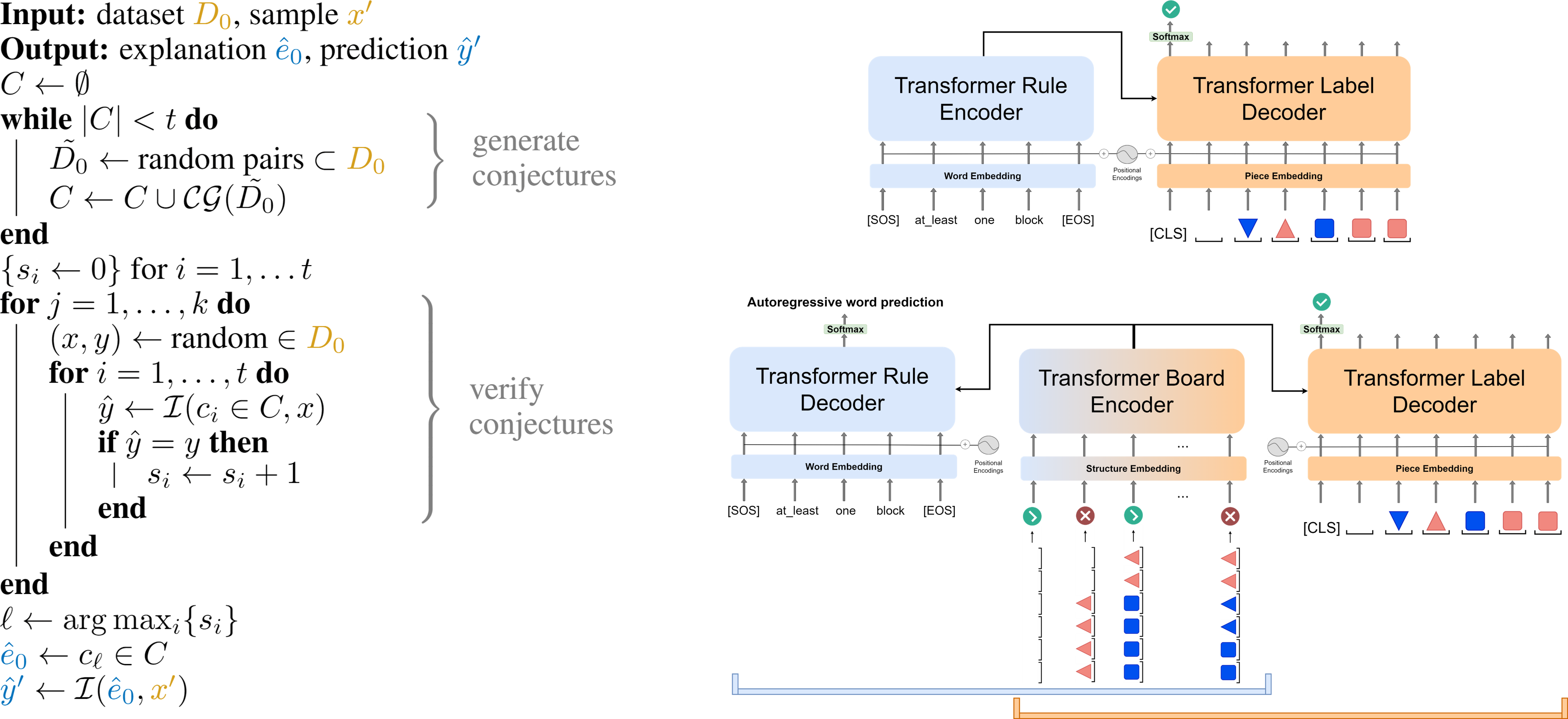}
    \put(52,38){{\color{black}\Large\I}}
        \put(47.5,3){{\color{blue}\Large\CG}}
    \put(91,1.2){{\color{orange}EMP-R}}
\end{overpic}

    \caption[Test-time algorithm of CRNs and Transformer architectures used]{\textsc{Left:} Test-time algorithm of CRNs. \textsc{Right:} CRNs are implemented using encoder-decoder transformers blocks, details of the parameters in Appendix B. \textsc{Right-top:} \I~denotes the interpreter model (rule encoder and label decoder). \textsc{Right-bottom:} The conjecture generator \CG~is composed by blue blocks. The ``radical empiricist'' (EMP-R) is composed by orange blocks. The ``conscious empiricist'' (EMP-C) baseline model consists of all the transformer blocks in the right-bottom figure, board encoder with rule and label decoders (all the blue and orange blocks).}
    \label{fig-implementation}
\end{figure*}

At test time, we are given a trained \CG{} and a trained \I{}, and we must predict whether some $x' \notin D_0$ belongs to $P_0$ or not. The idea is to first generate $t$ conjectures by applying \CG{} $t$ times to the dataset $D_0$; then, each conjecture is verified by counting how many times the interpreter \I{} outputs a correct prediction over $D_0$.
The conjecture with the highest hit rate is our candidate explanation $\hat{e}_0$ for $P_0$. Finally, we obtain the prediction $\hat{y}'$ as \I$(\hat{e}_0, x')$. See Figure~\ref{fig-implementation} (left) for a step-by-step pseudo code.

\textsc{Remarks.} The interpreter \I{} is a crucial component of our approach. A poor \I{} may fail to identify ${e}_0$ among the generated conjectures, or yield a wrong prediction $y'$ when given the correct ${e}_0$. On the other hand, 
we can work with a \CG{} of any quality and safely return as output an \emph{unknown} token, rather than a wrong prediction, whenever ${e}_0$ does not appear among the generated conjectures. 
The role of \CG{} is to trade-off performance for computational cost, and is controlled by the parameter $t$. Larger values for $t$ imply more generated conjectures, corresponding to exhaustive search if taken to the limit (as done, e.g., in \citet{clip}).
  This potential asymmetry in quality between \CG{} and \I{} is tolerated, since the learning problem solved by \CG{} is generally harder.

Secondly, although a CRN is implemented using neural networks, as we shall see shortly, its working hypothesis does not coincide with a snapshot of the countless network's parameters; rather, the working hypothesis is but the small conjecture analyzed at a given moment. This way, the CRN hypothesis is detached from the model and can only be accepted or refused in its entirety, rather than being slightly adjusted at each new data sample (Figure \ref{el-fig-teaser} C, the hypotheses are in orange).

\textsc{Implementation.} Figure~\ref{fig-implementation} (right) illustrates the architecture of CRNs, which we implement using standard encoder-decoder transformers \citep{Vaswani2017}. 
The figure also shows the architecture of the baseline methods EMP-R and EMP-C, corresponding to the end-to-end NN model and its variant with a textual head, respectively. We refer to the Appendix for further details.

\section{Experiments}\label{sec-exp}

We extensively compared CRNs to the radical (EMP-R) and conscious (EMP-C) empiricist models over the Odeen EL problem, and analyzed several fundamental aspects.

\subsection{Generalization power and data scaling laws} 
\label{par-gen}
Seeing the generalization power of a learning algorithm as its ability in discovering new knowledge from little data, the Odeen challenge asks to explain 1132 unknown phenomena for which only 32 observations are available. We measure the performance on this task through a proof of knowledge based on the successful tagging of 1176 new structures per phenomenon (NRS).
The information available at training time consists of symbolic explanations from 1438 known phenomena paired with $m$ observations each (see Fig. \ref{el-fig-teaser}A), we evaluated several settings with $m$ ranging from $10K$ to $50$.
No test explanation is equivalent to the ones seen at training\footnote{An important example is the bigram ``exactly two'', which appears in the test set, but was deliberately excluded from training; the training rules only contain ``at least/most two'' and ``exactly one''. With $m=1K$, the CRN guessed $40\%$ of the 72 test rules with ``exactly two'', while the empiricist models (EMP-C, EMP-R) scored $4\%$ and $0\%$ respectively.}. Some example games can be found in Appendix D.

\begin{table}[h]
    \centering
    \begin{small}
    \begin{sc}
    \begin{tabular}{l|ccc}
        \toprule
        Model &  $\text{{NRS}}$ & $\text{{T-Acc}}$ & \text{{R-Acc}} \\
        \midrule
        \CRN{}    & \textbf{\SoftRes{0.777}{0.780}} & \textbf{0.980} &  \textbf{0.737}\\
        \EA{} &  \SoftRes{0.225}{0.240} & 0.905 & 0.035 \\
        \ER{}    &  \SoftRes{0.156}{0.161}  & 0.898 & - \\
        \bottomrule
    \end{tabular}
    \end{sc}
    \end{small}
    \caption[Rationalist vs Empiricist models]{\textsc{Rationalist vs Empiricist models.} Performance of different models on 1132 unknown phenomena. \textsc{NRS} is our proxy to measure an effective discovery of the unknown rule governing a phenomenon. \textsc{T-Acc} is the absolute tagging accuracy, while \textsc{R-Acc} traces if the rule provided in output is equivalent to the ground-truth rule with respect to the hradcoded interpreter.}
    \label{tab:zendo_data_results}
\end{table}

Table \ref{tab:zendo_data_results} reports the full results for $m=1K$. Here \textsc{NRS} is our proxy to measure if the new phenomenon is understood by the model. \textsc{T-Acc} is the absolute tagging accuracy, while R-Acc is our proxy to measure if the understanding is properly explained: it tracks how frequently the output explanation is equivalent to the correct one. Specifically, here we say that two rules $A$ and $B$ are equivalent if the tags assigned by the hard-coded interpreter to all the $\sim$117k structures in $U$ are the same for $A$ and $B$. 

The best Empiricist model understands only $22.5\%$ of the new phenomena, and the explanations it provides are usually broken (R-Acc $3.5\%$). This gap of almost $20$ points is not surprising given that EMP-C gives no guarantee that the predicted explanation will be consistent with the tags prediction.
Conversely, the CRN understands $77.7\%$ of the new phenomena, and the explanations provided are almost always comprehensible: $73.7\%$ of them could be used by the hard-coded interpreter to make sensible predictions on the new phenomena. Even if small, this $4\%$ gap may be surprising: If CRN's predictions stem from its explanations, how can flawed explanations still yield accurate taggings? The underpinnings of this disparity will be delved into in the following lines.

\begin{figure}[h]
    \centering
    \includegraphics[trim=0.18cm 0.5cm 0.1cm 0.3cm,clip,width=0.8\linewidth]{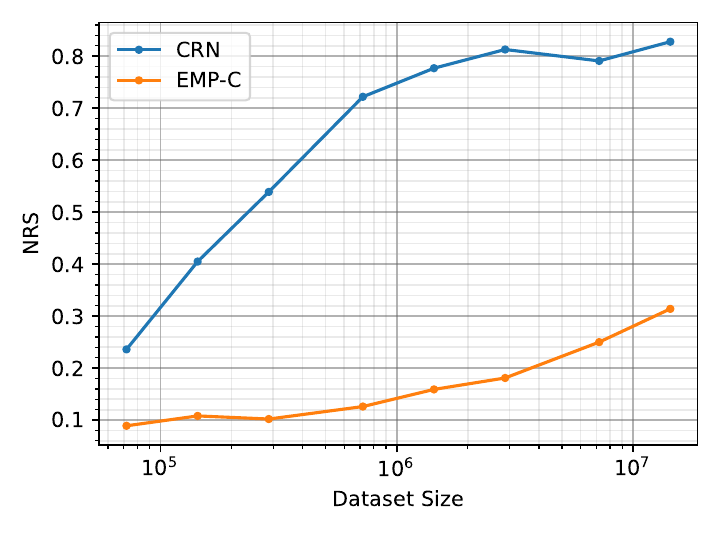}
    \caption[The rationalist model is more data efficient than the empiricist model]{\textsc{The rationalist model is more data efficient than the empiricist model.} Despite the EMP-C and CRN have approximately the same number of learnable parameters ($\approx 6M$), the CRN performance grows faster and earlier as data scales. The EMP-C barely improves before training at least on 1 million samples, and achieves the first CRN score ($21\%$) with a delay of two orders of magnitude ($m = 50$ vs $5K$)}
    \label{fig:zendo_data_trends}
\end{figure}

\subsection{Handling ambiguity and contradiction}
\label{par-amb}
One may reasonably expect that a CRN equipped with the ground-truth interpreter used to generate the dataset, would perform better than a CRN with a learned interpreter.
Remarkably, this is not always the case, as reported in Table~\ref{tab-ambiguity}.

The better performance of the fully learned interpreter over the ground-truth one 
is due to its ability to process ill-formed conjectures generated by the \CG{}. The conjecture ``at least one pointing up'' makes the hard-coded interpreter fail, since ``pointing up'' must always follow the word ``pyramid'' by the grammar. Yet, in Odeen, pyramids are the only objects that point, and the learned \I~interprets the conjecture correctly. Other examples include: ``exactly one red block touching pyramid blue'' (``pyramid'' and ``blue'' are swapped), or the contradictory ``at least one two pyramid pointing up and exactly one red pyramid'', which was interpreted correctly by ignoring the first ``one''. When the learned interpreter is not very accurate, the negative effect of errors in tagging prevails.

\begin{table*}[ht]

\vskip 0.15in
\begin{center}
\begin{small}
\begin{sc}
\begin{tabular}{ll|cc|c}
\toprule
\multicolumn{2}{l}{} &  \multicolumn{2}{c}{NRS of a CRN with} & T-Acc  \\
\multicolumn{2}{l}{Train Data} &  Learned \I{} & Hardcoded \I{} & 
Learned \I{} \\

\midrule
10K struct. &1438 rules & \textbf{0.813} & 0.801 & 0.997\\

1K struct. &1438 rules & \textbf{0.777} & 0.754 & 1.000\\

100 struct. &1438 rules & 0.402 & \textbf{0.406} & 0.987\\

\midrule
10K struct. &500 rules & 0.354 & \textbf{0.377} & 0.923\\

1K struct. &500 rules & 0.319 & \textbf{0.336} & 0.924\\

100 struct. &500 rules & \textbf{0.109} & 0.101 & 0.920\\

\bottomrule
\end{tabular}
\end{sc}
\end{small}
\end{center}

\caption[Explanatory Learning vs Program Synthesis paradigm]{ \textsc{Explanatory Learning vs Program Synthesis paradigm.}
Performance comparison of a data-driven vs ground-truth interpreter in a CRN. The last column shows the tag prediction accuracy of the learned \I{}, when provided with the correct rule.}

\label{tab-ambiguity}
\end{table*}

Making sense out of ambiguous or contradictory messages\footnote{This is one of seven essential abilities for intelligence as found in \emph{GEB} \citep[Introduction]{hofstadter1979godel}.} 
is a crucial difference between a learned interpreter vs a hardcoded one. As \citet{Rota-pernicious} reminds us, a concept does not need to be precisely defined in order to be meaningful. Our everyday reasoning is not precise, yet it is effective. ``After the small tower, turn right''; we will probably reach our destination, even when our best attempts at defining ``tower'', as found, e.g., in the Cambridge dictionary, begin with ``a \emph{tall}, narrow structure...''.

\subsection{Explainability}
The predictions of a CRN are {\em directly caused} by a comprehensible explanation that is available in the output; this makes CRNs explainable by construction. Further, CRNs allow counterfactuals; one may deliberately change the output explanation with a new one  to obtain a new prediction. 
The bank ML algorithm spoke: ``Loan denied''; explanation: ``Two not paid loan in the past and resident in a district with a high rate of insolvents''. With a CRN, we can easily discard this explanation and compute a new prediction for just ``Two not paid loan in the past''.

Importantly, by choosing a training set, we control the language used for explanations; i.e., we explicit the biases that will steer the learning of generalizations \citep{mitchell1980need}.  This allows a CRN to ignore undesirable patterns in the data (e.g., skin color) if these can not be expressed in the chosen language.  If the Odeen training set had no rule with ``pointing up/down'', the learned interpreter would see all equal pyramids, even with unbalanced training data where 90$\%$ of pyramids point up.    
On the contrary, current explainability approaches for NNs (end-to-end empiricist models) either require some form of reverse engineering, e.g., by making sense out of neuron activations~\citep{microscope}, or introduce an ad-hoc block to generate an explanation \emph{given} the prediction, without establishing a cause-effect link between the two~\citep{hendricks2016generating, hind2019ted}. 
This practice produces explanations that are not reliable and can be misleading \citep{rudin2019stop}, on the contrary CRNs' explanations are faithful to what the model actually computes.

\subsection{Adjustable thinking time.} 
\begin{figure}[h!]
\centering
  \begin{overpic}
		[trim=0cm 0cm 0cm 0cm,clip,width=0.8\linewidth]{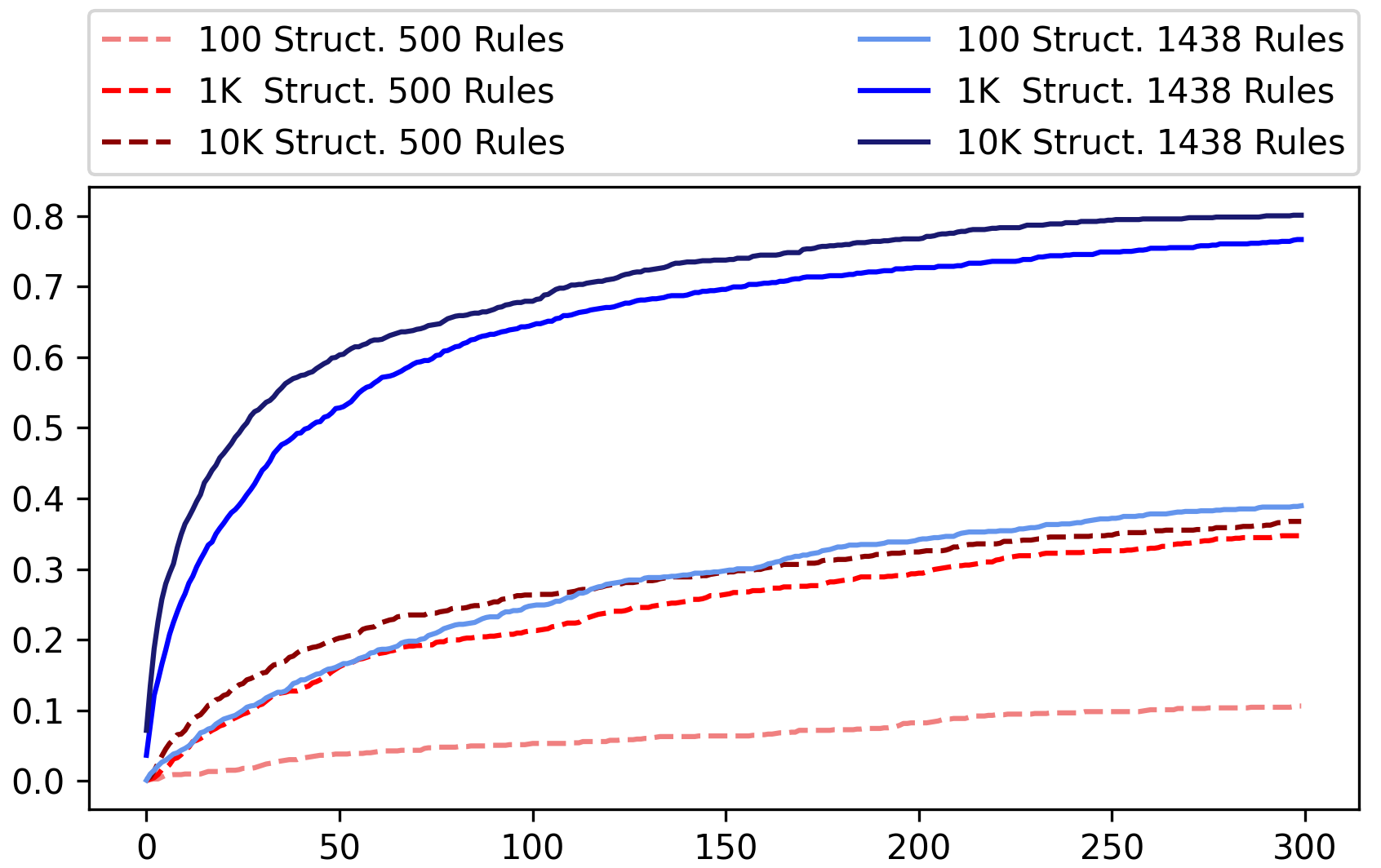}
		\end{overpic}
\caption[How generating more conjectures improves the performance of a CRN]{\textsc{How generating more conjectures improves the performance of a CRN.} Different colors show different sizes of the training dataset.}
\label{fig:conjecture-trend}
\end{figure}

End-to-end models do not exhibit a parameter to adjust their processing to the complexity of the incoming prediction. 
By contrast, CRNs have a test-time parameter $t$, corresponding to the number of generated conjectures, which trades off computational cost for performance. In Figure \ref{fig:conjecture-trend}, we plot the cumulative R-Acc score ($y$ axis) against the number $t$ of generated conjectures ($x$ axis). The curves show that $>60\%$ of correct explanations are found within the first $50$ candidates, and $>80\%$ are within the first $300$. As a reference, a brute force exhaustive search would reach $100\%$ over a search space of $24,794$ possible explanations.

\setlength{\columnsep}{3pt}

\subsection{Prediction confidence} 
As explained in Sec.~\ref{sec-crns}, at test time the CRN selects the conjecture with the highest hit rate among the ones generated by the \CG{}. Alternatively, one may keep only the conjectures coherent with {\em all} the structures in the table, returning an ``unknown explanation'' signal if no such conjectures are found.
If the interpreter is sufficiently accurate, this stricter condition barely deteriorates the CRN performance, and 
 it will never return a prediction based on a possibly wrong explanation. 
For example, tested in a setting with $n=1438$, $m=1000$, this stricter CRN discovers the correct explanation for 861 out of 1132 new phenomena ($76\%$), and admits its ignorance on the other 271. Conversely, evaluating the confidence of an end-to-end neural network remains an open problem \cite{meinke2019towards}.

\section{Related Work}\label{sec:related}

\textsc{Epistemology.} The deep learning model we propose in this work, CRNs, is designed according to the epistemological theory of critical rationalism advanced by \citet[]{popper1935logic}, where knowledge derives primarily from conjectures, criticized at a later stage using data.
\citet{deutsch2011beginning} remarks that to make this critique effective, conjectures should not be adjustable but can only be kept or rejected at each new data sample, as done in CRNs at test time.
Only in this way we can discover explanations with ``reach", namely that maintain predictive power in novel situations.

\textsc{Machine learning.} Explanatory Learning enriches the fundamental problem of modern program synthesis \citep{balog2019deepcoder, sun2018neural, ellis2020dreamcoder} by including the interpretation step among what should be learned. 
As seen in Section \ref{par-amb}, a learned \I{} can grant better performance by exploiting the ambiguity of language to impose new meaning on arbitrary substrates, which \citet{santoro2021symbolic} recognize as a fundamental trait of symbolic behavior.

Despite the similar underlying motivation, EL fundamentally differs from current meta learning approaches \citep{VinyalsBLKW16, Santoro16Meta, wang2016learning, weng2018meta}, since it does not prescribe any parameter adaptation at test time. In this sense, EL resembles more the Domain Generalization setting, which involves designing a specific model that is robust to domain shift \cite{li2018learning}. Unlike DG, EL requires labels to be symbolic sequences rather than a single symbol, but makes no assumption of identical labels between training and test domains\footnote{The DG version of the Odeen test set would contain the same rules of the training set, changing just the accompanying structures.}. EL requires just that all symbols present in the test sequences are seen at training. While current SOTA approaches in DG do not significantly outperform Empirical Risk Minimization \cite{gulrajani2020search}, in the EL setting CRNs overcome the performance of standard ERM ({\sc EMP-C}) by a large margin.
Recent literature finds few yet remarkable approaches that fit our EL paradigm, such as CLIP \cite{clip} in the vision area, Generate \& Rank \cite{shen2021generate} for Math Word Problems in NLP, and the Socratic Models of \citet{zeng2023socratic}, that use language as an intermediate representation between different neural models. 
Starting from the latter idea, \citet{andreas2017learning} train an interpreter and a captioning model to solve an abstract problem similar to Odeen. Differently from this work, their task is easier and solved also by an empiricist approach \cite{mu2019shaping}. Indeed, they focus on the benefits of representing samples using natural language, rather than on finding the unique explanation of a set of observations. This latter goal also led us to show the fundamental difference between a learned interpreter and a given fixed one.

The Odeen challenge continues the tradition of AI benchmarks set in idealized domains \cite{mitchell2021abstraction}, that started with Bongard problems \cite{bongard1967pattern}. Unlike CLEVR~\citep{clevr} and ShapeWorld~\citep{alex2017shapeworld}, Odeen focuses on abduction rather than deduction. Unlike ARC \cite{chollet2019measure}, Odeen is a closed environment providing all it takes to learn the language needed to solve it. Unlike the ShapeWorld adaptation of \citet{andreas2017learning}, its score is measured in terms of discovered explanations rather than sparse guessed predictions; further, the test and training set do not share any phenomenon.

\textsc{Learning theory.} Finally, we point out that the expression \emph{Explanatory Learning} was previously used by~\citet[Sec.~7]{aaronson2013philosophers} to argue about the necessity of a learning theory that models ``predictions about phenomena different in kind from anything observed''.
The author pointed to the work of \citet{Angluin}, who generalized the PAC learning model 
by moving the goal from successful predictions to comprehensive explanations.

\section{Conclusions}
\label{sec-conc}

Recently, the attention on the epistemological foundations of deep learning has been growing. 
The century-old debate between empiricists and rationalists about the source of knowledge persists, with two Turing prizes on opposite sides; 
\citet[]{epistemologyLecun} argues that empiricism still offers a fruitful research agenda for deep learning, while \citet[]{pearl2021radical} supports a rationalist steering to embrace model-based science principles.
This new debate is relevant, since as Pearl notes, today we can submit the balance between empiricism and innateness to experimental evaluation on digital machines.

\textsc{Limitations and future directions.}
EL models the essential part of the knowledge acquisition process, namely the interval that turns a mute sequence of symbols into an explanation with reach. 
However, our modeling assumes a representative set of observations $D_0$ to be given (the $k=32$ structures of the new phenomenon). A more comprehensive explanatory model would allow the player to do without these observations, including an interaction phase with the environment where the $D_0$ itself is actively discovered. We see this as an exciting direction for follow-ups.

Odeen has potential as a parametric environment  to  experiment  with  EL  approaches. The structure length (6 in this work), the number of shapes (3), attributes (2), and the grammar specifications (see Appendix A) can be easily tweaked to obtain either simpler or significantly more complex environments. The design choices of this work provide a good starting point; the resulting benchmark is far from being saturated, but experimenting with different variants is a possibility for the future.

%% file: Chapters/Chapter04.tex
\newtheorem{theorem}{Theorem}[chapter]
\newtheorem{definition}[theorem]{Definition}

\definecolor{rr}{RGB}{114,161,11}
\definecolor{ar_i}{RGB}{71,186,178}
\definecolor{ar_t}{RGB}{135,155,179}

\newcommand{\highlight}[1]{#1}

\chapter{ASIF: Coupled Data Turns Unimodal Models to Multimodal without Training}\label{ch:asif} \chaptermark{ASIF: Coupled Data Turns Unimodal Models to Multimodal}

\section*{Chapter abstract}
In this chapter we focus on learned Interpreters, a fundamental component of agents solving Explanatory Learning problems. Specifically, we will dissect the functioning of multimodal vision-language foundation models, among the most used learned interpreters nowadays. 
The CLIP \cite{clip} model has opened this research direction, proving that aligning visual and language spaces is key to solving many vision tasks without explicit training, but required to train image and text encoders from scratch on a huge dataset. LiT \cite{lit} improved this by only training the text encoder and using a pre-trained vision network. In this chapter, we show that a common space can be created without any multimodal training, using single-domain encoders (trained with or without supervision) and a much smaller amount of image-text pairs.
Furthermore, our model has unique properties. Most notably, deploying a new version with updated training samples can be done in a matter of seconds.
Additionally, the representations in the common space are easily interpretable as every dimension corresponds to the similarity of the input to a unique image-text pair in the multimodal dataset.
Experiments on standard zero-shot visual benchmarks demonstrate the typical transfer ability of image-text models. Overall, our method represents a simple yet surprisingly strong baseline for foundation multimodal models, raising important questions on their data efficiency and on the role of retrieval in machine learning.

\blfootnote{This chapter is based on the paper \textit{"ASIF: Coupled Data Turns Unimodal Models to Multimodal without Training"}, by Antonio Norelli, Marco Fumero, Valentino Maiorca, Luca Moschella, and Emanuele Rodolà.}

\newpage

\begin{figure}[h]
\centering
		
         \begin{overpic}
		[trim=0cm 0cm 0cm 0cm,clip,width=0.8\linewidth]{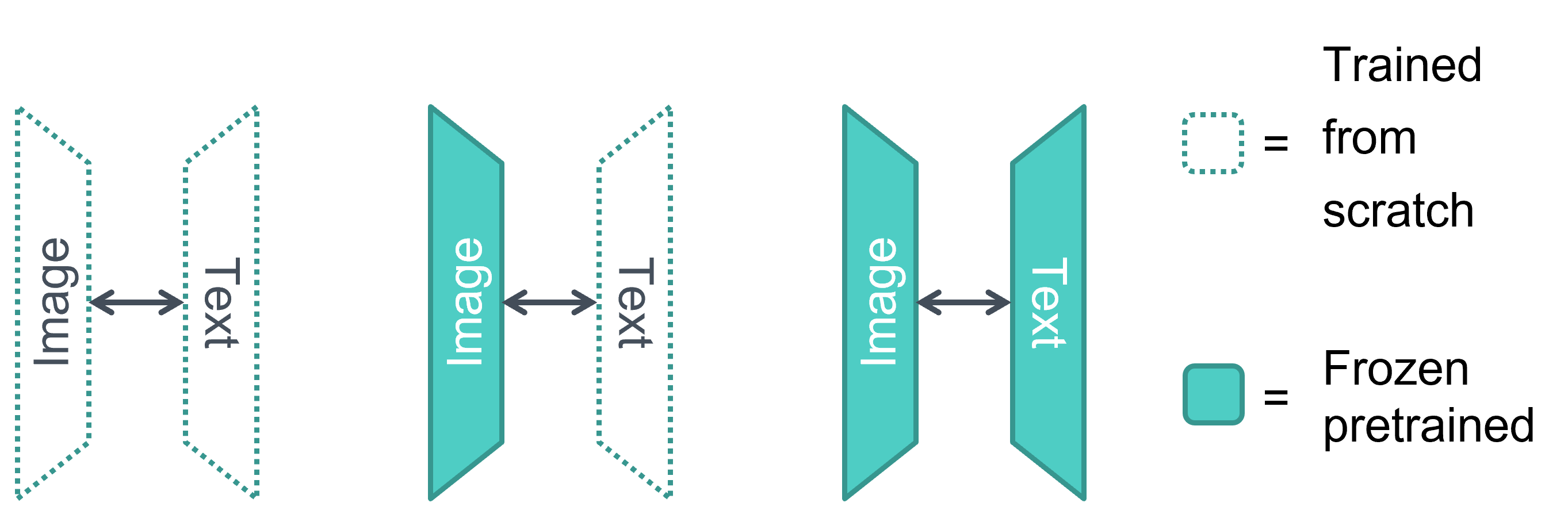}
\put(5,28){\small CLIP}
\put(32.8,28){\small LiT}
\put(57.5,28){\small ASIF}
 		\end{overpic}
   		 		   
   \caption[ASIF is a simple recipe to align the representations of two frozen pre-trained encoders]{ASIF is a simple recipe to align the representations of two frozen pre-trained encoders.}
   \label{fig-models}
\end{figure}

\section{Introduction}

Large multimodal models such as CLIP~\cite{clip} are rapidly becoming the standard for foundation models~\cite{bommasani2021opportunities} in computer vision. This is largely due to their zero-shot and open-world capabilities that enable diverse suites of downstream tasks, from classification to detection and visual search. 

Overall, \citet{clip} demonstrated that treating image recognition as language interpretation allows generalizing to a multitude of tasks without training explicitly for them. 
By building an interpreter, CLIP changed the way computer vision is approached: rather than extracting the ``dog'' label from an image like a CNN \cite{lecun1989backpropagation}, CLIP tests the hypothesis of a dog being in the image against all the other hypotheses.
The success of this image-text association is a testament to the power of deep learning: the CLIP model was the first of its kind, and building it required a joint training of two large neural encoders on a vast collection of image-text pairs.

Still, training neural networks at such scale presents several challenges beside the obvious infrastructure and training costs. 
Notably, it requires collecting massive training sets, making it difficult to interpret the predictions of the model in light of their training data. Additionally, the training assets are often not owned by the institution training the model~\cite{sun2017revisiting}. 
\begin{figure}[h]
\centering

    \begin{overpic}[trim=0cm 0cm 0cm 0cm,clip,width=0.8\linewidth]{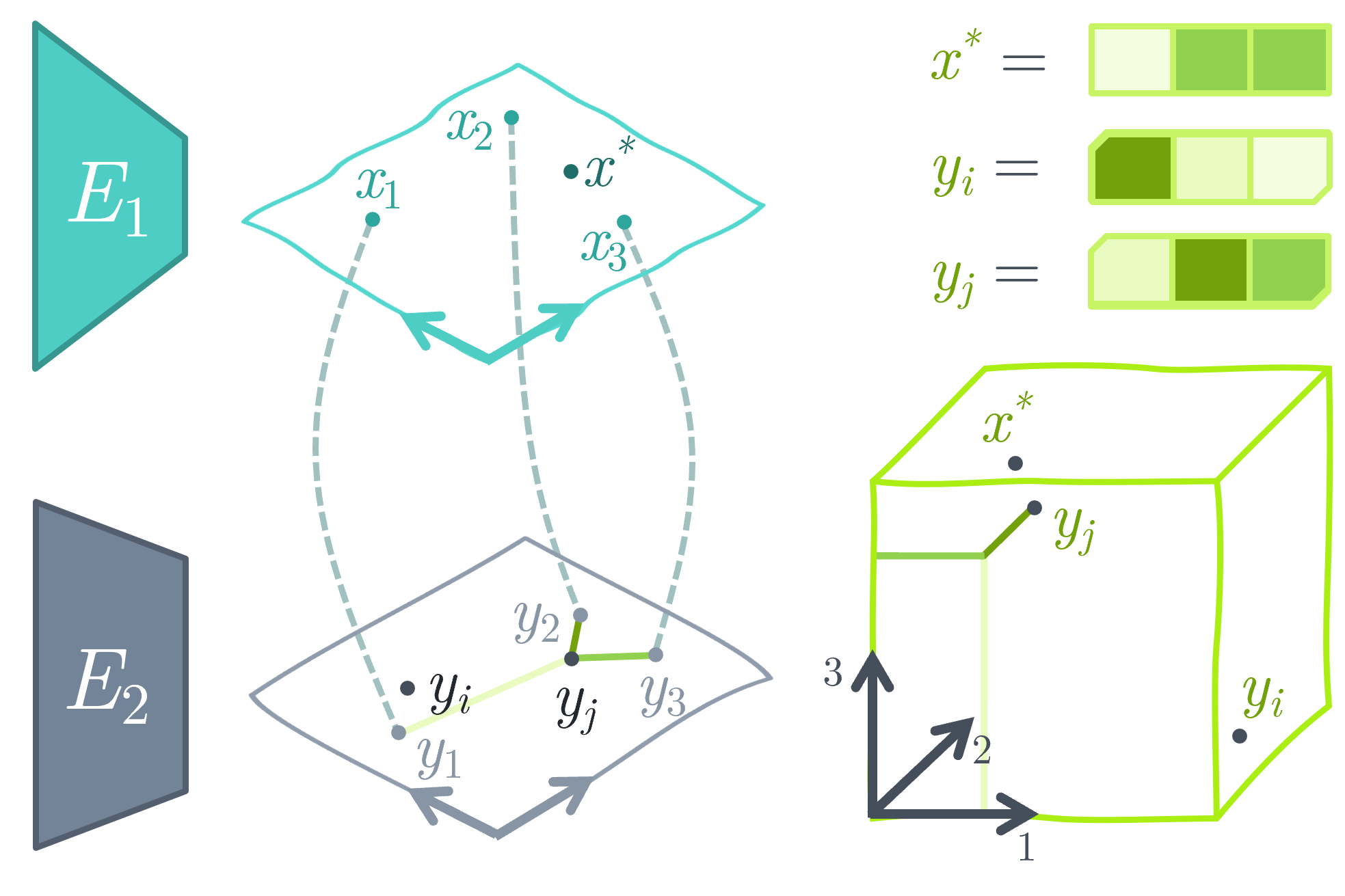}
                                    \end{overpic}
        \caption[The ASIF construction]{\textsc{The ASIF construction.} An ASIF model is defined by two unimodal pretrained encoders and a collection of coupled embeddings. This is sufficient to compare elements from different modes through representations made of similarities with ground-truth pairs:
    ${\color{rr}{y_j}}$ is more similar to ${\color{rr}{x^*}}$ than ${\color{rr}{y_i}}$.  
        }
    \label{fig-teaser}   
\end{figure}
This introduces several additional challenges, from reproducibility to the difficulty of ensuring that an asset owner can remove their data from the model~\cite{golatkar2020forgetting,ginart2019making,guo2020certified}.  Overall, these considerations make large multimodal models relatively inaccessible to researchers and practitioners until checkpoints are released or access to demo is granted. And even then, the ability to tweak the models by adding or removing training data or to interpret their results is limited.

In this chapter, we present ASIF, a simple non-parametric procedure that turns pre-trained uni-modal image and text encoders into a multimodal model using a \textit{relatively small}\footnote{CLIP \cite{clip} experiments used from 400M to 15M captioned images as training samples, LiT \cite{lit} from 901M to 10M. Our experiments use 1.6M.} 
collection of image-text pairs and no additional training, as depicted in Figure~\ref{fig-models}. The resulting model is functionally equivalent to CLIP, effectively producing aligned representations of images and their captions.

\begin{figure}[h]
\centering
		
     \begin{overpic}
				[trim=0cm 0cm 0cm 0cm,clip,width=0.7\linewidth]{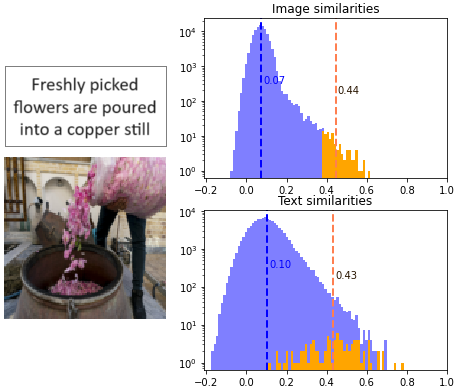}
 		\end{overpic}
   
   \caption[Captions of similar images are themselves similar]{\textsc{Captions of similar images are themselves similar.} Distribution of similarities of 100k embedded pairs in the training set versus the above image and caption embeddings. We highlighted in orange the 1000 pairs with the highest image similarity. }
 		 		\label{fig-similar}
\end{figure}

\textsc{Intuition.} The key insight is that captions of similar images are themselves similar (Fig. \ref{fig-similar}), and therefore a representation crafted using just similarities to ground-truth multimodal pairs is quasi mode-invariant (Fig. \ref{fig-teaser}).

The ASIF procedure is not only efficient but also has several intriguing features built-in. One of the key advantages is the ability to easily edit the model - adding new image-text pairs or removing outdated ones is as simple as encoding or canceling the corresponding embeddings. Furthermore, the multimodal representations are highly interpretable, as each dimension corresponds to the similarity of the input to a specific entry in the multimodal dataset.

 \textsc{Contribution. }Our results are surprising and raise several questions. Despite (1) the simplicity of the approach, (2) a multimodal dataset that is up to 250 times smaller than in prior work and (3) the lack of actually training the model on multimodal data, ASIF achieves zero-shot classification accuracy on downstream datasets that is comparable to CLIP~\cite{clip,lit}. This raises important questions on the data efficiency in foundation models, making ASIF a very powerful and cheap baseline for future work, and opening new doors for data centric AI~\cite{datacentric}.

In summary, we:

\begin{itemize}
    \item Introduce the ASIF procedure, which turns two pretrained unimodal black-box encoders into an interpretable multimodal model without tuning a neuron.
    \item Demonstrate the effectiveness of ASIF models on zero-shot image classification tasks, where they achieve performance in the same ballpark of CLIP with significantly fewer image-text pairs.
    \item Discuss the unique properties of ASIF, its implications on the role of memory and retrieval in machine learning, and the new opportunities it opens.
\end{itemize}

\section{Aligning Pre-Trained Models with ASIF}
\label{sec-method}

 In the following we present how a collection of captioned pictures implicitly defines a common space for images and texts through relative representations \cite{moschella2022relative}, allowing to build a multimodal model without training.  
Here we focus exclusively on vision and language as modalities due to the wider availability of relevant models and paired data. However, we anticipate that our procedure could be more generally applicable.
Indeed, subsequent research by other teams has already utilized ASIF as a baseline in other modalities, such as audio \cite{wang2023understanding}.
Before delving into the specifics of our method, we will briefly review existing techniques for establishing this common space.
\newpage

    {\centering
    \begin{overpic}[trim=0cm 0cm 0cm 0cm,clip,width=0.75\linewidth]{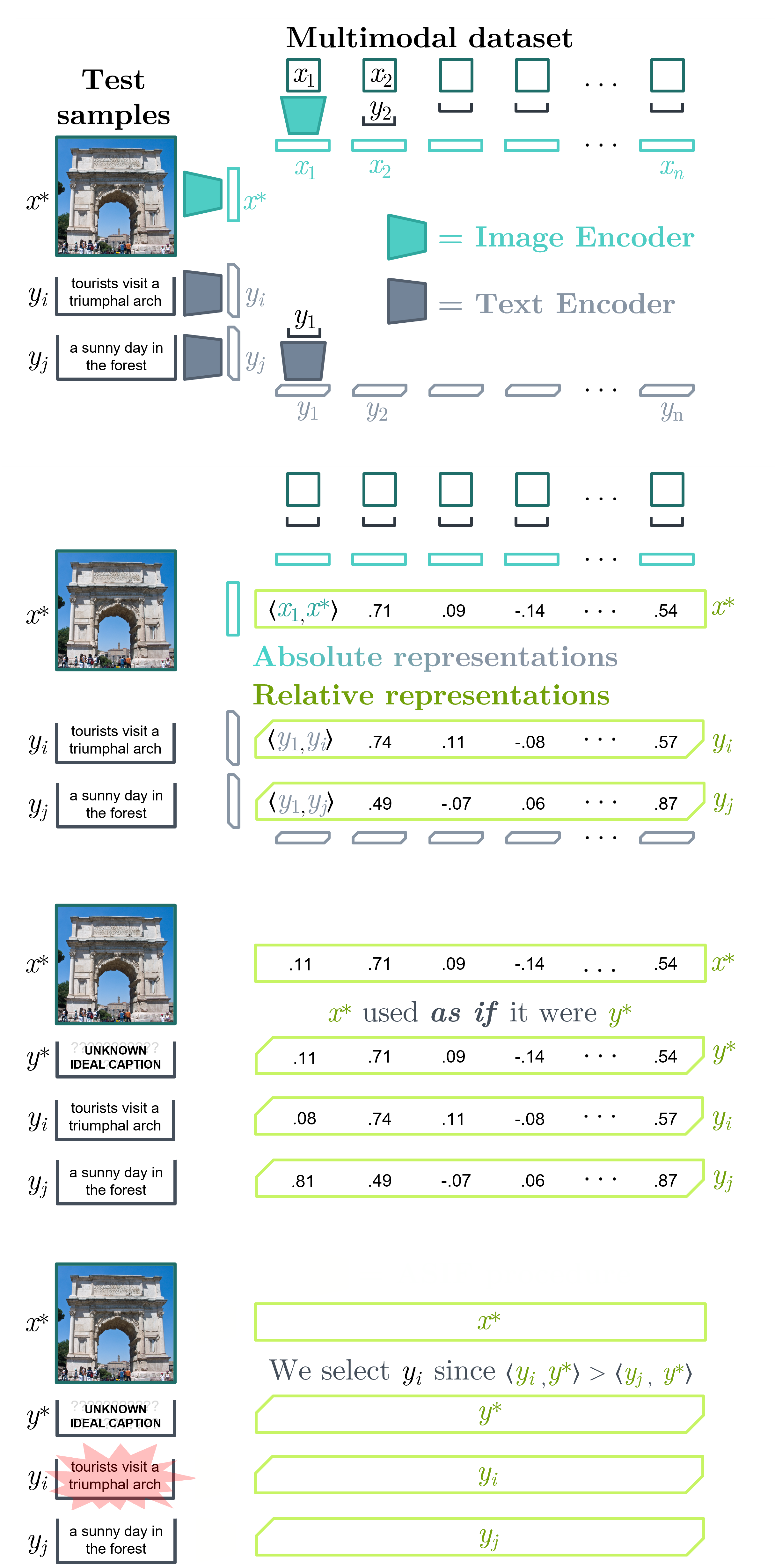}
        \put(49,83){\textsc{a}  }\put(49,31){\textsc{c}  }\put(49,56){\textsc{b}  }\put(49,8){\textsc{d}  }
		\end{overpic}
    \captionof{figure}[Zero shot classification with ASIF]{\textsc{Zero shot classification with ASIF.} In this example we determine the best caption for the image $x^*$ from the two candidates, $y_{i}$ and $y_{j}$. \textsc{a.} Compute and store the embeddings of the multimodal dataset and the test samples. \textsc{b.} Compute the relative representation of the test image and the candidate captions. \textsc{c.}  We consider the relative representation of $x^*$ with respect to the image collection $x_1, \dots, x_n$ \textit{as if} it was the relative representation of $y^*$ -- the ideal caption for $x^*$ -- with respect to the corresponding caption collection $y_1, \dots, y_n$. \textsc{d.} We choose the candidate caption most similar to the ideal one.}   
                \label{fig-main}
}
\newpage

\subsection{Contrastive training to build a common space. } \ With multimodal models, we refer to architectures that 
embed inputs of diverse modes into the same space. The better is a multimodal model, the closer are representations of different modes of the same object.
So far, this common space has been obtained as the result of a contrastive training of one \cite{lit} or both the neural mode encoders \cite{clip, align}. Using a collection of image-text pairs as training set, a contrastive loss promotes the closeness of image and text embeddings from the same pair, while spacing out the ones from distinct pairs. \citet{lit} train just the text encoder to match the image embeddings of a pretrained visual encoder.
Once the encoders are trained, a multimodal model can be adapted to perform any visual classification task just by crafting a caption for each label and selecting the one with the embedding closer to the embedding of the input image (zero-shot image classification).

\subsection{Relative representations.} \ 
Our idea to build a common latent space is to use a collection of coupled data as a ``rosetta stone'' between modalities, and represent each new data point as its similarities to the points of the same modality in the collection. In other words, we compute a \textit{relative representation} for each new data point:

\label{def:rel-rep}
\textsc{Definition.} Given an encoder ${\color{ar_i}{E}}: X \to {\color{ar_i}{\mathbb{R}^d}}$ and a subset of samples $\{x_1, \dots, x_n\}$ denoted as anchors, we define the relative representation of $x'$ as the $n$-dimensional vector:
$${\color{rr}{x'}} = [\text{sim}({\color{ar_i}{x'}}, {\color{ar_i}{x_1}}), \;\dots\;, \text{sim}({\color{ar_i}{x'}}, {\color{ar_i}{x_n}})]$$
for some similarity function sim, e.g. the cosine similarity. 
$x_i \in X$ are input samples, e.g. images or texts; ${\color{ar_i}{{x_i}}} \in {\color{ar_i}{{\mathbb{R}^d}}}$ are the embeddings of $x_i$ obtained with the encoder ${\color{ar_i}{{E}}}$, i.e. the absolute representations of $x_i$; while ${\color{rr}{{x_i}}} \in {\color{rr}{{\mathbb{R}^n}}}$ are the relative representations of $x_i$.

We observe that when each anchor is available in two or more modalities, we can compute relative representations of samples from those modalities using the same subset of anchors. Most notably, these relative representations will all live in the same space, even when they are representing samples from different modalities. This foundational insight is illustrated in Figure \ref{fig-teaser}.

\subsection{Relation with Kernel methods.} Definition~\ref{def:rel-rep} may not look surprising to readers familiar with the literature on kernel methods \cite{hofmann2008kernel}. Instead of presenting ${\color{rr}{x'}}$ as a kernel, we say it is a relative representation to stress that  (1) we want to \textit{explicitly} represent the coordinates in our ASIF procedure as opposed to operating in an implicit feature space and (2) we do not aim at learning regression parameters, although we note this may help with the inference speed. Instead, we rely on a simpler procedure that may be interpreted as a hard version of the Watson-Nadaraya~\cite{nadaraya1964estimating,watson1964smooth} regression with a distance threshold. Nevertheless, it is worth noting that while the ASIF procedure hinges on the ability to compute similarities between samples, such computation can be achieved using a kernel function, thus sidestepping the need for explicit representations generated by unimodal encoders.
Although integrating kernel methods could offer benefits, as alluded to earlier, delving into these prospects is beyond the scope of this work. Our central focus remains on illustrating how single-domain pre-trained networks can be aligned without additional training.

\subsection{ASIF: relative representations inducing a meaningful common space.} \ 
Consider the embedding spaces of any two good text and visual encoders,
we expect captions of images that are close in the visual space to be themselves close in the language space. 
This fact makes a representation defined in terms of similarities against a set of ground-truth multimodal pairs almost mode-invariant, i.e. an image and its caption share almost the same representation.

That is why we can assign the best caption to a new image $x^*$ just by performing nearest neighbors in this new space: we can consider the
relative representation of $x^*$ respect to the image collection $(x_1, \dots , x_n)$ \textit{as if} it was the relative representation of its ideal caption $y^*$ with respect to the counterpart collection $(y_1, \dots, y_n)$, see Figure \ref{fig-main}.
The whole procedure to set up an ASIF model and use it to find the best caption for a new image follows.

\begin{mdframed}[
  backgroundcolor=black!10,
  usetwoside=false,
  innermargin=0,
  outermargin=0,
  leftmargin=0,
  rightmargin=0,
  rightline=false,
  bottomline=false,
  leftline=false,
  topline=false
  ]
\textsc{ASIF recipe.} 
Ingredients:
\begin{itemize}
    \item Two good encoders, each mapping a single data modality to a vector space.     Let $X$ and $Y$ be the mode domains, for instance a pixel space and a text space, we need ${\color{ar_i}{E_1}}: X \to {\color{ar_i}{\mathbb{R}^{d1}}}$ and ${\color{ar_t}{E_2}}: Y \to {\color{ar_t}{\mathbb{R}^{d2}}}$.
    \item A collection of ground truth multimodal pairs: $D = \{(x_1, y_1), $ $\dots, (x_n, y_n)\}$, for instance captioned images.
\end{itemize}

Procedure to find the best caption among a set of original ones $\hat{Y} = \{\hat{y}_1, \dots, \hat{y}_c \}$ for a new image $x^*$:
\begin{enumerate}
    \item Compute and store the embeddings of the multimodal dataset $D$ with the encoders ${\color{ar_i}{E_1}}, {\color{ar_t}{E_2}}$ and discard $D$. Now in memory there should be just $D_E = \{({\color{ar_i}{x_1}}, {\color{ar_t}{y_1}}), \dots, ({\color{ar_i}{x_n}}, {\color{ar_t}{y_n}})\}$;
    \item Compute the $n$-dimensional relative representation for each candidate caption ${\color{rr}{\hat{y}_i}} = [\text{sim}({\color{ar_t}{\hat{y}_i}}, {\color{ar_t}{y_1}}), \dots, \text{sim}({\color{ar_t}{\hat{y}_i}}, {\color{ar_t}{y_n}})]$, where $\text{sim}$ is a similarity function, e.g. cosine similarity. Then for each ${\color{rr}{\hat{y}_i}}$ set to zero all dimensions except for the highest $k$, and raise them to $p\geq1$. Finally normalize and store the processed $c$ vectors ${\color{rr}{\hat{y}_i}}$. Choose $k$ and $p$ to taste, in our experiments $k=800$ and $p=8$;
    \item Compute the relative representation of $x^*$ using the other half of the embedded multimodal dataset $D_E$ and repeat the same processing with the chosen $k$ and $p$;
    \item We consider the relative representation of the new image $x^*$ \textit{as if} it was the relative representation of its ideal caption $y^*$, i.e. we define ${\color{rr}{y^*}} \coloneqq {\color{rr}{x^*}}$. So we choose the candidate caption $\hat{y}_i$ most similar to the ideal one, with  $i = \text{argmax}_i(\text{sim}({\color{rr}{y^*}}, {\color{rr}{\hat{y}_i}}))$.
            \end{enumerate}
To assign one of the captions to a different image $x^{**}$ repeat from step 3.
\end{mdframed}

\subsection{Properties of ASIF models.} \ The above construction yields several intriguing properties for free:

\textsc{No training and ``data centric''}.
As we have seen, an ASIF model is built on top of two independently pretrained encoders and the embeddings of a multimodal dataset, and so without training or finetuning any neuron. 
Being deployable or updatable in seconds, an ASIF model is radically ``data centric''~\cite{datacentric}. For example, it is trivial to adjust the model by adding or forgetting specific samples.
The latter use-case is particularly important, as the right to use specific assets may change over time and removing the effect of specific samples from a trained network requires sophisticated forgetting techniques, e.g. \cite{golatkar2021mixed,golatkar2020forgetting,golatkar2020eternal,ginart2019making,guo2020certified}. In our procedure, the encoders should be pre-trained with established data sets that do not change over time, while removing the effect of a multimodal pair is as simple as deleting its embeddings.

\textsc{Data efficiency: }Being able to exploit two pretrained encoders, ASIF models require far less ground-truth multimodal pairs to become effective. As confirmed by our experiments, ASIF reaches competitive zero-shot performance on diverse classification benchmarks by using a fraction of the multimodal data of its predecessors, reaching a respectable accuracy even with thousands of pairs  (we refer to Section \ref{sec-experiments} for more details). This is in line with classical work in computer vision, where prototypical networks~\cite{proto} are a strong baseline in the extremely few-shot regime.

\textsc{Interpretability:} Sparse relative representations make the ASIF models interpretable classifiers. In fact, we can trace back every prediction to a small set of data points in the multimodal dataset  --corresponding to the dimensions that are nonzero in both the image and the label relative representations-- accountable for the outcome (at most $k$), see Figure \ref{fig-deepdive}. This enables visualizations of the relevant samples contributing to a correct or incorrect prediction at no extra cost, in stark contrast with other approaches that are often costly \cite{Achille_2021_CVPR} and fragile~\cite{koh2017understanding,basu2020influence}.

\subsection{Relation to k-Nearest Neighbors.} Like k-NN, ASIF is a non-parametric supervised learning algorithm that requires an explicit representation of every entry in the training dataset at test time. Differently from k-NN, ASIF can perform open-ended classification, as shown e.g. in Figure \ref{fig-main} with two competing brand new captions. Indeed, ASIF is functionally equivalent to CLIP, and can function as a drop-in replacement in applications using CLIP-like models.

\subsection{Implementation that scales.} \  Clearly, our method pays the price of avoiding training with a larger memory footprint and increased computation at inference time, since we need to compute not only the embeddings but also the cosine similarities against the multimodal dataset. 
As such, our approach should not be considered a general one-stop replacement for CLIP, although in our experiments we managed to scale ASIF to 1.6M pairs while maintaining a reasonable inference speed. Our non-optimized implementation of ASIF is less than 2x slower than CLIP.
On a positive note, there are two well-established techniques that could radically enhance the efficiency of ASIF and potentially enable scalability to billions of entries.
The memory footprint required to store all the embeddings of the multimodal dataset can be dramatically reduced using product quantization \cite{product-quantization}, while inverse indexing \cite{inverse-indexing} can be used to circumvent the need for computing the cosine similarities against the entire dataset at test time. These techniques are both implemented e.g. in the FAISS library \cite{faiss}. 
Finally, we find that the distribution of pairs chosen during inference is rather short-tailed, presenting opportunities to massively prune the model even \textit{at deployment time}, deleting from memory the data that is never used. It should be noted, however, that the assessment of the performance of large-scale optimized ASIF models is beyond the scope of this work. While it is an interesting direction for future research, the focus of our current study is on establishing the potential the ASIF method.

\subsection{Design choices and implementation of ASIF models.} 
\textsc{Curating the multimodal dataset.} \ 
While neural models like CLIP or LiT are defined just by the weights of the two encoders, to univocally identify an ASIF model we also need the embeddings of the multimodal dataset.
\highlight{Even if two image-text ASIF models rely on the very same encoders, they comply to different visual understandings of the world if they are based on different collections of image-text pairs, therefore achieving different performances on the same tasks.}
Unlike conventional neural vision models, ASIF enables effective curation of the training dataset through swift iterations, given the absence of training and the smaller datasets. Furthermore, ASIF provides the means to assess the impact of each training pair, as demonstrated in Figure \ref{fig-deepdive}.

\textsc{Salient hyperparameters.} \ 
While using the raw relative representations already produces an ASIF multimodal model with non-trivial capabilities, we found that two simple treatments greatly improve performance and efficiency, and also foster interpretability.
\\
\begin{itemize}
    \item \textit{Sparsification.} We set to 0 all the entries of the $n$-dimensional relative representation except for the top $k$. In our experiments $n$ and $k$ are respectively in the order of millions and hundreds. In this way we cut off the small noisy signals from the dissimilar entries, that accumulated during comparisons would destroy the signal from the few similar entries. Furthermore we get highly interpretable representations that can be efficiently compared, since we have just $k$ nonzero features, each one linked to a single entry in the multimodal dataset.
    \item \textit{Exponentiation.} We raise all the nonzeroed similarities $\text{sim}({\color{ar_i}{x'}}, {\color{ar_i}{x_i}})$ to $p$, with $p \geq 1$. This non-linearity weighs more the contribution of the most similar entries in the relative representation.
\\
Besides the pivotal choice of the ground-truth multimodal pairs, the number of non-zero elements $k$ and the exponent $p$ are the salient hyperparameters to consider when deploying an ASIF model. In general, we found that picking a $p \neq 1$ may help, while choosing a $k \ll n$ is always crucial.  
For more details see
Section \ref{sec-experiments}.
\end{itemize}

\section{Related Works}

\textsc{Classics.}\ In \cite{rota}, Stanley Ulam affirms that a mathematical formalization of the word ``as''--on a par with the connectives ``and'', ``or'', ``implies'' and ``not''--would be a key milestone to artificial intelligence. This idea of analogies being at the core of cognition is shared by  \citet{hofstadter2001analogy}, who states that a concept is a collection of analogies, in line with what the ASIF procedure prescribes.

\textsc{Retrieval augmented foundation models.}  
Recent works in NLP enhance unimodal language models with retrieval to reduce the size of the architectures and the training dataset while also making results more transparent \cite{retro, atlas}. Our work is in line with this view, that the ASIF procedure extends to the multimodal case. Importantly, ASIF offers a new perspective on data centric AI~\cite{datacentric}, where data and the external memory implement the alignment between modalities. Networks with discrete Key-Value bottlenecks~\cite{trauble2022discrete} are closely related to our approach, with the critical differences that our memory is not learned and that their decoder is trained for classification. 
Retrieval and memory-augmented models have also been successful in Reinforcement Learning~\cite{goyal2022retrieval}, physical reasoning~\cite{alias2021neural}, and code generation~\cite{liu2021retrieval}.
Finally, we notice that making predictions on new samples by exploiting the similarity with a dictionary of previous data points is a common approach in computer vision~\cite{proto} for few-shot learning. Our procedure is also related to compressed sensing algorithms where a signal (in our case an image) is sensed as a sparse combination of fixed atoms~\cite{wang2012generalized,mallat1993matching} with an iterative projection procedure~\cite{locatello2017unified,locatello2018matching} and only transmitting the coefficients to the receiver (text modality).

\begin{table*}[ht]
\vskip 0.15in
\begin{center}
\begin{small}
\begin{sc}
\begin{tabular}{lccccc}
\toprule
& \textsc{Dataset} & \textsc{ImgNet} & \textsc{CIFAR} & \textsc{Oxford} & \textsc{ImgNet} \\
\textsc{Method} & \textsc{Size} & & \textsc{100} & \textsc{Pets} & \textsc{v2} \\
\midrule
CLIP{\tiny ~\cite{clip}}        & 400M (priv)                 & 68.6                       & 68.7                       & 88.9                   & -                             \\
CLIP{\tiny ~\cite{clip}}        & 15M (pub)                   & 31.3                       & -                          & -                      & -                             \\
LiT{\tiny ~\cite{lit}}         & 10M (pub)                   & 66.9                       & -                          & -                      & -                             \\
CLIP {\tiny ~\citealt[\texttt{uu}]{lit}}        & 901M (priv)                 & 50.6                       & 47.9                       & 70.3                   & 43.3                          \\
LiT{\tiny ~\cite{lit}}          & 901M (priv)                 & 70.1                       & 70.9                       & 88.1                   & 61.7                          \\
\midrule
ASIF {\tiny(sup vis. encoder)}                    & 1.6M (pub)                  & 60.9$^*$                      & 50.2                       & 81.5                   & 52.2  \\
ASIF {\tiny(unsup vis. encoder)}                    & 1.6M (pub)                  & 53.0$^*$                      & 46.5                       & 74.7                   & 45.9  \\
\bottomrule
\end{tabular}
\end{sc}
\end{small}
\end{center}
\caption[Zero shot classification accuracy of different multimodal designs]{\textsc{Zero shot classification accuracy of different multimodal designs.} CLIP and LiT implementations vary by dataset and the visual transformer used as image encoder. The first CLIP and LiT entries use a VITb16 as ASIF, the last CLIP and LiT entries use a VITb32 (larger patch size). The public dataset of CLIP is a curated subset of YFCC100m~\cite{thomee2016yfcc100m}, while LiT and ASIF use CC12M. \tiny{$^*$We used a subset of the ImageNet validation set to tune the two hyperparameters of ASIF which were then used on the other data sets. The number reported in the table is a test set. When tuning on different datasets, accuracies stay consistent, see the appendix.}
}
\label{tab-main}
\end{table*}

\textsc{Learning multimodal models.} \ 
Following the intuition outlined by early works on aligning text and image embeddings \cite{earlyfrome2013devise, earlykarpathy2015deep},
today large multimodal models are conquering the computer vision scene thanks to their wide applicability and easy transfer to new downstream tasks \cite{clip, lit, align, coca}. 
\\
We identify two key leaps respect to traditional models like ResNet \cite{He2015DeepRecognition}: (i) Free text allows to learn visual concepts beyond a finite set of predefined categories and to exploit the structure of language to compose them, as masterfully seen in Dall-E \cite{dalle2}. (ii) The recognition tag transitioned from being an output pulled out of the image by the neural stack (the label) to become an input that should be interpreted, and therefore processed by its own encoder (the free text caption). 
This corresponds to an epistemological perspective shift, as we discussed in chapter \ref{ch:explanatory}. 
Data and learning efficiency are clear challenges for large multimodal models, that often require hundreds of millions of examples. Efforts such as~\cite{lit, atlas} attempt to reduce this. ASIF presents a different take on this problem, showing how much can be achieved by simply remembering the training data efficiently.

\section{Empirical Evidence}
\label{sec-experiments}
In the following we compare ASIF to traditional multimodal models based on contrastive training, CLIP and LiT. We then take a closer look to the classification of a single image, unpacking the relative representations and following the classification algorithm step by step.
As a prelude to the above, we start by discussing the pretrained encoders and dataset forming the ASIF models we tested.

\subsection{Pretrained encoders and multimodal dataset used.}  For our primary experiment, we utilized vision transformers as image encoders, pretrained either in a supervised manner (DEIT base, \cite{deit}) or in an unsupervised manner (DINO VITs8, \cite{dino}), on Imagenet 1k \cite{deng2009imagenet} and 21k \cite{imagenet21k} respectively. 
The embedding size was 768 for DEIT and 384 for DINO. This configuration aligns with that used in LiT ~\cite{lit}, with the sole distinction being that, unlike LiT, we used a pre-trained, frozen text encoder. Regarding the text encoder, we employed the SentenceT transformer \cite{sentencet}, trained on a dataset comprising more than 1 billion sentences obtained from the internet. 
We employed the first 1.6M  entries of the Conceptual Caption dataset (CC12M, \cite{cc12m}) as our multimodal dataset. This dataset amasses images and filtered alt-texts collected from the internet. To optimize performance on a single Tesla T4 GPU, we limited our analysis to the initial 1.6M pairs. In our ``scaling-laws'' experiments, we also utilized DEIT tiny and small vision transformers \cite{deit} and two smaller SentenceT encoders.

\subsection{Zero-shot performance.} \
We assessed the quality of our ASIF multimodal model by comparing its zero-shot classification performance against CLIP and LiT on four datasets: CIFAR100, Imagenet, Imagenetv2, and PETS \cite{ deng2009imagenet, cifar, imagenetv2, pets}; see Table \ref{tab-main}. 
We crafted label prompts as in \citet[Table 11]{lit}.

Remarkably, we achieve competitive performance with CLIP and LiT using two frozen pretrained encoders and a fraction of the image-text pairs. 

The two ASIF models reported in Table \ref{tab-main} differ
for the vision encoder, that is respectively supervisedly (DEIT) and unsupervisedly pretrained (DINO). We tuned $k$ and $p$ on the ImageNet validation set, in both cases we used $k=800$ and $p=8$.
The comparison between the two versions is not completely fair since the visual transformer architecture of DINO is smaller (e.g. the codings are 384-dimensional rather than 768) but corroborates the effectiveness of ASIF with encoders pretrained using no supervision. In the appendix we report the results of a wider collection of ASIF models based on different visual and textual backbones.

\textsc{Summary:} Overall, we observe that our ASIF procedure can achieve competitive zero-shot results with a fraction of the image-text pairs used in prior work (assuming access to other pre-trained, even unsupervisedly, unimodal models).

\begin{figure}[h]
\centering
  \begin{overpic}
		[trim=0cm 0cm 0cm 0.55cm,clip,width=0.999\linewidth]{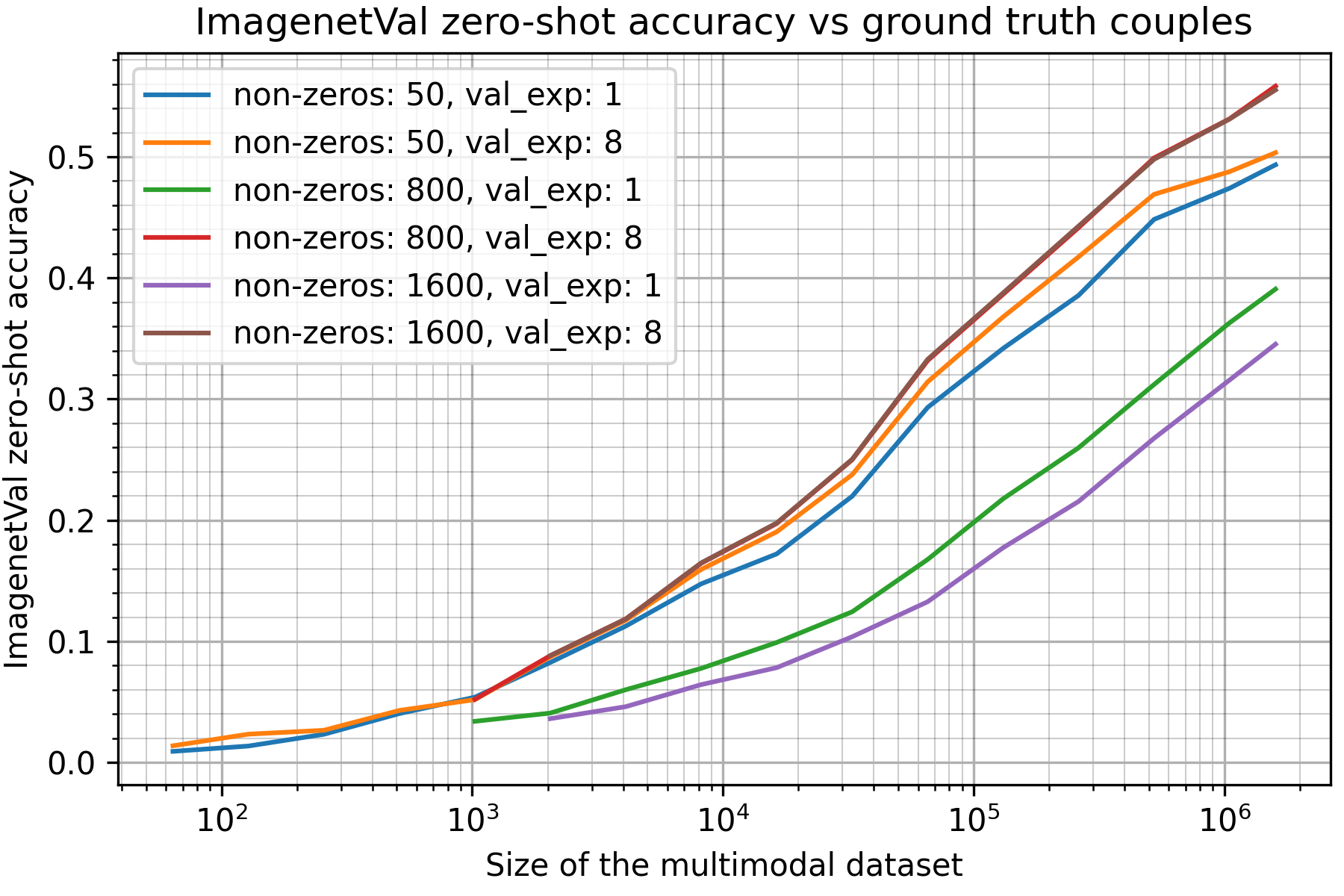}
		\end{overpic}

		\caption[ASIF is a learning algorithm]{\textsc{ASIF is a learning algorithm:} Imagenet accuracy improves smoothly as we increase the size of the multimodal dataset. Colors show the impact of \textit{k} and \textit{p} (non-zeros, val exp).}\label{fig-hyper}
\end{figure}

\subsection{ASIF scaling laws.} In Figure \ref{fig-hyper} we show the full zero-shot accuracy trend on Imagenet as the size of the multimodal dataset grows for different choices of $k$ and $p$. ASIF models become effective  very quickly: we 
reach a non-trivial $18\%$ classification accuracy using just the first 10,000 pairs in the multimodal dataset. 
We recall that building an ASIF model from scratch still requires a lot of unimodal data to pretrain the encoders, but now this data may come untied and even without any label.

We tested ASIF further with smaller image and text encoders on multimodal datasets of different sizes ($10^2$ to $10^6$ image-text pairs from CC12M) to provide early evidence about ASIF scaling laws. We used DEIT tiny, small, and base vision transformers \cite{deit}, along with two smaller SentenceT encoders. As we see in Figure \ref{fig-hyper} and in the appendix, Imagenet classification accuracy grows smoothly with the size of the multimodal dataset, and performance does not saturate earlier with smaller encoders. These results are promising but still preliminary, further experiments with larger multimodal datasets are left for future work. 

\textsc{Summary:} Our experiments show a steady improvement in ASIF's performance as the size of the multimodal dataset increases. Performance deteriorates with smaller encoders, but even then there is no sign of saturation or plateau.

\begin{figure}[t]
    \centering
        \begin{overpic}[trim=0cm 0cm 0cm 0cm,clip,width=1.05\linewidth]{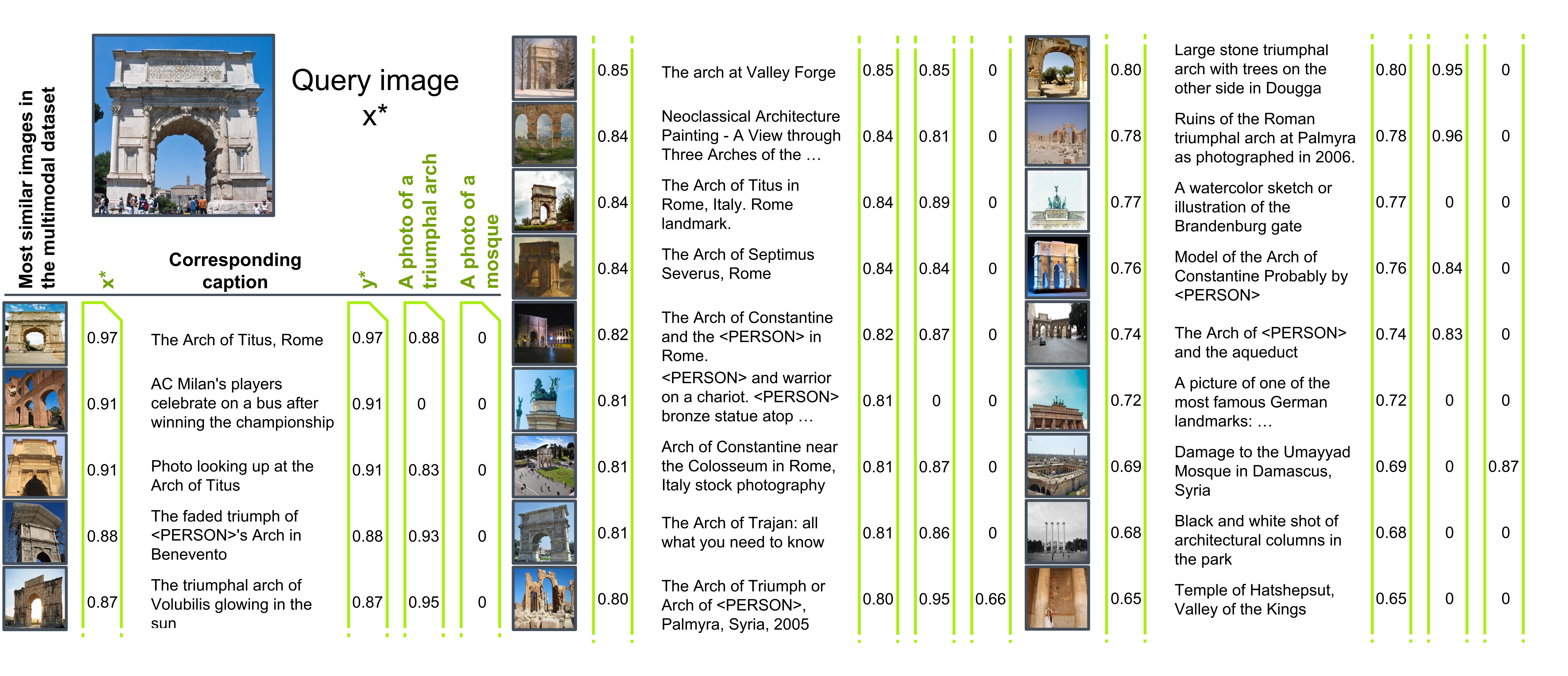}
                                    \end{overpic}
    
        \caption[ASIF representations are interpretable and amendable]{\textsc{ASIF representations are interpretable and amendable.} Thorough analysis of the relative representations --the four vectors in green-- produced by ASIF to assign the best caption to a given image $x^*$. Every dimension corresponds to the similarity of the input image (text) to a unique image (text) entry in the multimodal dataset. We can visualize the training samples leading to correct or incorrect predictions, which is useful to curate better data. For example, the second training sample is broken, we can remove it and produce an updated ASIF model in seconds. 
        }
    \label{fig-deepdive}
    
    \end{figure}

\subsection{Adjusting an ASIF model in seconds.} Consider classifying EuroSAT \cite{helber2019eurosat} satellite images using ASIF. Initially, the zero-shot performance outperforms random chance but is not impressive  ($29.4\%$ unsup. configuration, 10 classes). CLIP, while better, also falls short with a $54.1\%$ performance rate. 

Now, imagine we acquire 100 new image-text pairs from EuroSAT. Our goal is to develop a new multimodal model based on an updated training set, anticipating the need to process more satellite images in the future. With CLIP, this would require us to fine-tune or retrain the model from scratch using the updated dataset.
In contrast, ASIF requires only to obtain and store the new pairs' embeddings. The model retains its usability for all the previous images, but it now also makes accurate predictions for satellite images, achieving an $82.2\% \pm 2.0$  accuracy on EuroSAT.\footnote{Why does CLIP perform better zero-shot? Likely, its 400M pairs have more samples close to satellite images than ASIF's 1.6M samples from CC12M, evidenced by ASIF's drastic improvement with just 100 new samples.}

Similarly, if we need to remove samples from the multimodal training dataset—either because they are faulty (as shown in Figure \ref{fig-deepdive}) or because we have lost the license to use them—the process is as simple as deleting the corresponding embeddings to get a new model.

\textsc{Summary:} 
ASIF enables quick model adjustments and data handling without the need for retraining, unlike traditional models such as CLIP. This demonstrates its practicality in real-world scenarios, such as the emergence of a new setting or the loss of rights for assets used during training.

\subsection{Deep dive into a classification.} To better understand why and how the ASIF procedure works, we are going to follow step by step the single classification depicted in Figure \ref{fig-deepdive}, showing the entries of the multimodal dataset used to build the relative representations. For simplicity we assume $k=23$.

We want to find the Imagenet label of the upper left image in Figure \ref{fig-deepdive}. The first step is to compute its relative representation with respect to the multimodal dataset, this procedure selects the 23 most similar images in our collection. The most similar are indeed triumphal archs that--as similarity decreases--turn to generic structures with archs or ancient monuments. No other image in the multimodal dataset is involved in the classification, we could replace all the entries in our collection except for these 23, and as long as no new image is more similar to the input image, every computation in the classification process would remain the same. This consistency of the ASIF procedure is in contrast to traditional approaches like CLIP and LiT, where the parameters defining the encoder computations are shaped by the whole multimodal dataset through contrastive learning.

Now, we approach \highlight{the pivotal step of the ASIF procedure: jumping into the text space by treating the relative representation of the input image \textit{as if} it were the relative representation of its ideal caption. The vector remains the same, but its meaning changed: now it signifies how much the ideal caption should be similar to 23 captions.} 

The efficacy of this jump hinges entirely on the quality of the multimodal dataset. Looking at the 23 captions in Figure \ref{fig-deepdive}, we are confident that the most fitting candidate caption corresponds to one of the prompts associated with the \texttt{triumphal\_arch} label in ImageNet, such as \textit{``a photo of a triumphal arch''}.
Conversely, a dataset featuring 23 non-informative captions, such as file names \textit{``IMG\_20180823.jpg''} or camera settings \textit{``D90 18.0-70.0 mm f/3.5-4.5''}, would not allow the model to recognize the correct class even with the best image and text encoders. 
This limitation is actually a defining feature of the ASIF procedure, as \highlight{the meaning attributed to the input is ultimately determined by the multimodal dataset and not by the encoders}: by acting just on the image-text pairs, we have full control over the model's output.

\textsc{Summary:} Our simple (non-cherry picked) example showcases how the ASIF predictions can be easily attributed to specific examples in the multimodal training data by construction. This feedback can be used to explain predictions and grow high quality datasets in data centric AI, for example by inspecting which examples contribute to incorrect classifications.

\section{Discussion}
The effectiveness of the ASIF procedure raises questions on the role of memory and retrieval in machine learning, while at the same time opens new opportunities for products based on multimodal models, opening many avenues for future works. In the following we will discuss these aspects.

\subsection{Perception and interpretation disentangled.} \ 
In ASIF there is no trace of the multimodal data in the weights of the encoders, which are pretrained on different unimodal datasets.
Nonetheless, the relative representations and the outcome of any classification task fundamentally depend on the multimodal dataset.
This state of affairs reflects \highlight{the factorization of perception and interpretation in the two stages constituting an ASIF model}; the encoding and the construction of the relative representations.
Such factorization \highlight{is desirable because it relieves the black-box neural encoders from the responsibility of attributing meaning to their inputs}, as envisaged by \citet[Par. 3.3.1.3]{eco-kant}. \highlight{Therefore, we can consider the neural encoders as mere sensors and focus only on the second stage to analyze, explain, and act on the interpretations of an ASIF model}, as shown in Figure \ref{fig-deepdive}.

\subsection{Learning or retrieval?} \ 
As we have seen, the ASIF procedure requires no training: it does not distill the multimodal data into any learnable parameter. Rather, it prescribes a rigid memorization of the embeddings of the multimodal dataset, 
where each entry has its fixed-size spot, similarly to a retrieval process.
On the other hand it seems impossible to not describe ASIF as a learning algorithm; for instance it satisfies the fundamental characterization given by Mitchell \cite{mitchell1997machine}:
the more the multimodal data the more ASIF improves, as we can clearly see in Figure \ref{fig-hyper}. Ultimately, an ASIF model is functionally comparable to CLIP.
ASIF blurs the border between learning and retrieval by questioning the effectiveness of storing information only in the weights, 
and rather advocates to combine learning representations with external memories. We encourage more work on memory augmented neural networks and towards understanding the implications of memory for generalization.

\subsection{Generalization to new distributions.} \ The empirical performance of ASIF calls for a discussion on zero-shot and out-of-distribution generalization in foundation models trained with huge data sets. Clearly, the performance of ASIF will depend strongly on the multimodal data used for alignment. As an example, the good performance on Imagenet may not be particularly surprising in light of the qualitative evaluation seen in Figure~\ref{fig-deepdive}. There, our query image might as well had been part of our multimodal data set, as the semantic gap with its neighbours appears to be small. Despite this, our choice of downstream evaluation and pre-training data set is identical compared to prior work~\cite{lit}. As such, while it appears clear that ASIF should fail when the semantic gap between downstream task and ``training data'' is large, it is unclear why it should be different for more standard models like CLIP~\cite{clip} and LiT~\cite{lit}: if a gap does exist, future work should work to address it. In the meanwhile, we recommend that future foundation models are benchmarked on significantly broader sets of downstream tasks, ideally with some analysis of the semantic distance between test and training data (if any). Alternatively, our results may be interpreted in light of the strong performance of unimodal models. There may be a limited benefit of training from scratch on less curated multimodal data sets compared to standard and well established unimodal data sets, although we posit that at even larger scales the difference may be more significant.

\subsection{Limitations.} \
The simple ASIF procedure presented here offers a strong baseline for multimodal models, but its performance still falls apart to CLIP and LiT when the mutimodal dataset is abundant and the cost of training is not a concern. Additionally, the large dimensionality of the relative representations, even if sparse, poses challenges for directly applying ASIF to tasks like text-to-image generation. We recognize that the experiments reported in this work do not provide a comprehensive examination of the myriad of downstream tasks multimodal models are known to adeptly handle, it is important to note that such broad coverage was explicitly out of scope for this work. Our primary objective here was to introduce and justify the ASIF procedure, illustrating its effectiveness on the representative task of zero-shot classification. In making this choice we followed \cite{lit}, that used the very same datasets to showcase the benefits of a locked image encoder, that is their main claim. We anticipate and welcome a more extensive evaluation of ASIF in the context of a wider range of tasks in future research endeavors.

\section{Conclusions.} \ 
We presented a simple procedure called ASIF to assemble a fully functional multimodal model like CLIP from two unimodal pretrained encoders and a collection of image-text pairs without tuning a single neuron. While achieving competitive zero-shot classification results with its predecessors using a fraction of the data, the proposed ASIF procedure enriches multimodal models with editability--a new model based on different pairs can be deployed in seconds--and interpretable representations.
The effectiveness of ASIF models also clashes with the dominant view of a learning algorithm as a way to distill data into the parameters of a model, and raises questions on the role of memory and retrieval in machine learning.

%% file: Chapters/Chapter05.tex
\newcommand{\Square}{\square}
\newcommand{\CheckedBox}{\checkmark\hspace{-8pt}\square}

\chapter{Are Large Language Models Sparks of Artificial Scientists?}
\label{ch:llms} \chaptermark{Are Large Language Models Sparks of Artificial Scientists?}

\section*{Chapter abstract} Large Language Models (LLMs) such as GPT \cite{gpt2020}, Claude \cite{claude2023}, Palm \cite{chowdhery2022palm}, and LLaMA \cite{touvron2023llama} have been hailed as sparks of Artificial General Intelligence \cite{bubeck2023sparks}, leading us to envision them as preliminary versions of the artificial scientists discussed throughout this thesis. Their performance has been nothing short of remarkable, showing a wide range of capabilities--advanced reasoning tasks, including code and math, question answering, translation and multilingual proficiency, summarization--and demonstrating human-level performance on various professional and academic benchmarks \cite{openai2023gpt4}. However, it is crucial to acknowledge that, as of the current state of affairs, these models are prone to hallucination \citep{hallucination}, fundamentally incapable of recognizing the limits of their knowledge, and exhibit traits that are inherently inconsistent with scientific practice. While an exhaustive exploration of these issues could well warrant a separate Ph.D. thesis in the years to come, our aim in this chapter is to identify these challenges, and discuss LLMs eligibility as early versions of artificial scientists. Finally, we will also test several LLMs on a task derived from Odeen, the environment we used to simulate the scientific endeavor introduced in chapter \ref{ch:explanatory}.  

\newpage
\section{The tension between truth and LLMs functioning}

\begin{flushright}{\slshape    
    Bullshit is a greater enemy of the truth than lies are.} \\ \medskip
    --- \defcitealias{frankfurt2005bullshit}{Harry}\citetalias{frankfurt2005bullshit} \citet{frankfurt2005bullshit}
\end{flushright}

\begin{flushright}{\slshape    
    To copy the truth can be a good thing, but to invent the truth is better, much better.} \\ \medskip
    --- 
    Giuseppe Verdi
\end{flushright}

Despite their striking semblance to a human assistant, endowed with vast knowledge and exceptional mastery over language, Large Language Models (LLMs) have been labeled as “stochastic parrots” \citep{parrots} or prolific producers of bullshit \citep{bullshit}. This characterization stems from their frequent tendency to generate entirely fabricated information, be it incorrect sports results, historical inaccuracies, or references to nonexistent works. Such behavior is utterly unacceptable for a machine aspiring to embody the role of an artificial scientist. This was demonstrated by the fate of Galactica, the Large Language Model (LLM) built by Meta to assist scientists, which was retired after only three days \cite{heaven2022meta}. We are thus compelled to ask: is this a flaw inherent to the early iterations of LLMs, one that could be remedied by advancing along our current trajectory of increased computational power and more extensive data? Or, alternatively, does their fundamental functioning harbor an incompatibility with the relentless pursuit of truth that moves a genuine scientist?

\subsection{Do LLMs just need more training data?}

The empirical relationship between training data and performance in Large Language Models (LLMs) reveals a power-law trend, suggesting that more data generally leads to better performance \citep{kaplan2020scaling}. However, it’s crucial to note that the performance is measured in terms of test loss, which may not adequately represent an LLM’s capability of genuine understanding or generation of scientific knowledge.

To shed light on this issue, Timothy B. Lee’s thought experiment with a college physics major offers a valuable perspective \cite{lee2012trainingdata}. Imagine charting a student’s performance on physics tests against the number of textbook pages she has read. Throughout her undergraduate studies, you would likely observe a consistent relationship: as she reads more, her performance improves. Extrapolating this trend would lead you to predict a continuous growth in her understanding, transforming her into a physics prodigy greater of Einstein and Fermi.

However, a critical transition occurs as she enters a Ph.D. program and begins to grapple with the frontiers of physics knowledge. Her learning curve, once steep and promising, starts to flatten. It’s not that she has run out of textbooks to read; she has run out of new knowledge that can be readily assimilated from them. She now faces the challenge of generating original insights, stepping into territories uncharted by her predecessors.

Applying this analogy to LLMs, we can envision a scenario where, despite achieving human-level understanding across diverse topics, providing them with more training data ceases to be beneficial. The LLMs, akin to the physics student, might find themselves swimming in a sea of words that merely rephrase what they already know. The bottleneck would not be the quantity of data; but the quality and novelty of the knowledge it can provide. For instance, if the data is sufficient to craft a world model useful to properly simulate new circumstances, as we will discuss in the next section.

\subsection{Reality check}

When scientists, journalists, or anyone craft their narratives about reality, they are often confronted with potential continuations of their texts that, while seemingly fitting and satisfying from a storytelling perspective, are then discarded due to their deviation from the truth. Humans have an internal mechanism that signals this kind of deviation, and can use it to preserve the integrity of their work and their alignment with reality.

LLMs lack this feedback, and instead keep that continuation. 

This mental experiment materializes a popular belief about a fundamental limitation of LLMs: they are missing an actionable model of the world, where statements can be simulated, leading eventually to a signal when they clash with the rules of their world model; a circumstance that is independent of the soundness.
Indeed, the very notion of truth as the property of being in accord with reality \cite{merriamwebster2005} seems inaccessible for computational systems that have never had any direct experience with the real world outside of text, as noted by Douglas \citet{hofstadter2023}.

However, it is important to acknowledge evidence pointing towards the capability of neural network-based AI models, to build and effectively use a world model. One seminal work in this domain is "World Models" by \citet{ha2018world}. In their research, they developed generative neural network models of popular reinforcement learning environments, creating a compressed spatial and temporal representation of the environment. Remarkably, they demonstrated that an agent could be trained entirely within a dream environment generated by its world model, and the learned policy could then be successfully transferred back to the actual environment.

Moving specifically to LLMs, a very recent work by Kenneth Li et al. titled "Emergent World Representations" \cite{li2023othello} showcases the presence of an Othello world model in LLMs fine-tuned on Othello moves. In this study, a variant of the GPT model was applied to predict legal moves in the board game Othello. The research uncovered evidence of an emergent nonlinear internal representation of the board state, showing the ability to control the output of the network by making interventions on this state.

In light of these findings, we can posit that while current LLMs do lack a comprehensive model of the world, the excessive reliance on surface statistics seems to be a consequence of their experience limited to text, rather than a fundamental incapacity to build and use world models. 

However, the mentioned works also highlight a significant challenge. The Othello world model developed by Li was derived from text representations of game moves, and also the Odeen task, with its reliance on emojis for creating real-world observations, operates within the bounds of standard UTF-8 text encoding. Developing a comprehensive model of the world seems to necessitate the integration of data modalities beyond text.

Addressing this need represents one of the prevailing open research questions in AI today. While some works propose that the predict-the-next-token paradigm of LLMs could be extended to seamlessly incorporate image tokens alongside text tokens  \cite{peng2023kosmos2, alayrac2022flamingo}, the ability to effectively train on video data appears to demand a new and as yet undiscovered breakthrough \cite{lecun2023}.

\subsection{LLMs uncritically accept every input sample}

We have discussed that more data for LLMs is not always useful, but can it be harmful? How do LLMs behave when new training data contradicts old, sedimented information? Is making an LLM believe a falsehood as simple as stating the falsehood in its training data? 

These questions highlight another fundamental issue of LLMs: they uncritically accept every single training sample, meaning that all training samples have the same right to alter their parameters. It is actually worse than this, since data points that starkly contrast with the model’s established beliefs result in larger, stronger weight updates; in other words, contradicting data points are taken as a higher truth than agreement, in the words of Andrew \citet{trask2023}.

This reveals another fundamental incompatibility with the idea of utilizing Large Language Models (LLMs) as foundational elements for an artificial scientist. Scientists approach observations with a degree of skepticism, meticulously examining surprising or unexpected data before accepting or refuting it. As Karl Popper said, observations are theory-laden, and it is the duty of a scientist to critically assess them.

\subsection{LLMs are not able to assess what they do not know}

Here we discuss the incapacity of LLMs to assess their ignorance.

Large Language Models (LLMs) have demonstrated numerous emergent capabilities through standard autoregressive training; however, ignorance awareness—the ability to recognize and acknowledge knowledge limitations—appears notably absent from this repertoire. While research in this area is nascent, its significance for developing artificial scientific agents warrants attention, even though empirical findings remain limited. Some researchers have hypothesized that explicitly incorporating ignorance statements (e.g., "I do not know") into training data might adversely affect model performance. The concern is that frequent exposure to such statements during training could lead to their overuse, potentially resulting in an overly cautious model that, while not harmful, would be functionally limited.

Building on this, it has been hypothesized that assessing ignorance is a cognitive capacity that requires more than one LLM to be modeled. Specifically, this ability may require two steps: first, two or more continuations are produced, and then the alternatives are compared and analyzed before producing the final answer. For instance, if multiple continuations are recognized as not coherent with each other, then the final answer would be “I do not know”. Importantly, these steps may well be performed by multiple independent calls to one or more LLMs, but not with a single call.

Linking back to the broader implications of this discussion, the profound unawareness of its ignorance places the LLM in a precarious position, particularly when viewed through the lens of an artificial scientist. This unawareness not only impairs the LLM's ability to accurately assess situations and provide informed responses but also diminishes the fundamental motive that underlies the scientific endeavor—the desire to know and understand. Without a mechanism to recognize and admit ignorance, the LLM misses out on the critical reflexive process that drives inquiry and knowledge advancement.

\section{An experiment: LLMs tested on Odeen}

The final contribution of our analysis of LLMs in the context of artificial scientific discovery involves an experimental test of LLMs within Odeen, the environment we introduced in Chapter \ref{ch:explanatory}, serving as a miniature model of the scientific endeavor. These experiments took place within the Big-Bench project \cite{srivastava2023beyond}, a collaborative effort aimed at developing a diverse and challenging set of tasks to assess the performance of Large Language Models. Renowned LLMs and human raters participated in these tasks, with the results being documented in the Big-Bench paper.

\blfootnote{This section is partially based on the paper \textit{"Beyond the Imitation Game: Quantifying and extrapolating the capabilities of language models"}, by 450 authors including Antonio Norelli, Andrea Santilli, Giorgio Mariani, Luca Moschella, Giambattista Parascandolo, Simone Melzi, and Emanuele Rodolà.}

Our Big-Bench submission was a multiple-choice questionnaire, \\
named the Symbol Interpretation Task (SIT) and based on Odeen. SIT was selected as one of the 24 tasks out of 204 that were also tested on humans. Remarkably, our task demonstrated the largest performance gap between human participants and LLMs: all tested LLMs, including Palm and GPT-3, did not perform better than random chance ($20\%$), whereas the average human rater achieved a score of $36\%$, with the top raters even reaching a perfect score of $100\%$.

\begin{figure}[h]
    \centering
    \includegraphics[width=0.99\linewidth]{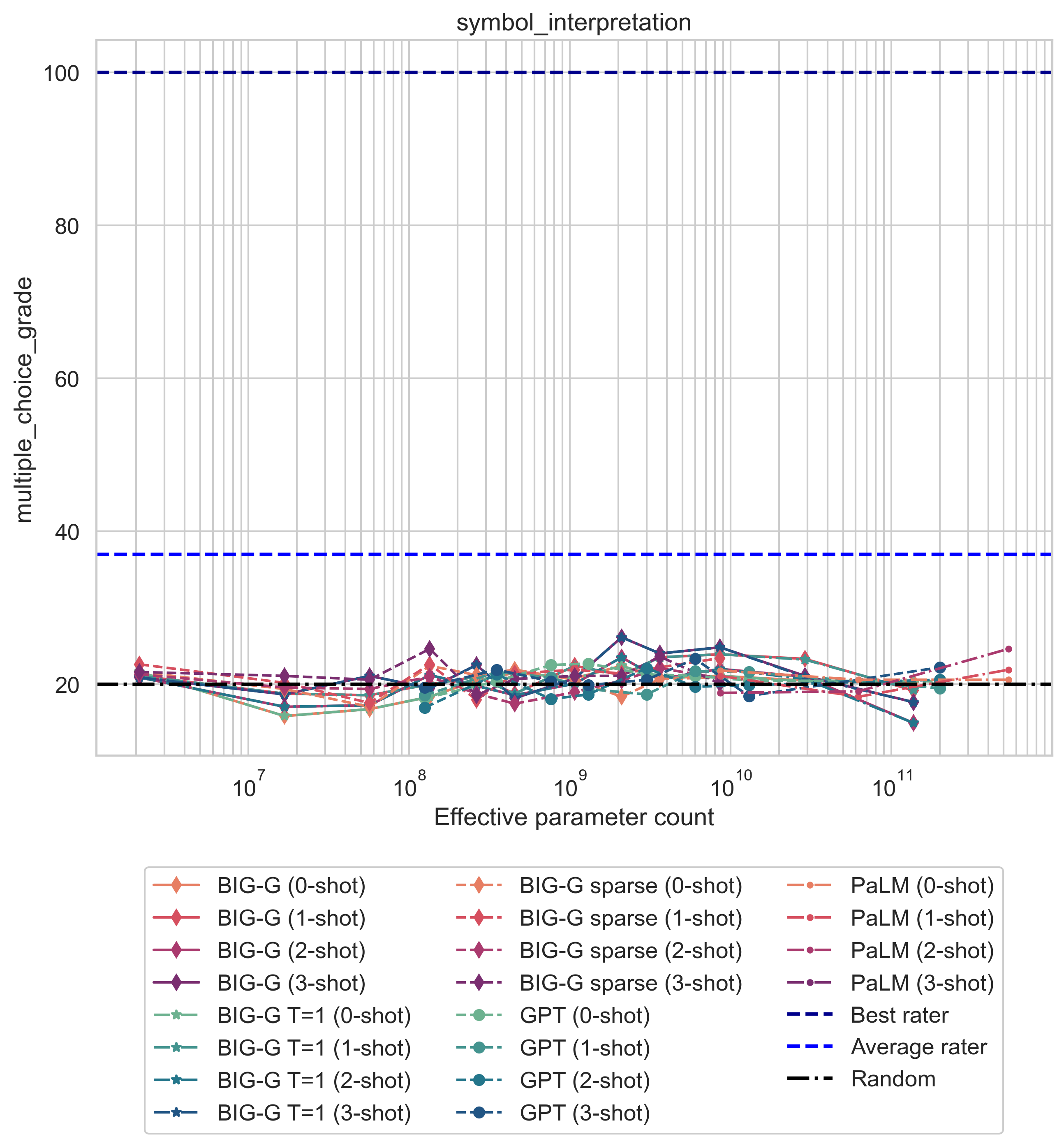}
    \caption[LLMs performance on our Symbol Interpretation Task]{Comparison of multiple-choice grade performance across various models with different effective parameter counts on the symbol\_interpretation task within the Big-Bench project. While renowned LLMs, including BIG-G, BIG-G sparse, PaLM, and GPT, hover around the baseline of random chance ($20\%$), human raters exhibit a considerably superior understanding, with an average score of $36\%$ and top performers achieving a perfect score.}
    \label{fig:zendo-sit}
\end{figure}

In the sections to follow, we will provide a concise overview of the Big-Bench initiative, introduce our Symbol Interpretation Task grounded in Odeen, and share the performance results of both human raters and various LLMs on this task in Figure \ref{fig:zendo-sit} and Table \ref{tab:zendo-sit}.

\subsection{Big-Bench}
Big Bench serves as a comprehensive benchmark to evaluate and understand the capabilities and limitations of contemporary language models \cite{srivastava2023beyond}. Developed collaboratively by 450 authors from 132 institutions, including myself, the benchmark consists of 204 diverse tasks covering areas such as linguistics, math, common-sense reasoning, science, and more, aiming to challenge models with problems believed to be beyond their current abilities.

This project puts various models to the test, including OpenAI's GPT models, dense transformer architectures, and Switch-style sparse transformers, with sizes ranging from millions to hundreds of billions of parameters. A team of human expert raters also participated, providing a robust baseline for comparison.

Key findings highlight that while model performance and calibration improve with scale, they remain poor in absolute terms and when compared to human performance. The models' performance was surprisingly consistent across different architectures, though sparse models showed some advantages. Tasks involving substantial knowledge or memorization showed gradual and predictable improvement, while those requiring multiple steps or components, or relying on brittle metrics, displayed breakthroughs at specific scales. 

\subsection{The Symbol Interpretation Task (SIT)}
\label{sec:SIT}
The Symbol Interpretation Task (SIT) is a particular challenge derived from the Odeen problem, which we have detailed in Chapter \ref{sec:Odeen}. This task necessitates that models choose a sentence that is congruent with two given structures, each represented by a sequence of six emojis. An example follows:

\bigskip\noindent
\textit{In the SIT-plain world, a structure is a sequence of six emojis. Below are the emojis used, along with their descriptions:}
\begin{itemize}
    \item \includegraphics[height=1em]{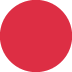} is a red circle;
    \item \includegraphics[height=1em]{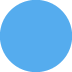} is a blue circle;
    \item \includegraphics[height=1em]{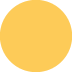} is a yellow circle;
    \item \includegraphics[height=1em]{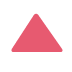} is a red triangle pointing up;
    \item \includegraphics[height=1em]{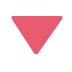} is a red triangle pointing down;
    \item \includegraphics[height=1em]{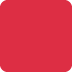} is a red square;
    \item \includegraphics[height=1em]{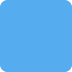} is a blue square;
    \item \includegraphics[height=1em]{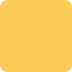} is a yellow square;
    \item \texttt{\_} is an empty space.
\end{itemize}

\bigskip\noindent
\textit{Choose the sentence consistent with the structure} \includegraphics[height=1em]{emoji-images/yellow-square.png} \includegraphics[height=1em]{emoji-images/red-square.png} \includegraphics[height=1em]{emoji-images/yellow-square.png} \includegraphics[height=1em]{emoji-images/yellow-square.png} \includegraphics[height=1em]{emoji-images/yellow-circle.png} \includegraphics[height=1em]{emoji-images/blue-circle.png} \textit{and not consistent with the structure} \includegraphics[height=1em]{emoji-images/blue-square.png} \includegraphics[height=1em]{emoji-images/red-square.png} \includegraphics[height=1em]{emoji-images/red-triangle-up.png} \includegraphics[height=1em]{emoji-images/red-square.png} \texttt{\_} \includegraphics[height=1em]{emoji-images/yellow-circle.png}\textit{:}

\begin{itemize}
    \item[$\Square$] There are zero yellow pieces.
    \item[$\Square$] There is exactly one blue piece.
    \item[$\Square$] There is at most one yellow piece.
    \item[$\CheckedBox$] There is exactly one red square.
    \item[$\Square$] There are at most two yellow squares.
\end{itemize}

As we can see, the prompt is divided in two parts. The first part explicitly states the map between signs (the emojis) and their meanings, while the second part expresses the actual question and provides the five alternatives.

\subsection{What is SIT trying to measure?} 
With SIT we wanted to assess the language model's capability to reason and interpret a simple scene depicted solely through natural language. Here the model is required to anchor its observations in natural language and deduce the relationships between the objects present in the scene. 

The successful completion of this task necessitates the integration of several crucial abilities:

\begin{itemize}
    \item \textsc{Domain Separation of Text Tokens:} Discriminating text tokens as either language tokens or references to abstract objects.
    \item \textsc{Language Grounding:} Mapping language tokens to objects in a world.
    \begin{itemize}
        \item Parsing and immediately utilizing descriptions of new objects, i.e., assimilating semantic maps during inference.
        \item Maintaining invariance to incorrect object symbols and/or nonsensical object names, demonstrating an abstract understanding of the underlying language conventions.
    \end{itemize}
    \item \textsc{Perception:} Recognizing objects in the scene.
    \begin{itemize}
        \item \textit{Object Identification:} Determining the types of pieces (square, circle, triangle).
        \item \textit{Attribute Identification:} Recognizing attributes of the pieces (color, orientation).
    \end{itemize}
    \item \textsc{Reasoning:} Making deductions about the objects in the scene.
    \begin{itemize}
        \item \textit{Counting:} Quantifying the pieces in the structures.
        \item \textit{Relational Reasoning:} Understanding positional relationships between pieces and applying common-sense meaning within the SIT context.
        \item \textit{Logical Reasoning:} Executing logic operations (and/or).
    \end{itemize}
\end{itemize}

These capabilities collectively contribute to identifying the most probable explanation. 

\subsection{SIT subtasks}

What we have just discussed is only one of the five subtasks of SIT, the \textit{plain} one. The others differ in the map between signs and their meanings, making it not trivial:

\begin{itemize}
    \item \textbf{Plain:} ``\includegraphics[height=1em]{emoji-images/red-square.png} is a red square.''
    \item \textbf{Agnostic emoji-side:} ``\includegraphics[height=1em]{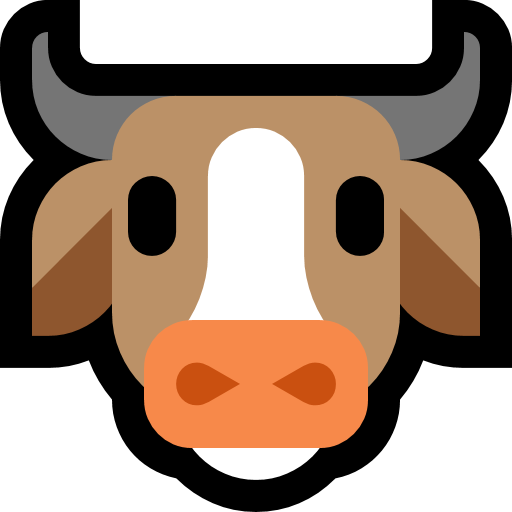} is a red square.''
    \item \textbf{Agnostic name-side:} ``\includegraphics[height=1em]{emoji-images/red-square.png} is a X Y.''
    \item \textbf{Tricky:} ``\includegraphics[height=1em]{emoji-images/red-square.png} is a der reauqs.''
    \item \textbf{Adversarial:} ``\includegraphics[height=1em]{emoji-images/yellow-circle.png} is a red square.''
\end{itemize}

Overall these subtasks want to test to what degree LLMs are malleable in their use of symbols. 
Malleability is a fundamental trait of proficiency in the use of symbols and, therefore, an ability that any system aspiring to be recognized as an artificial scientist should express. In the words of \citet{santoro2021symbolic}: \textit{Fluent symbol users understand that the conventionality of meaning allows for change. Meaning can be altered by context, by the creation of other symbols or concepts, or by intentionally redefining the symbol.} 

The motivations for each individual subtask are summarized as follows:
\begin{itemize} 
    \item \textsc{SIT Agnostic Emoji-side:} Assesses the ability to connect meaningful words to unrelated world objects. We expect this to be the easiest task: an arbitrary definition of a set of symbols.
    \item \textsc{SIT Agnostic Name-side:} Evaluates the capacity to use arbitrary placeholders instead of meaningful words. An ability fundamental e.g. in understanding computer code.
    \item \textsc{SIT Tricky:} Tests the ability to compose capabilities, such as working with words in reverse order. Here we are interested in particular in seeing if there is a gap with \textit{SIT Agnostic Name-side}, to see if the stage of symbol definition is accessible for the general capabilities of the model, like reversing a string. 
    \item \textsc{SIT Adversarial:} Challenges the model to work with an unnatural mapping, against common-sense emoji-meaning associations seen in training. We expect this to be the hardest subtask, since the model is required to erase and redefine the existing default map between emojis and their meanings at test time, not just to define a new one. 
\end{itemize}

\subsection{SIT subtasks results}

\begin{figure}[h]
    \centering
    \includegraphics[width=0.49\linewidth]{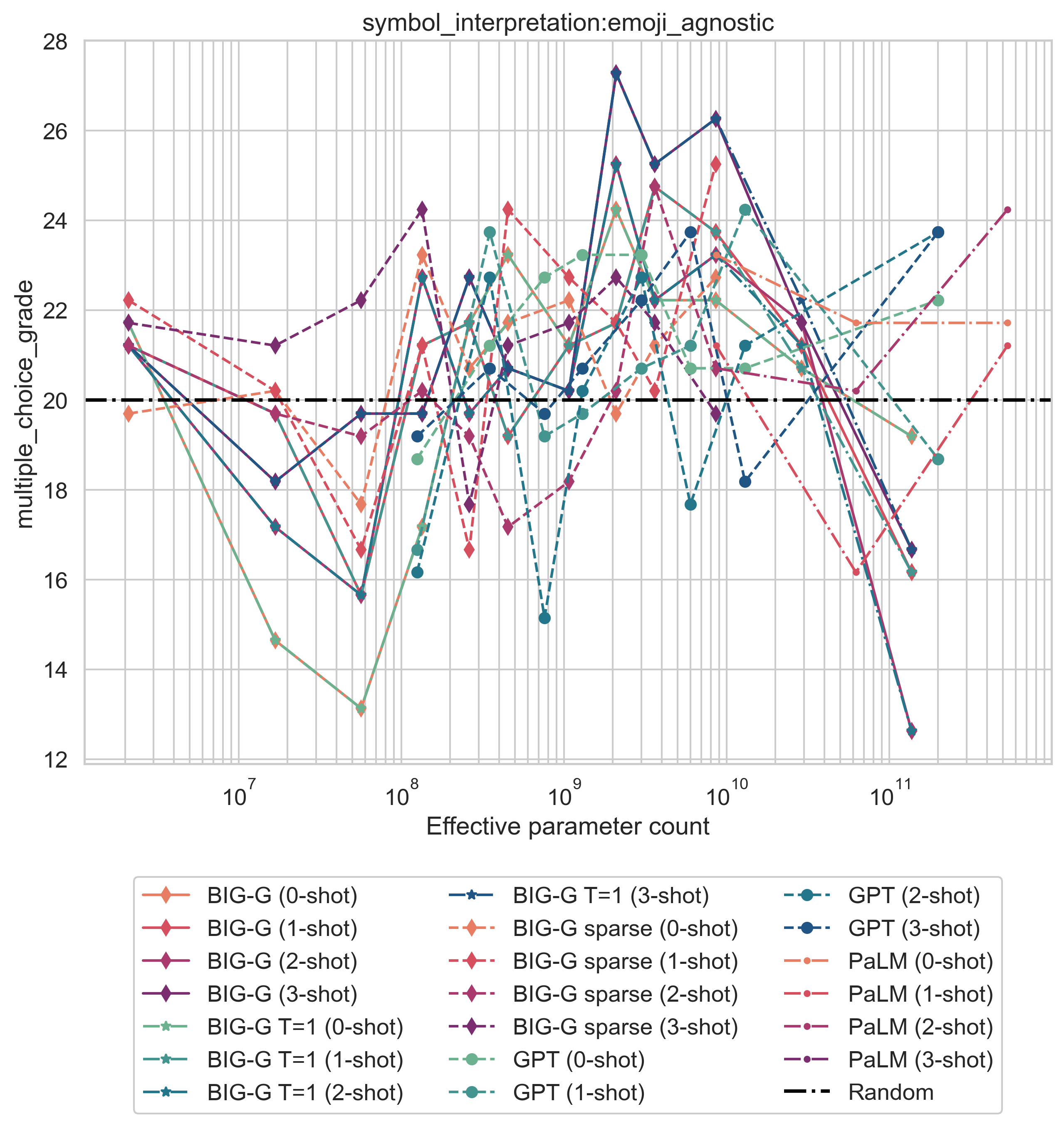}
    \includegraphics[width=0.49\linewidth]{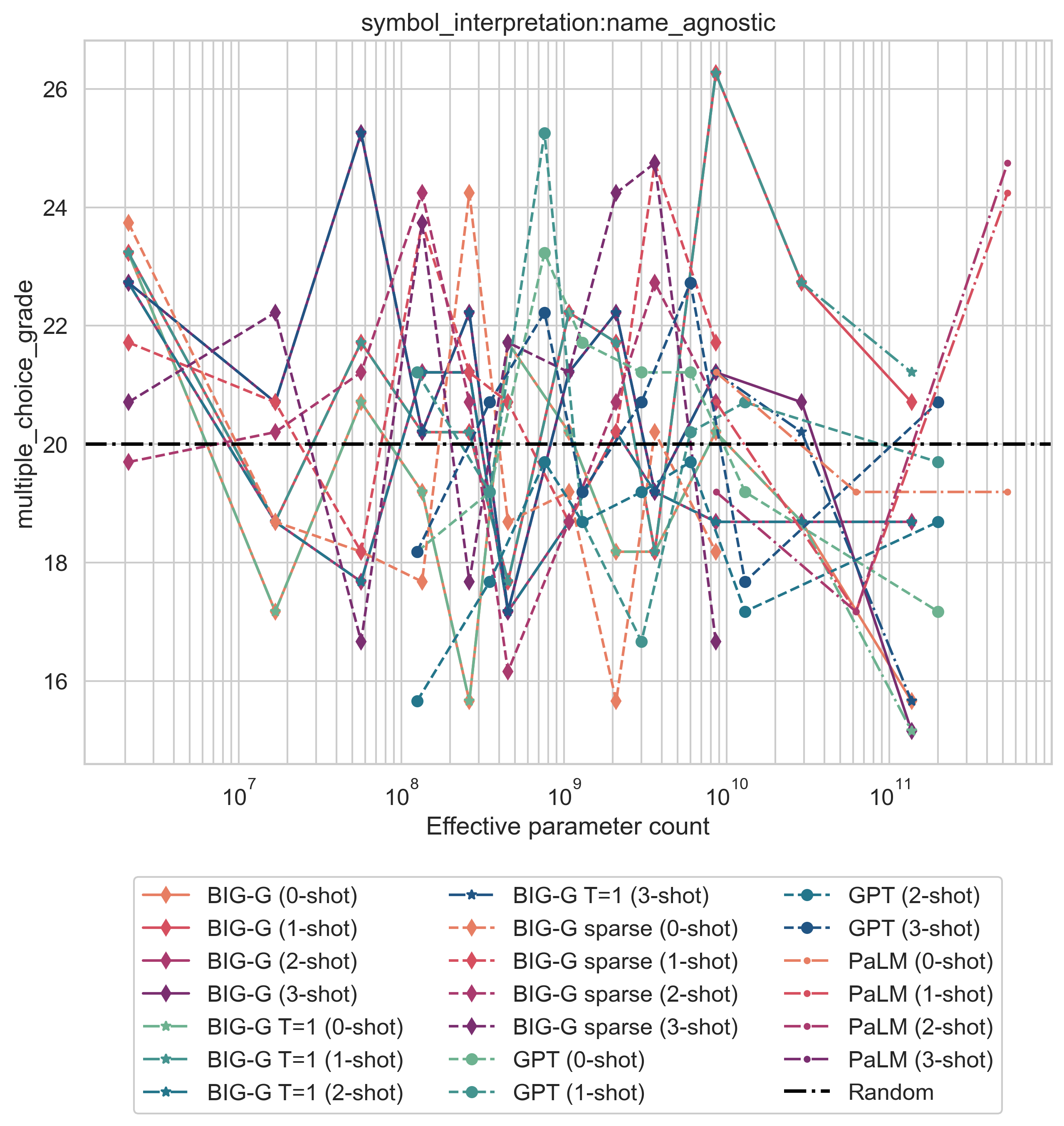}
    \includegraphics[width=0.49\linewidth]{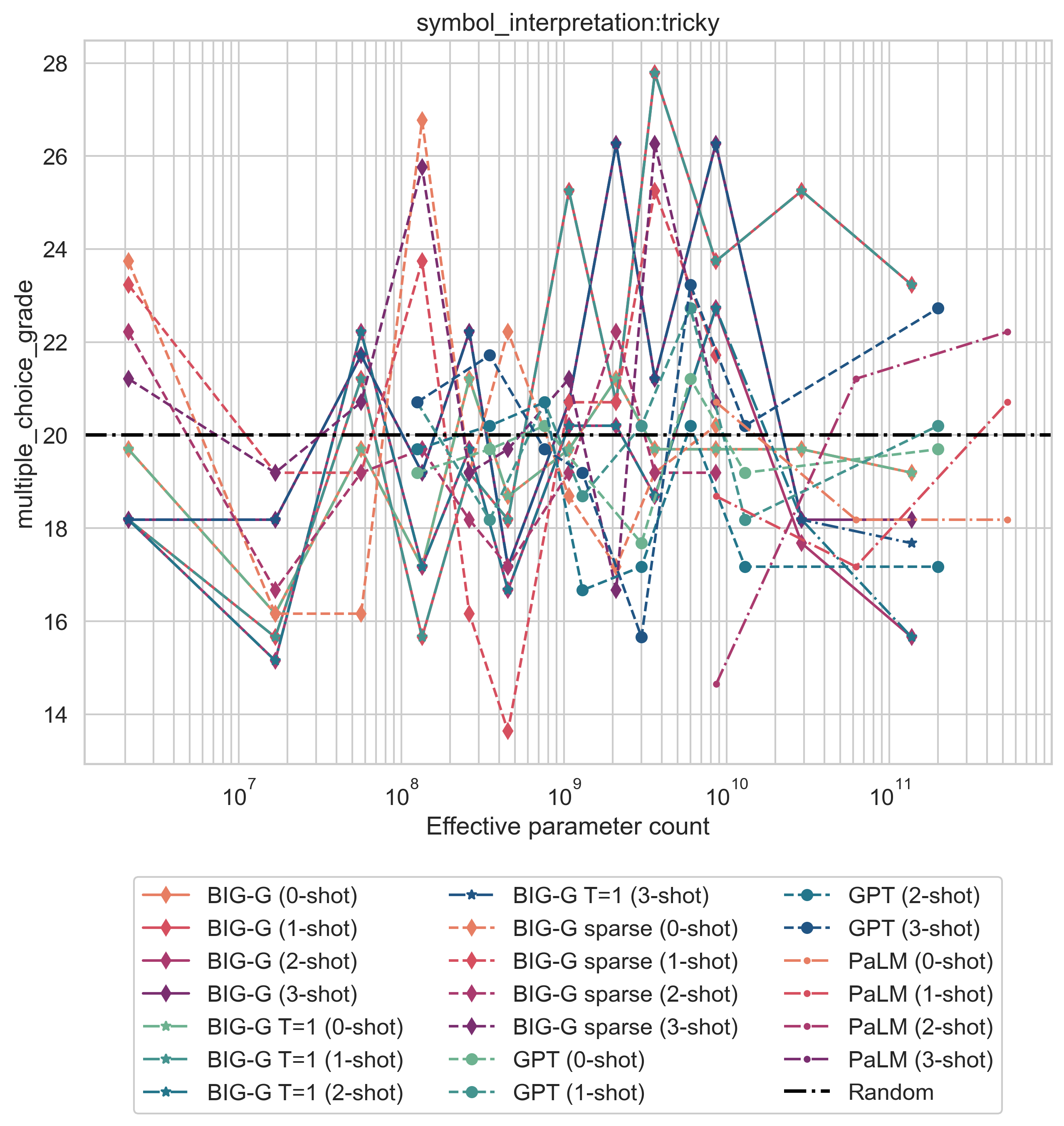}
    \includegraphics[width=0.49\linewidth]{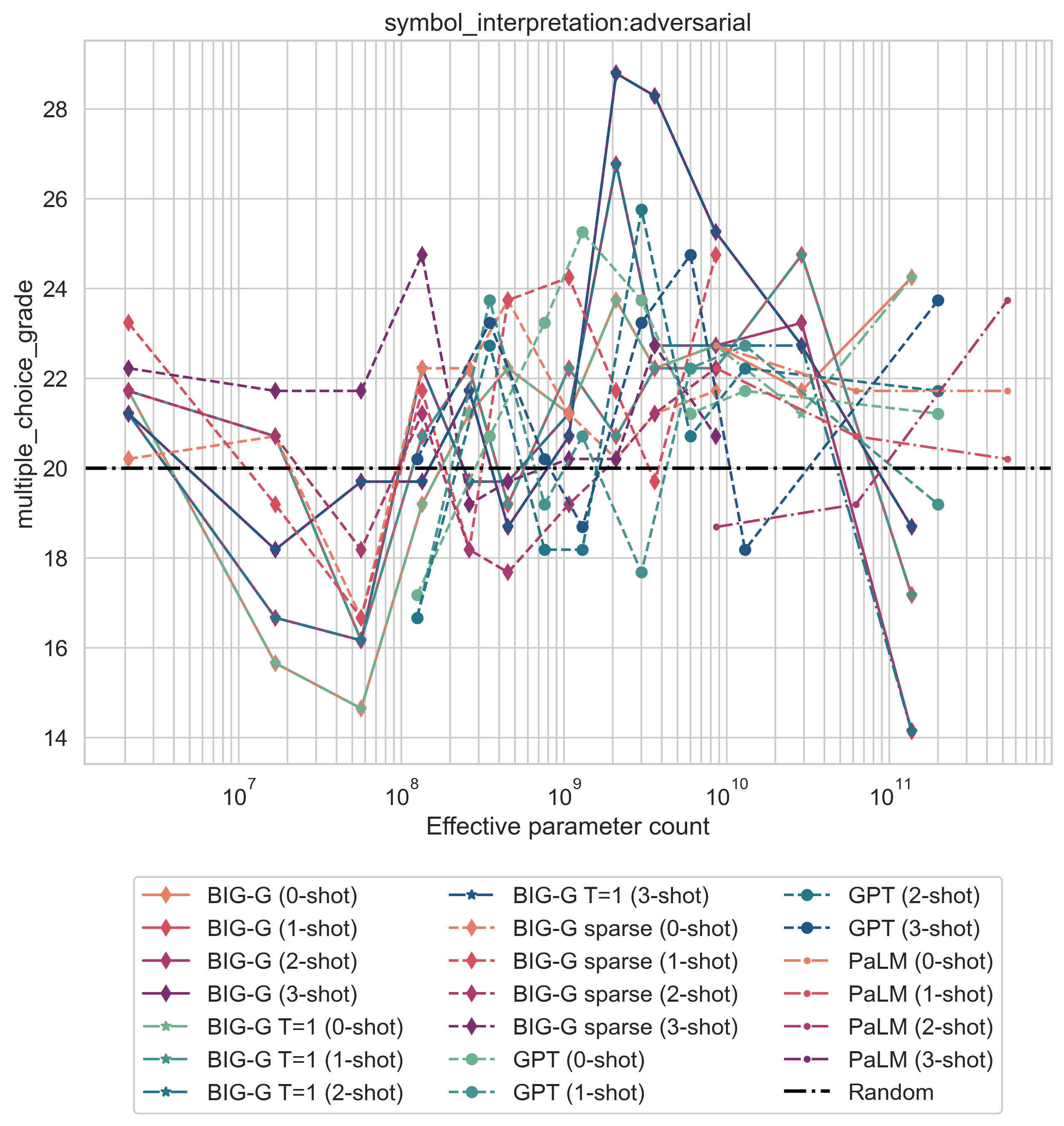}
    \caption[LLMs performance on the SIT subtasks]{Comparison of multiple-choice grade performance across various models with different effective parameter counts on the symbol\_interpretation subtasks within the Big-Bench project. No LLM perform significantly better than random chance. }
    \label{fig:sit-subtasks}
\end{figure}

Unfortunately, as shown in figure \ref{fig:sit-subtasks}, current SOTA LLMs do not go beyond random chance in every subtask, not making it possible to measure any gap. This result was expected given the same performance on the \textit{plain} subtask, which we consider to be the easiest. We look forward to the advent of more powerful models capable of surpassing random chance on the \textit{plain} task to then assess their effective malleability in the use of symbols.

\begin{table}
\centering
\caption[The Symbol Interpretation Task shows the largest performance gap between human participants and LLMs.]{Results obtained by PaLM LLMs increasing in size across 24 BIG-bench Lite tasks on 5-shot evaluation. Our Symbol Interpretation Task shows the largest performance gap between human participants and LLMs.}
\label{tab:zendo-sit}
\begin{tabular}{l*{5}{r}}\toprule
& \multicolumn{3}{c}{PaLM} & \multicolumn{2}{c}{Human} \\
Task & 8B & 62B & 540B & (Avg.) & (Best) \\
\midrule
\footnotesize{auto\_debugging} & 14.7 & 38.2 & 38.2 & 13.7 & 50.0 \\
\footnotesize{bbq\_lite\_json} & 71.2 & 91.0 & 96.3 & 59.9 & 88.4 \\
\footnotesize{code\_line\_description} & 15.0 & 26.7 & 90.0 & 60.6 & 100.0 \\
\footnotesize{conceptual\_combinations} & 25.9 & 59.7 & 81.3 & 83.2 & 100.0 \\
\footnotesize{conlang\_translation} & 40.1 & 52.1 & 66.3 & 21.6 & 54.8 \\
\footnotesize{emoji\_movie} & 25.0 & 55.0 & 91.0 & 92.6 & 100.0 \\
\footnotesize{formal\_fallacies\_syllogisms\_negation} & 48.8 & 49.5 & 53.3 & 54.1 & 80.0 \\
\footnotesize{hindu\_knowledge} & 34.9 & 77.7 & 95.4 & 61.9 & 100.0 \\
\footnotesize{known\_unknowns} & 50.0 & 58.7 & 73.9 & 80.2 & 100.0 \\
\footnotesize{language\_identification} & 11.8 & 12.9 & 36.0 & 16.1 & 55.0 \\
\footnotesize{logic\_grid\_puzzle} & 33.3 & 36.5 & 42.4 & 45.1 & 100.0 \\
\footnotesize{logical\_deduction} & 24.4 & 31.2 & 43.3 & 40.1 & 88.9 \\
\footnotesize{misconceptions\_russian} & 36.7 & 38.8 & 59.2 & 64.7 & 100.0 \\
\footnotesize{novel\_concepts} & 46.9 & 59.4 & 71.9 & 67.2 & 100.0 \\
\footnotesize{operators} & 23.8 & 45.2 & 59.0 & 45.6 & 90.6 \\
\footnotesize{parsinlu\_reading\_comprehension} & 27.6 & 42.3 & 49.2 & 1.5 & 30.0 \\
\footnotesize{play\_dialog\_same\_or\_different} & 43.2 & 51.2 & 53.1 & 52.2 & 100.0 \\
\footnotesize{repeat\_copy\_logic} & 6.2 & 15.6 & 53.1 & 38.8 & 83.3 \\
\footnotesize{strange\_stories} & 41.9 & 63.8 & 87.9 & 80.1 & 100.0 \\
\footnotesize{strategyqa} & 55.4 & 65.4 & 73.9 & 62.9 & 90.0 \\
\textbf{\footnotesize{symbol\_interpretation}} & \textbf{19.0} & \textbf{19.9} & \textbf{21.9} & \textbf{36.8} & \textbf{100.0} \\
\footnotesize{vitaminc\_fact\_verification} & 38.9 & 44.4 & 70.2 & 63.7 & 100.0 \\
\footnotesize{winowhy} & 55.7 & 61.0 & 65.9 & 52.3 & 84.4 \\
\footnotesize{linguistics\_puzzles} & 0.0 & 0.1 & 0.1 & 0.0 & 0.0 \\
\bottomrule
\end{tabular}
\end{table}

%% file: Chapters/Chapter06.tex
\chapter{Conclusions}\label{ch:conclusions}

\section{Recap}

\begin{flushright}{\slshape
    If you will tell me precisely what it is that a machine cannot do, then I can always make a machine which will do just that.} \\ \medskip
    --- John von Neumann
\end{flushright}

In these pages, we have tried to get more precise in what traits of a human scientist are still missing in machines. Our journey began by confronting a fundamental tension highlighted early on: the gap between a machine's ability to discover knowledge and its capacity to communicate it meaningfully.
This was vividly illustrated by {\sc Olivaw}, our AlphaGo Zero-inspired Othello agent. While it autonomously mastered the game to a world-class level using minimal resources, its knowledge remained locked within its neural weights, unable to articulate the principles it had learned in a way that could enlighten human players. 

This critical limitation—the inability to explain—propelled us towards defining Explanatory Learning (EL). We sought agents capable not just of prediction, but of justification within a shared symbolic framework. This required moving beyond traditional approaches that assume a fixed, human-defined interpreter for the language of explanation. We proposed Critical Rationalist Networks (CRNs) as a model embodying this, capable of autonomous symbol interpretation, learning the meaning of explanatory symbols directly from observations. We tested these ideas within the Odeen environment, a microcosm designed to simulate the challenge of scientific inquiry where understanding the language is part of the discovery process itself.

Recognizing that autonomous interpretation is key, we then dissected the process of building meaning between different modalities, a cornerstone of how humans often ground abstract concepts. This led to ASIF, a novel, training-free method for creating multimodal understanding (specifically vision-language). By leveraging independently pretrained unimodal encoders and a small set of coupled data (image-text pairs) as a 'Rosetta Stone', ASIF demonstrated that aligned meaning spaces could be constructed transparently and efficiently. This approach offers unprecedented model editability (adding/removing knowledge by simply managing the coupled data embeddings) and raises fundamental questions about the roles of explicit end-to-end training versus structured retrieval and composition in building foundation models.

Finally, we turned our lens to the current state-of-the-art, Large Language Models (LLMs) like those powering ChatGPT. Are these the nascent artificial scientists we seek, perhaps needing only scale or refinement? Our analysis, culminating in the Symbol Interpretation Task (SIT) derived from Odeen and run within the Big-Bench collaboration, revealed a stark contrast. Despite their remarkable fluency and breadth of knowledge, current LLMs exhibited critical shortcomings inconsistent with scientific practice: a propensity for fabrication ("hallucination"), a fundamental inability to gauge their own ignorance, and an uncritical acceptance of input data. Their failure on SIT – performing at random chance where humans demonstrated clear understanding and even mastery – underscored the significant, qualitative gap that remains.

Thus, from the silent competence of {\sc Olivaw} to the fluent but flawed responses of LLMs, this thesis has charted a course attempting to delineate the essential capabilities of an artificial scientist. We've moved from identifying the communication bottleneck to proposing frameworks for learnable interpretation (EL/CRNs) and efficient mechanisms for the construction of cross-modal meaning (ASIF), ultimately providing a clearer, empirically grounded picture of the specific, non-trivial challenges that still lie between current AI and the goal of autonomous, communicable scientific discovery.

\section{Beyond Intelligence}

The remarkable triumphs of deep learning over the past decade have propelled artificial intelligence from the realm of science fiction into tangible reality. The aspiration to build an artificial scientist—a machine capable of generating novel ideas, conducting autonomous research, and explaining it to us advancing human knowledge—has shifted from a distant fantasy to an ambitious, yet attainable, objective.

With the advent of models like ChatGPT, intelligence has become an abundant resource: accessible, affordable, and seamlessly integrated into numerous aspects of our daily lives. November 30, 2022—the launch date of ChatGPT—may well be commemorated in future history books as a pivotal moment in the evolution of AI. However, this monumental achievement has also illuminated a critical realization: intelligence alone, as we currently understand and implement it, is insufficient to automate progress.

One prevailing argument posits that an intelligent agent is fundamentally limited without the capacity to interact with the physical world, emphasizing the indispensable role of robotics in achieving genuine understanding and agency. Conversely, perspectives like DeepMind's motto, "solving intelligence, and then using that to solve everything else" \cite{google2016ai}, and OpenAI's decision to disband its robotics team \cite{oitzman2021openai}, suggest that intelligence is the singular, all-encompassing quest.

In reflecting upon these viewpoints, 
I contend that there is indeed a primary quest, but it transcends raw intelligence. It is the capacity for scientific discovery--our ability to distill our skills into a physical form, a knowledge base, that enables other humans to benefit from the same skill without having to discover it from scratch. We call those who produce the symbolic sequences that materially constitute this pile of knowledge "scientists," and I believe that building a machine capable of adding pieces to this pile should be our next goal.

Achieving this objective necessitates addressing several critical challenges. First, we must enhance our model of intelligence with good reasoning capabilities, a trait that is still lacking in present-day Large Language Models, as we have seen in Odeen \cite{srivastava2023beyond}; and full multimodality, machines must autonomously interpret information across diverse sensory signals. Second, besides intelligence, we should understand agency—the capacity to make autonomous decisions, such as doing experiments—and curiosity—the drive to explore the unknown, e.g. posing relevant questions. These attributes fuel the scientific endeavor and must be encapsulated within an artificial scientist. While recent attempts in modeling the agency and curiosity traits are promising \cite{pathak2017curiositydriven, kaur2021ask}, our understanding is still very shallow, and constitutes one of the most significant limitations of today's flagship LLMs.  

\medskip 

As we stand on the cusp of this new frontier, I believe a growing wave of AI labs worldwide will join the quest to build the first artificial scientist.

Eventually it will happen, just as it did with Go, opening up a new stream of discoveries. In the words of \citet{deutsch2011beginning}, a second beginning of the infinity.

%% file: Chapters/Chapter0A1.tex
\def\I{$\mathcal{I}$}
\def\CG{$\mathcal{CG}$}
\def\Zendo{{Odeen}}
\definecolor{myblue}{rgb}{0,0.447,0.741}
\definecolor{myorange}{rgb}{0.837,0.625,0.112}

\newcommand{\regimen}{training regimen}
\newcommand{\Regimen}{Training regimen}
\newcommand{\regimens}{training regimens}
\newcommand{\Regimens}{Training regimens}

\newcommand{\Softscore}{NR-score}
\newcommand{\SoftscoreAbbr}{NRS}
\newcommand{\NRS}{\SoftscoreAbbr{}}

\newcommand{\NumRules}{1438}
\newcommand{\BestRes}[1]{\textsc{#1}}

\newcommand{\RuleAccuracy}{Rule-Accuracy}
\newcommand{\RA}{R-Acc}

\newcommand{\LabelAccuracy}{Label-Accuracy}
\newcommand{\LA}{T-Acc}

\newcommand{\xmark}{\ding{55}}

\newenvironment{bnfsplit}[1][0.7\textwidth]
 {\minipage[t]{#1}$}
 {$\endminipage}

\chapter{Appendix Explanatory Learning}

\section{Further details on the Odeen dataset}

\paragraph*{Training set.}
The total number of rules produced by the \Zendo{}\footnote{Odeen is the Rational alien in \emph{The Gods Themselves} (1972), a novel by Isaac Asimov about a conspiracy against Earth by the inhabitants of a parallel universe with different physical laws.} grammar is 24,794. 
We consider training sets varying from 500 to 1438 rules.
We choose these rules such that each token and each syntactic construct appears at least once; then, we uniformly select the others from the distribution.
We removed from the training set any rule containing the 
bigram \texttt{exactly 2}, as well as any rule of the form \texttt{at\_least 2 X and at\_most 2 X}, equivalent to \texttt{exactly 2 X}.
Each rule is associated with a set of 100, 1,000, or 10,0000 labelled structures that unambiguously identify a rule equivalence class.

\paragraph*{Test set.}
We generate the 1,132 games that compose the test set the same way, with the additional constraint of excluding the rules belonging to an equivalence class that is already in the training set. In the test set 72 rules contain the bigram \texttt{exactly 2}. Rules in the test set are associated with just 32 labelled structures. The first 10 structures are chosen by searching pairs of similar structures with different labels, following a common human strategy in {Zendo}. 
The remaining 22 structures are selected to ensure the lack of ambiguity on the board.

\paragraph*{Formal definition of the \Zendo{} grammar.}

\begin{figure*}[ht]
\begin{bnf*}
\bnfprod{RULE}{
\bnfpn{PROP\_S}\bnfor
\bnfpn{PROP}\bnfor
\bnfpn{PROP\_S}\bnfts{ }\bnfpn{CONJ}\bnfts{ }\bnfpn{PROP\_S}
}
\\
\bnfprod{PROP}{
\bnfpn{QTY}\bnfts{ }\bnfpn{OBJ}\bnfts{ }\bnfpn{REL}\bnfts{ }\bnfpn{OBJ}
}
\\
\bnfprod{PROP\_S}{
\bnfpn{QTY}\bnfts{ }\bnfpn{OBJ}
}
\\
\bnfprod{OBJ}{
\bnfpn{COL}\bnfor
\bnfpn{SHAPE}\bnfor
\bnfpn{COL}\bnfts{ }\bnfpn{SHAPE}
}
\\
\bnfprod{QTY}{
\bnfts{at\_least }\bnfpn{NUM}\bnfor
\bnfts{exactly }\bnfpn{NUM}\bnfor
\bnfts{at\_most }\bnfpn{NUM}\bnfor
\bnfts{zero}
}
\\
\bnfprod{SHAPE}{
\bnfts{pyramid }\bnfpn{ORIEN}\bnfor
\bnfts{pyramid}\bnfor
\bnfts{block}
}
\\
\bnfprod{REL}{
\bnfts{touching}\bnfor
\bnfts{surrounded\_by}\bnfor
\bnfts{at\_the\_right\_of} }
\\
\bnfprod{ORIEN}{
\bnfts{pointing\_up}\bnfor
\bnfts{pointing\_down}
}
\\
\bnfprod{NUM}{
\bnfts{1}\bnfor
\bnfts{2}
}
\\
\bnfprod{CONJ}{
\bnfts{and}\bnfor
\bnfts{or}
}
\\
\bnfprod{COL}{
\bnfts{red}\bnfor
\bnfts{blue}
}
\end{bnf*}
\vspace{-0.75cm}
\caption{\label{fig:bnf}
Grammar productions for the Odeen Language.}
\end{figure*}

The context-free grammar in Figure \ref{fig:bnf} defines all the acceptable rules in \Zendo{}.
This grammar only formalizes which rules are \emph{syntactically correct}.
Token names (e.g. \texttt{red}, \texttt{1} or \texttt{touching}) do not imply any rule meaning.

{The hard-coded interpreter  formalizes how to interpret the rules. Similarly to compilers, it tokenizes and transforms the rule into an abstract syntax tree (AST). The interpreter then adds semantic information to the AST, establishing the truth value of each node based on the truth value of its children and the structure under evaluation.}

\subsection{The Odeen binary semantic representations.} 

\begin{figure}[h!]
\begin{center}
            \includegraphics[width=0.74\textwidth]{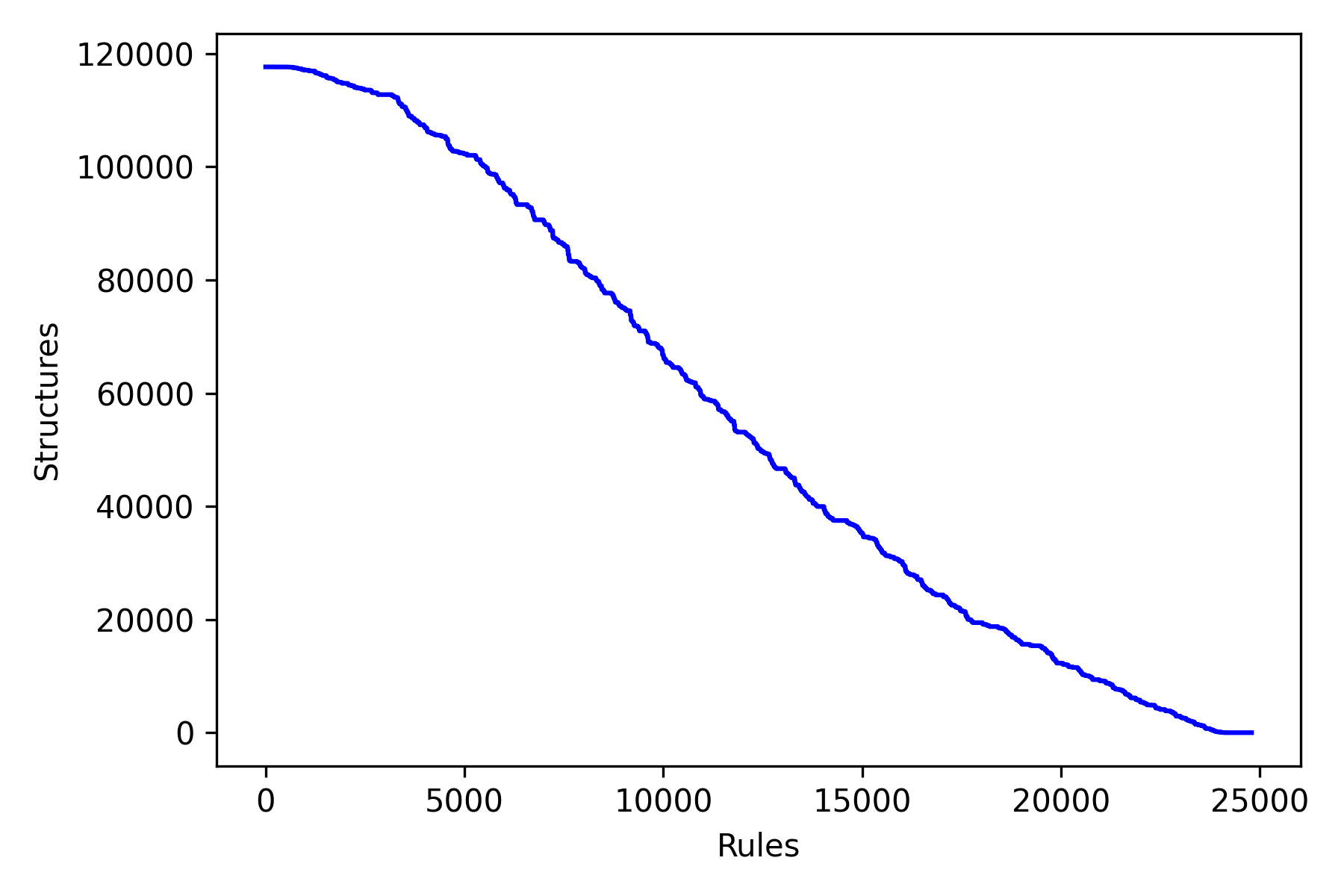}
                                                                \includegraphics[width=0.74\textwidth]{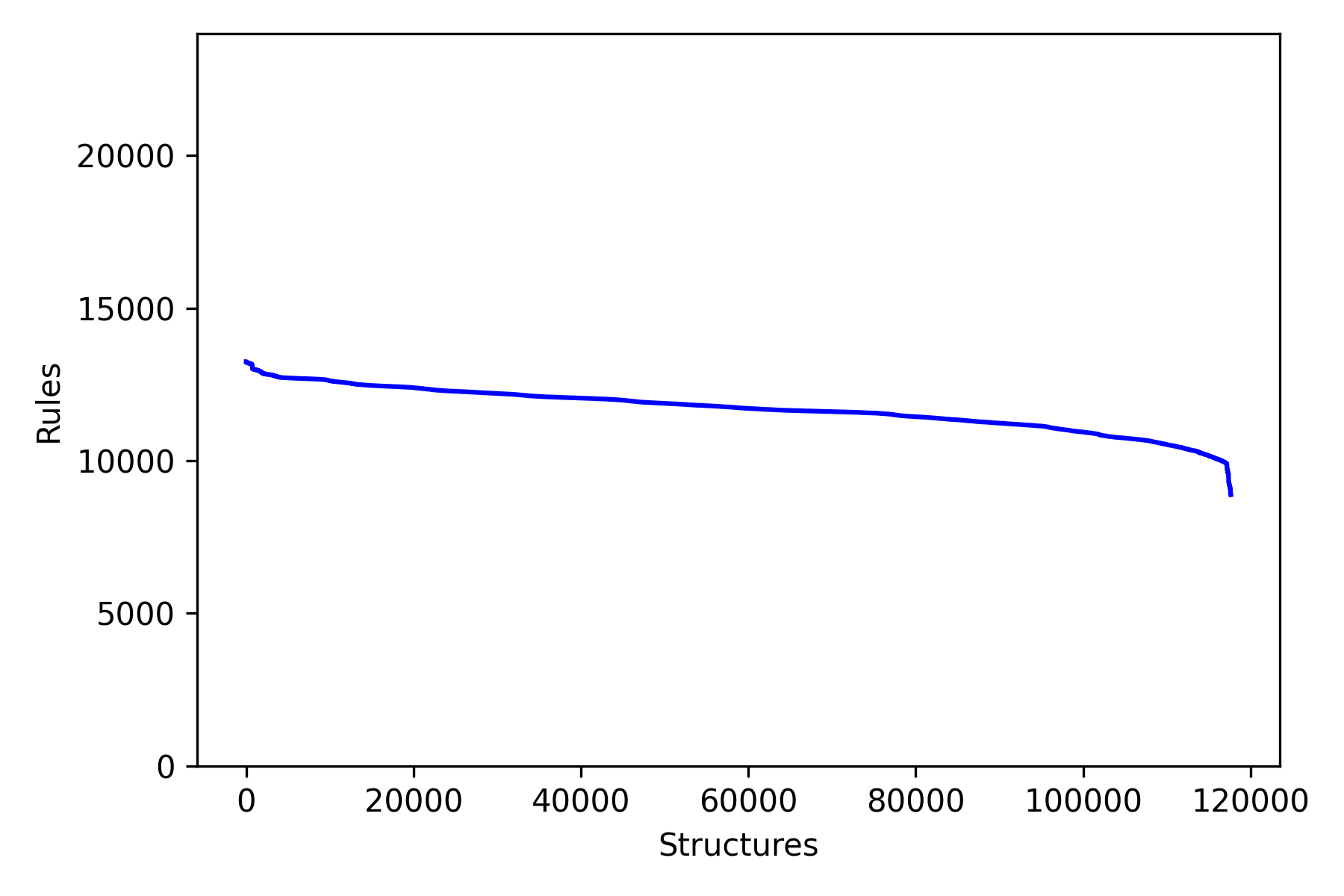}
                                    \caption[Hamming weight of structures and rules representations in Odeen]{\label{fig:equivalence_classes} Hamming weight of the binary semantic representation of each rule (up) and each structure (down). We sort them in descending order for visualization purposes. While rules span a wide range --we have rules satisfied by all, some, or zero structures, with a smooth transition-- structures are all satisfying around half of the rules. 
                }
\end{center}
\end{figure}
By simulating the process of scientific discovery, \Zendo{} offers a convenient simulation of a world described by a language. Besides the computational tractability, the simplicity and adjustable size of the \Zendo{} world allows us to explicit the whole semantics of its language. 

This semantics can be encoded in a binary \emph{semantic matrix} $S$ with the 24,794 rules $e_i$ on the rows and the 117,649 structures $x_j$ on the columns. The $s_{ij}$ element of this matrix is equal to 1 if 
the structure $x_j$ complies with the rule $e_i$  and 0 otherwise, see inset in Section 3, Metrics paragraph.
$S_{i*}$, the 117,649-dimensional binary vector coinciding with the $i$-th row of $S$, fully represents the meaning of rule $e_i$ in the \Zendo{} world.
Similarly, each structure $x_j$ is represented by the 24,794-dimensional binary vector coinciding with the column $S_{*j}$ of $S$. 

In Figure \ref{fig:equivalence_classes}, we analyze the distribution of the Hamming weights (i.e., the number of ones) in $\{S_{i*}\}_{i=1}^{117,649}$ (up) and $\{S_{*j}\}_{j=1}^{24,794}$ (down).
We observe an asymmetry between the rule and structure distributions.
On one hand, the semantic representation of a rule can be quite unbalanced, with populated extremes of rules evaluating \emph{all} structures with 1 (or 0) as shown in Figure \ref{fig:equivalence_classes} (up).
On the other hand, Figure \ref{fig:equivalence_classes} (down) shows that the semantic representations of structures are very balanced; most of them have around half zeros and half ones, with no structure with less than 10k or more than 14k ones.

\begin{figure*}
\begin{center}
    \includegraphics[width=0.89\textwidth]{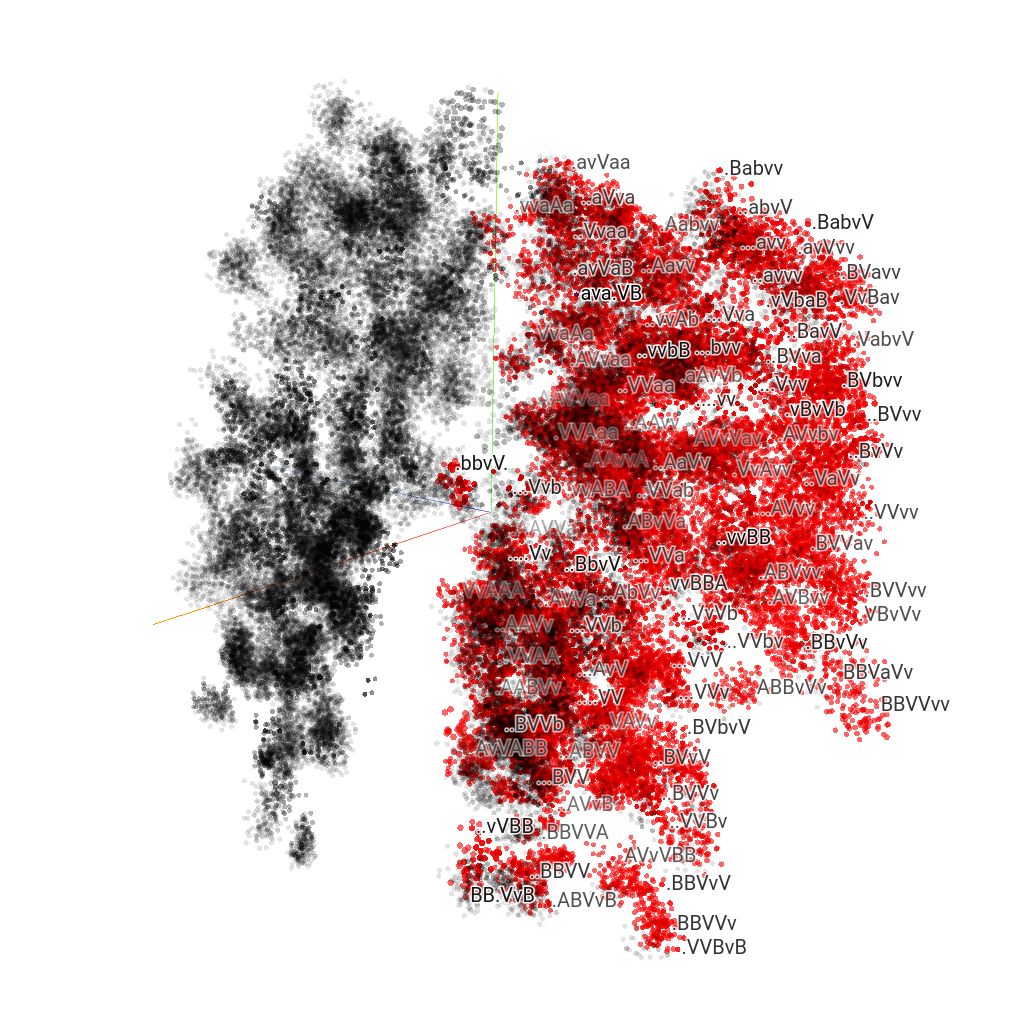}
    \vspace{-1em}
             \includegraphics[width=0.89\textwidth]{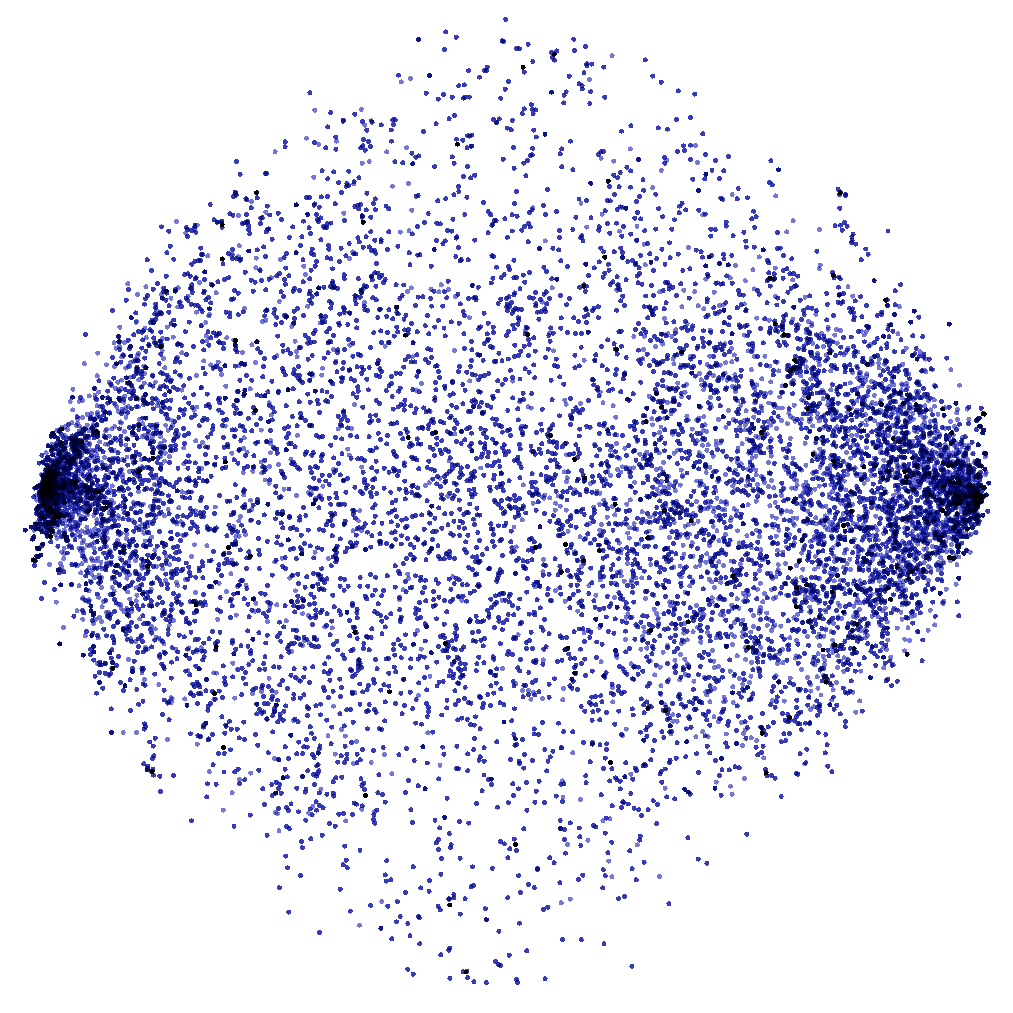}
    \vspace{1em}
                    \caption[PCA on the structures representations]{
PCA applied to the binary semantic representation of structures (up) and rules (down). We highlight in red the structures that have two touching pyramids pointing down. To ease the visualization, we represent structures with characters (\textit{A, V} are red pyramids pointing up and down; \textit{B} is a red block, lowercase chars are blue pieces and the \textit{.} is a space).
The rules distribution reflects what can be observed in Figure~\ref{fig:equivalence_classes} (up).
}
\label{fig:pca-rules}
\end{center}
\end{figure*}

This balanced trend, along with the well separable PCA of $\{S_{*j}\}_{j=1}^{24,794}$ (Figure \ref{fig:pca-rules}) suggests that the chosen language produce representations that are effective in separating structures. Conversely, the PCA of $\{S_{i*}\}_{i=1}^{117,649}$ is much less homogeneous (Figure \ref{fig:pca-rules}). Here we can recognize two poles, corresponding respectively to rules with all ones and all zeros.
We believe that this analysis of the binary semantic representations is only partial, and we leave further exploration for follow-up work.

\section{Implementation Details}\label{sec:trans}

In this paragraph, we give the implementation details of the models proposed and depicted in Figure 3 (right).
{All the models are based on a Transformer block composed of 4 layers and 8 heads.} We used a hidden dimension of 256 for all the models except for the interpreter, where we used a hidden dimension of 128.
The models differ primarily by the type of transformer block used (encoder/decoder), inputs and embeddings.
In detail:

\begin{itemize}
            
    \item {\sc Transformer Label Decoder}. This is a transformer block used to predict a label given a structure. The input structure is a sequence of six learned embeddings, one per piece. We add a sinusoidal positional encoding to each embedding as in the original transformer implementation. 
    The embedding size is 128 in the \I{} and 256 in the Empiricist models (EMP-C, EMP-R). 
    We used the standard transformer encoder block and added a special token \texttt{[CLS]} at the beginning of the structure to perform the classification task.

    \item {\sc Transformer Rule Decoder}. This is a transformer decoder block with embedding size of 128 and sinusoidal positional encoding. This decoder block is used to generate the rule by the EMP-C and \CG{} models.
            
    \item {\sc Transformer Board Encoder}. This is a transformer encoder block used to encode the (structure, label) pairs. The input is encoded a sequence of 32 learned embeddings, one per structure-label pair. The size of each embedding is 256.
        We did not add positional encodings, since the specific position of structure-label pairs among the 32 is not relevant. This block is used in all the models.

    \item {\sc Transformer Rule Encoder}. This transformer encoder block is used in \I{} to encode the rule. Its implementation is analogous to the {\sc Transformer Rule Decoder}, with the only difference that it does not use causal attention since it is an encoder layer.

                    \end{itemize}

\begin{table}[ht]
\caption{Number of training epochs for each training regimen in Odeen.}
\label{tab:epoch_numbers}
\vskip 0.15in
\begin{center}
\begin{small}
\begin{sc}
\begin{tabular}{ll|c}
\toprule
\multicolumn{2}{l}{Training regimen} &  Number of epochs \\

\midrule
10K struct. &1438 rules & 2 \\

1K struct. &1438 rules & 20\\

100 struct. &1438 rules & 200\\

10K struct. &500 rules & 6\\

1K struct. &500 rules & 58\\

100 struct. &500 rules & 576\\

\bottomrule
\end{tabular}
\end{sc}
\end{small}
\end{center}
\vskip -0.1in
\end{table}

\paragraph{Training Procedure.}
All the models are trained with a learning rate of $3\cdot10^{-4}$ using Adam optimizer, a batch size of 512 and early-stop and dropout set to 0.1 to prevent overfitting. 
 We train all the models on randomly sampled sets of 32 (structure, label) pairs to prevent overfitting on specific boards.
 Table \ref{tab:epoch_numbers} describes the number of epochs for each training regimen. Models are trained to: predict the label of a structure give the board (EMP-R); predict the label of a structure given the 32 pairs (structure, label) and the associated rule (EMP-C); predict the rule given the 32 pairs (\CG{}); predict the label of a structure given a rule (\I{}).

\section{Efficiency}
\label{ap:efficiency}

\paragraph{Data efficiency} 
In the Odeen challenge, CRNs require less training data to match the performance of empiricist models.
For instance, in the case of 1438 rules at training, we see in Table \ref{tab:table-data} that the CRN trained on 100 structures per rule (NRS$=40,2\%$) still overcomes the performance of empiricist models trained on a dataset 100 times bigger (NRS$=35.2\%$ on 10k structures per rule).

\begin{table}

\caption{T-Acc and NRS for different \regimens{}.}
\label{tab:table-data}
\vskip 0.15in
\begin{center}

\begin{small}
\begin{sc}
\begin{tabular}{l|lcc}
\toprule
 Train Data &Model &  \SoftscoreAbbr{} & \LA \\

\midrule
\multirow{3}{2cm}{10K struct. 1438 rules} &\CRN{}    & \textsc{\SoftRes{0.813}{0.813}} & \textsc{0.984} \\
&\EA{} &  \SoftRes{0.352}{0.366} & 0.930  \\
&\ER{}    &  \SoftRes{0.179}{0.181}  & 0.895  \\

\midrule
\multirow{3}{2cm}{1K struct. 1438 rules} &\CRN{}    & \textsc{\SoftRes{0.777}{0.780}} & \textsc{0.980} \\
&\EA{} &  \SoftRes{0.225}{0.240} & 0.905  \\
&\ER{}    &  \SoftRes{0.156}{0.161}  & 0.898  \\

\midrule
\multirow{3}{2cm}{100 struct. 1438 rules} &\CRN{}    & \textsc{\SoftRes{0.402}{0.404}}& \textsc{0.939} \\
&\EA{} &  \SoftRes{0.125}{0.136} & 0.865  \\
&\ER{}    &  \SoftRes{0.163}{0.183}  & 0.896  \\

\midrule
\midrule
\multirow{3}{2cm}{10K struct. 500 rules} &\CRN{}    & \textsc{\SoftRes{0.354}{0.354}}& \textsc{0.932} \\
&\EA{} &  \SoftRes{0.095}{0.107} & 0.869  \\
&\ER{}    &  \SoftRes{0.068}{0.070}  & 0.863  \\

\midrule
\multirow{3}{2cm}{1K struct. 500 rules} &\CRN{}    & \textsc{\SoftRes{0.319}{0.321}}& \textsc{0.930} \\
&\EA{} &  \SoftRes{0.088}{0.099} & 0.874  \\
&\ER{}    &  \SoftRes{0.084}{0.087}  & 0.876  \\

\midrule
\multirow{3}{2cm}{100 struct. 500 rules} &\CRN{}    & \SoftRes{0.109}{0.110}& \textsc{0.883} \\
&\EA{} &  \SoftRes{0.057}{0.064} & 0.823  \\
&\ER{}    &  \textsc{\SoftRes{0.117}{0.125}}  & 0.872  \\
\bottomrule

\end{tabular}
\end{sc}
\end{small}

\end{center}
\vskip -0.1in
\end{table}

\paragraph{Computational cost.}

In this section, we discuss the computational cost at test time of the rationalist and empiricists approaches.
Table \ref{tab:computational_cost} reports the costs of tagging $s$ new structures, while \ref{tab:computational_cost_RI} reports the costs of explicitly predicting the textual rule.
We evaluate the cost per game in two ways: i) by counting the number of calls of each trained neural network and ii) by measuring the absolute time in seconds of each method with the same hardware configuration.

We refer to the first quantity as the \emph{Computational Cost} and parameterize it in terms of the main blocks of the models. This value is independent of the batch size and the hardware adopted. 
As an example, the cost of tagging the new structures for a CRN using 300 conjectures is given by:
$$ 300 \cdot \mathcal{CG} + 300 \cdot  b \cdot \mathcal{I} +  s \cdot \mathcal{I} .$$

Where $300 \cdot \mathcal{CG}$ stands for the $300$ beams used to get $300$ conjectures from the conjecture generator $\mathcal{CG}$. Each conjecture ($300$) is then tested on all the board structures ($b$) by the interpreter \I. Finally \I{} is called to apply the chosen conjecture on each new structure ($s$).
As an upper bound, an exhaustive search algorithm ({\sc Exv src}) uses no conjecture generator, and thus has to evaluate with \I{} all admissible rules ($r$) on each structure of the board. Conversely, the empiricist approach provide label predictions through a single end-to-end model which is simply called $s$ times. Concerning the problem of inferring explicitly the textual rule,  
using more beams in the empiricists models does not provide any increase in performance, i.e. the true rule is not a more probable proposition accessible through a larger beam search.

We measured also the absolute time in seconds with the following hardware configuration for all the experiments: 1 single core hyper threaded Xeon CPU Processor with 2.2 Ghz, 2 threads; 12.7 GiB. of RAM; a Tesla T4 GPU, with 320 Turing Tensor Core, 2,560 NVIDIA CUDA cores, and 15.7 GDDR6 GiB of VRAM.

\begin{table}[h]
\caption[CRNs computational cost analysis on structures]{Computational cost of our models at test time to tag $s$ new structures. In \Zendo{} $r$=24,794, $b$=32, $s$=1,176. Notice how CRNs offer a good balance between computational efficiency and performance, this trade-off is regulated by a single parameter, the number of beams.} 
\label{tab:computational_cost}
\vskip 0.15in
\begin{center}
\begin{small}
\begin{sc}
\begin{tabular}{l|ccc}
\toprule
Model &  Computational Cost & T (s) & \SoftscoreAbbr{} \\

\midrule
{Ex. search} & {$ r \cdot  b \cdot \mathcal{I} +  s \cdot \mathcal{I}$} & {47.9} & {0.99}\\
{\RAT{} \scriptsize{[300b]}}  & {$300 \cdot \mathcal{CG} + 300 \cdot  b \cdot \mathcal{I} +  s \cdot \mathcal{I}$} & {0.79} & {0.81}\\
{\RAT{} \scriptsize{[10b]}}& {$10 \cdot \mathcal{CG} + 10 \cdot  b \cdot \mathcal{I} +  s \cdot \mathcal{I}$} & {0.43} & {0.35}\\
{\sc Emp} & {$  s \cdot $ {\sc Emp-R}}& {0.15} & {0.35}\\

\bottomrule
\end{tabular}
\end{sc}
\end{small}
\end{center}
\vskip -0.1in
\end{table}

\begin{table}[h]
\caption[CRNs computational cost analysis on rules]{Computational cost of our models at test time to produce the textual rule in output}
\label{tab:computational_cost_RI}
\vskip 0.15in
\begin{center}
\begin{small}
\begin{sc}
\begin{tabular}{l|ccc}
\toprule
Model &  Computational Cost & T (s) & R-Acc\\

\midrule
Ex. search & $ r \cdot  b \cdot \mathcal{I}$ & 47.8 & 0.99\\
\RAT{} \scriptsize{[300b]}  & $300 \cdot \mathcal{CG} + 300 \cdot  b \cdot \mathcal{I}$ & 0.72 & 0.77 \\
\RAT{} \scriptsize{[10b]}  & $ 10 \cdot \mathcal{CG} + 10 \cdot  b \cdot \mathcal{I}$ & 0.35 & 0.35\\
{\sc Emp} \scriptsize{[300b]} & $300 \cdot$ {\sc Emp-C} & 0.41 & 0.07\\
{\sc Emp} \scriptsize{[10b]} & $ 10 \cdot$ {\sc Emp-C} & 0.10 &  0.07\\
{\sc Emp} \scriptsize{[1b]}& $ 1 \cdot$ {\sc Emp-C} & 0.10 & 0.07\\

\bottomrule
\end{tabular}
\end{sc}
\end{small}
\end{center}
\vskip -0.1in
\end{table}

\newpage
\section{Odeen example games}
\label{sec-examplegames}
In the following pages we propose a collection of qualitative results showing a series of Odeen games from the test set and how they are solved by the proposed models. For each model, we report the predicted rule (only for EMP-C and the CRN), the accuracy on the structures labeling (T-acc), and a mark that indicates whether the nearest rule is the correct one (NRS). All the models are trained on 1,438 rules with 1,000 structures per rule.

\begin{figure*}[h]
\centering
\includegraphics[width=0.99\textwidth]{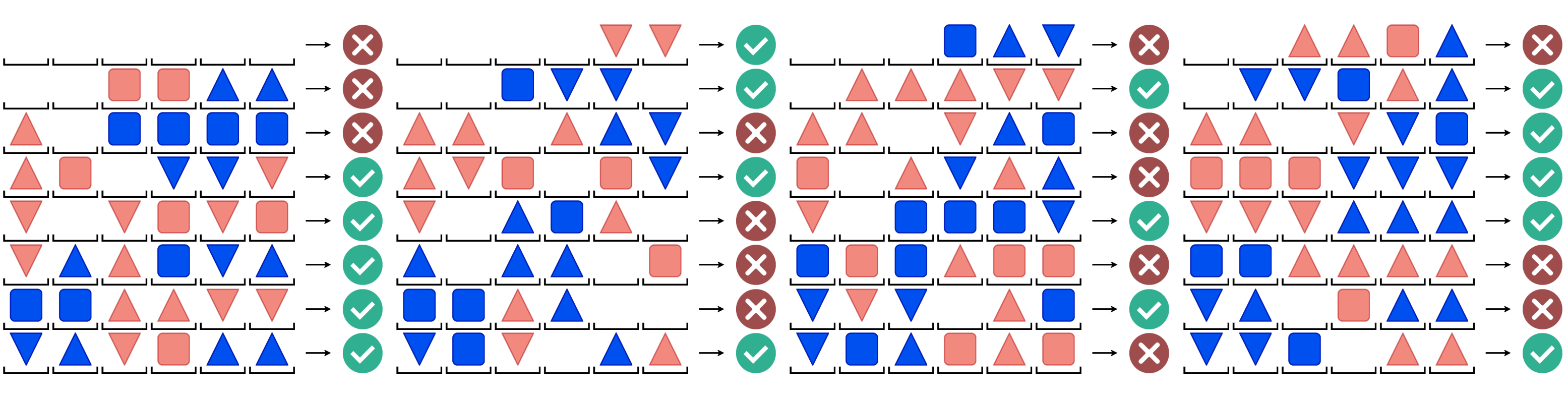}
\caption*{{Board 01} \\ {Golden Rule}: ``at\_least 2 pyramid pointing\_down" \\ \textsc{CRN}: ``at\_least 2 pyramid pointing\_down"; T-acc 1.0 \checkmark\\ \textsc{EMP-C}: ``at\_least 1 pyramid touching touching"; T-acc: 0.76 \xmark \\ \textsc{EMP-R}: T-acc 0.72 \xmark}
\end{figure*}

\begin{figure*}[h]
\includegraphics[width=0.99\textwidth]{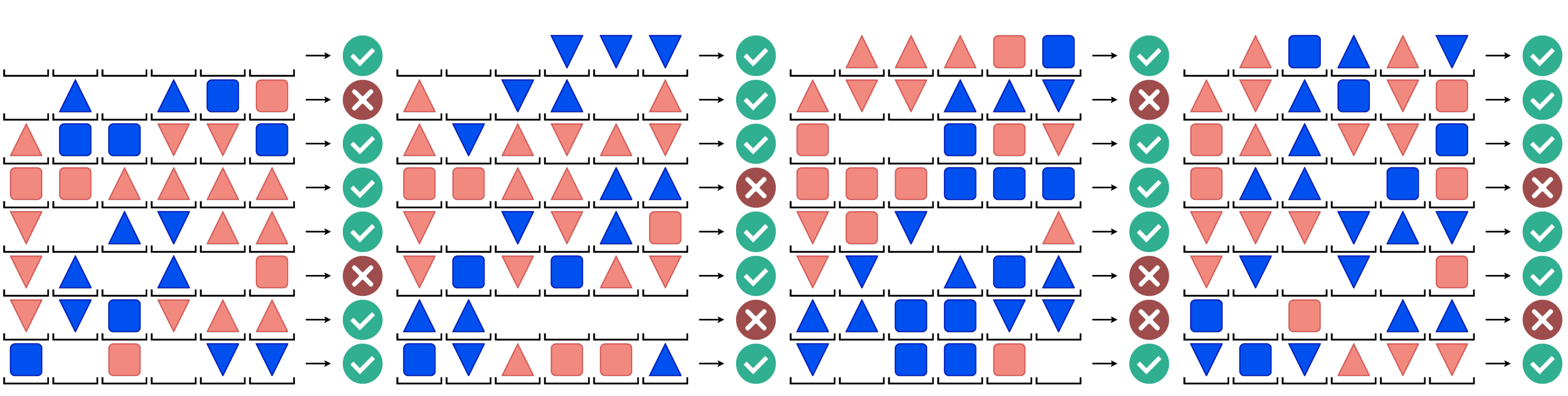}\caption*{{Board 04} \\ {Golden Rule}: ``at\_most 1 blue pyramid pointing\_up" \\ \textsc{CRN}: ``zero blue or at\_most 1 blue pyramid pointing\_up"; T-acc 1.0 \checkmark\\ \textsc{EMP-C}: ``zero 1 blue touching or or"; T-acc: 0.89 \xmark \\ \textsc{EMP-R}: T-acc 0.92 \checkmark}
\end{figure*}

\begin{figure*}[h]
\includegraphics[width=0.99\textwidth]{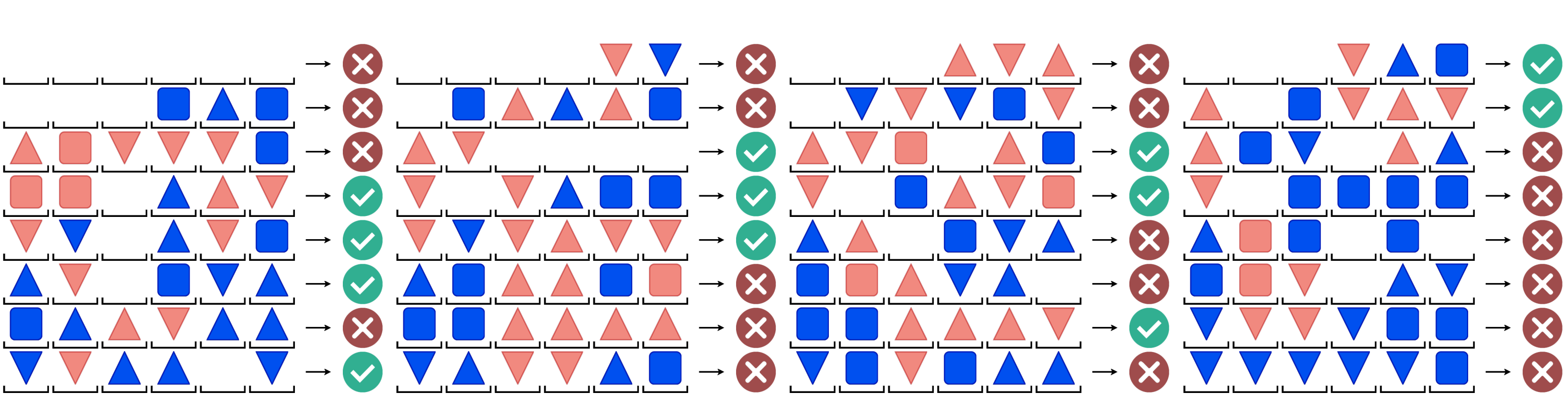}
\caption*{{Board 09} \\ {Golden rule}: ``exactly 1 pyramid pointing\_up touching red pyramid pointing\_down" \\ \textsc{CRN}: ``exactly 1 red pyramid pointing\_down touching pyramid pointing\_up", T-acc 0.95 \xmark \\ \textsc{EMP-C}: ``exactly 1 red at\_the\_right\_of and red", T-acc: 0.80 \xmark \\ \textsc{EMP-R}: T-acc 0.79 \xmark}
\end{figure*}

\begin{figure*}[h]
\includegraphics[width=0.99\textwidth]{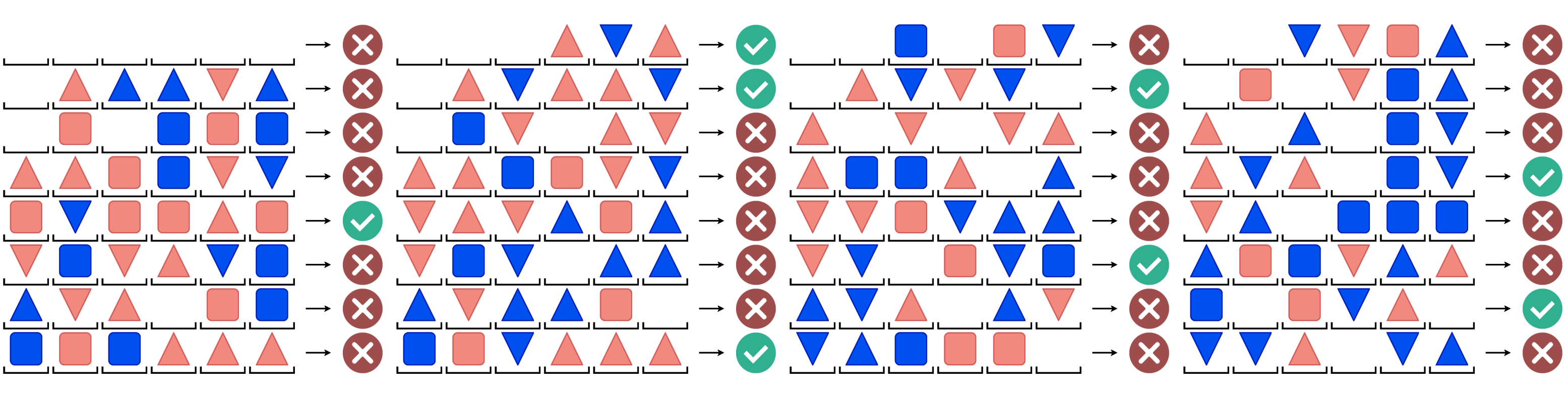}
\caption*{{Board 25}\\ {Golden rule}: ``at\_least 2 red touching blue pyramid pointing\_down" \\ \textsc{CRN}: ``at\_least 2 red touching blue pyramid pointing\_down", T-acc 1.0 \checkmark \\ \textsc{EMP-C}: ``at\_least 2 red touching blue pyramid", T-acc: 0.87 \xmark \\ \textsc{EMP-R}: T-acc 0.69 \xmark}
\end{figure*}

\begin{figure*}[h]
\includegraphics[width=0.99\textwidth]{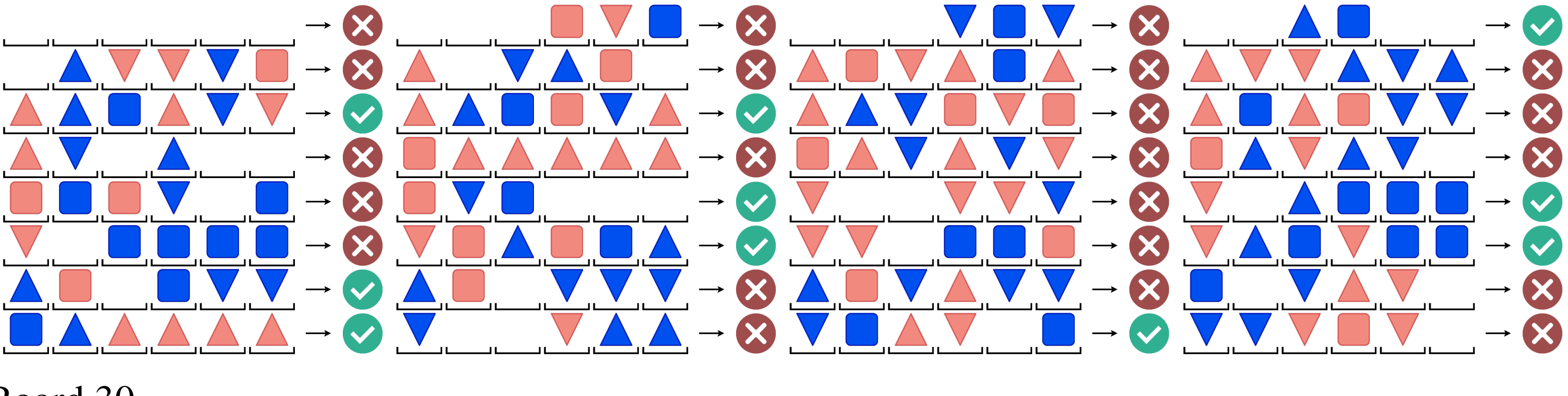}
\caption*{{Board 30}\\ {Golden rule}: ``exactly 1 blue pyramid touching blue block" \\ \textsc{CRN}: ``exactly 1 blue pyramid touching blue block", T-acc 1.0 \checkmark \\ \textsc{EMP-C}: ``exactly 1 blue pyramid touching block block", T-acc: 0.97 \checkmark \\ \textsc{EMP-R}: T-acc 0.79 \xmark}
\end{figure*}

\begin{figure*}[h]
\includegraphics[width=0.99\textwidth]{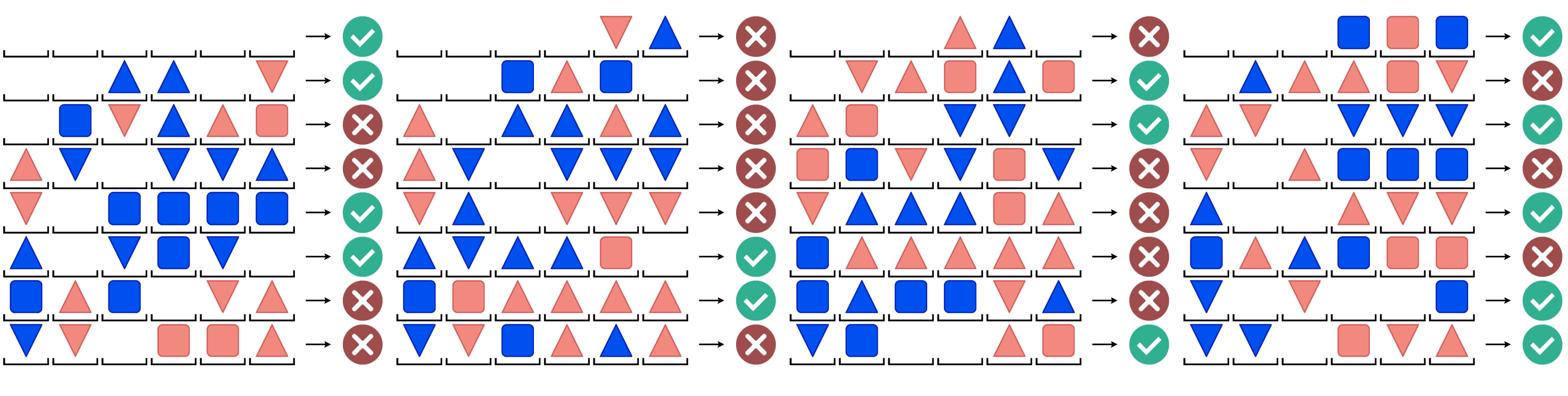}\caption*{{Board 75}\\ {Golden rule}: ``zero blue touching red pyramid" \\ \textsc{CRN}: ``zero blue touching red pyramid", T-acc 1.0 \checkmark \\ \textsc{EMP-C}: ``zero blue touching red", T-acc: 0.85 \xmark \\ \textsc{EMP-R}: T-acc 0.91 \checkmark}
\end{figure*}

\begin{figure*}[h]
\includegraphics[width=0.99\textwidth]{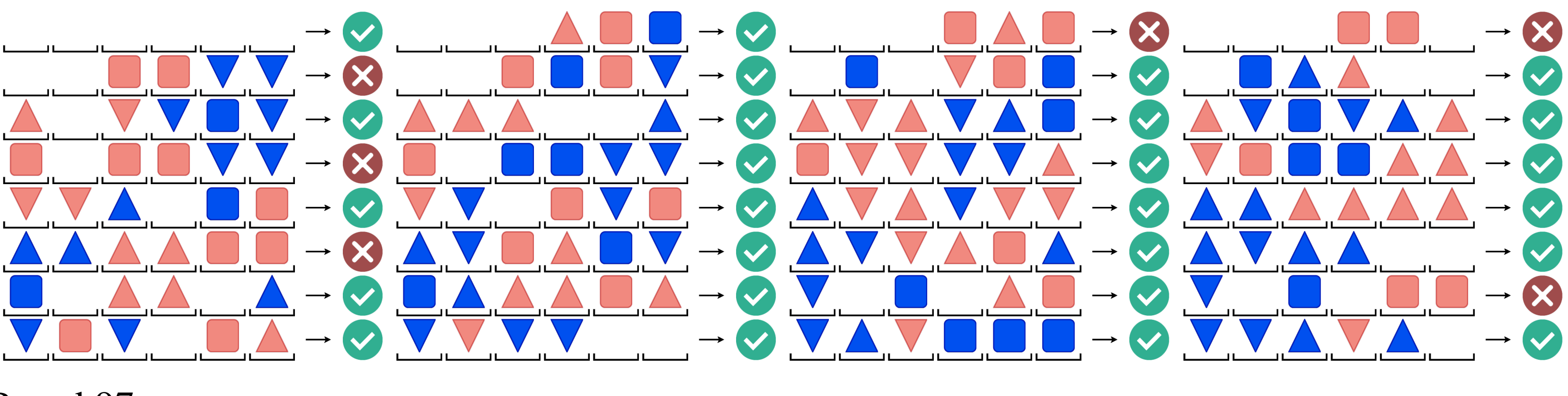}
\caption*{{Board 97} \\ {Golden rule}: ``at\_most 1 red block touching red" \\ \textsc{CRN}: ``at\_most 1 red block touching red", T-acc 1.0 \checkmark \\ \textsc{EMP-C}: ``at\_most 1 red touching at\_the\_right\_of red", T-acc: 0.98 \checkmark \\ \textsc{EMP-R}: T-acc 0.93 \xmark}
\end{figure*}

\begin{figure*}[h]
\includegraphics[width=0.99\textwidth]{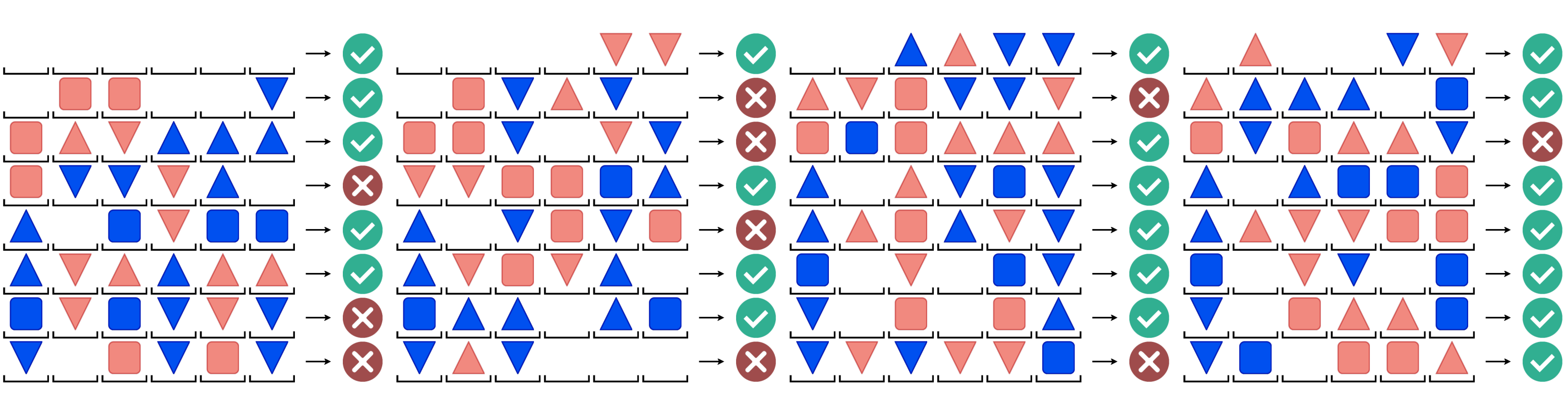}
\caption*{{Board 103} \\ {Golden rule}: ``at\_most 1 blue pyramid pointing\_down touching red" \\ \textsc{CRN}: ``at\_most 1 blue pyramid pointing\_down touching red", T-acc 1.0 \checkmark \\ \textsc{EMP-C}: ``at\_most 1 blue pyramid pointing\_down touching red", T-acc: 0.98 \checkmark \\ \textsc{EMP-R}: T-acc 0.85 \xmark}
\end{figure*}

%% file: Chapters/Chapter0A2.tex
\chapter{Appendix ASIF}

\label{sec-appendix}

\begin{figure}[h!]
    \centering
    \includegraphics[width=0.999\linewidth]{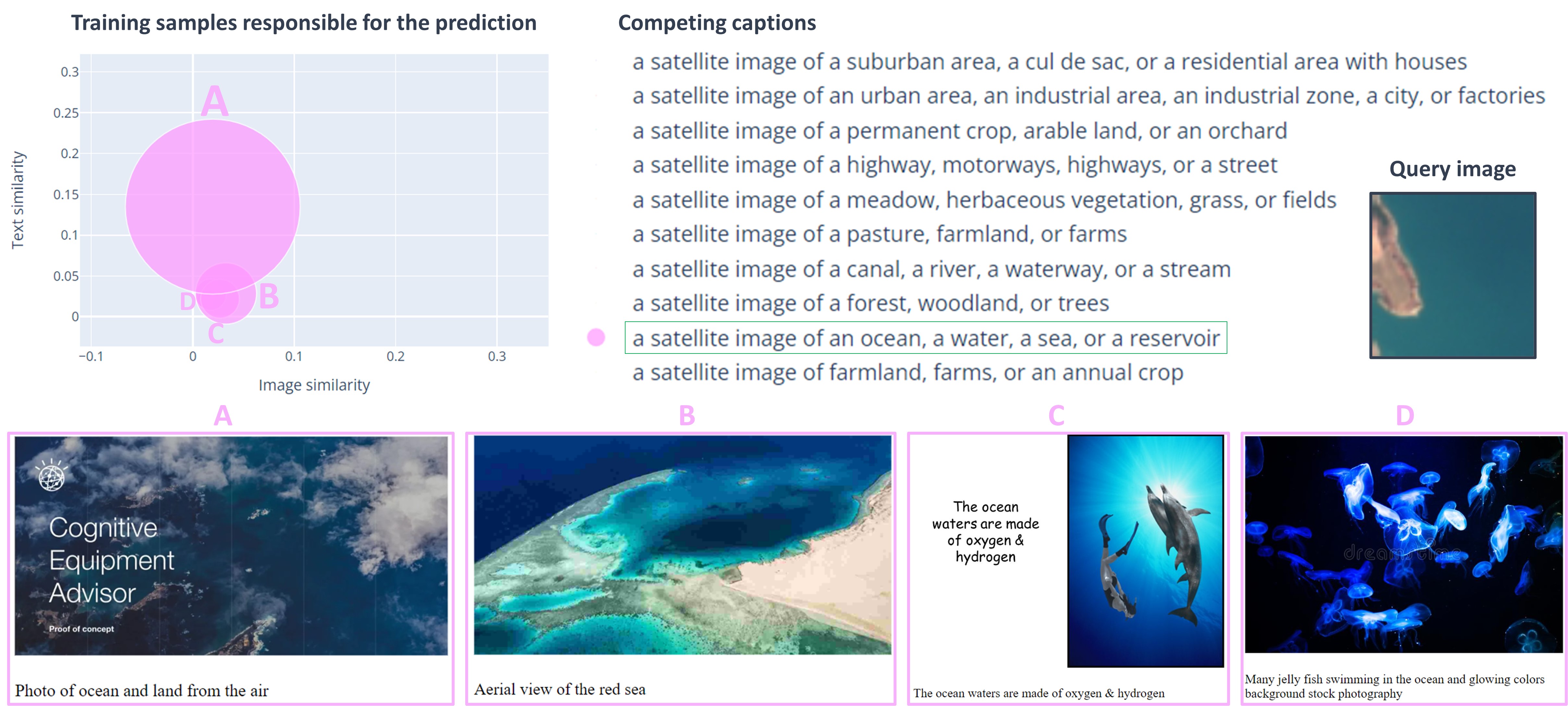}

  \includegraphics[width=0.999\linewidth]{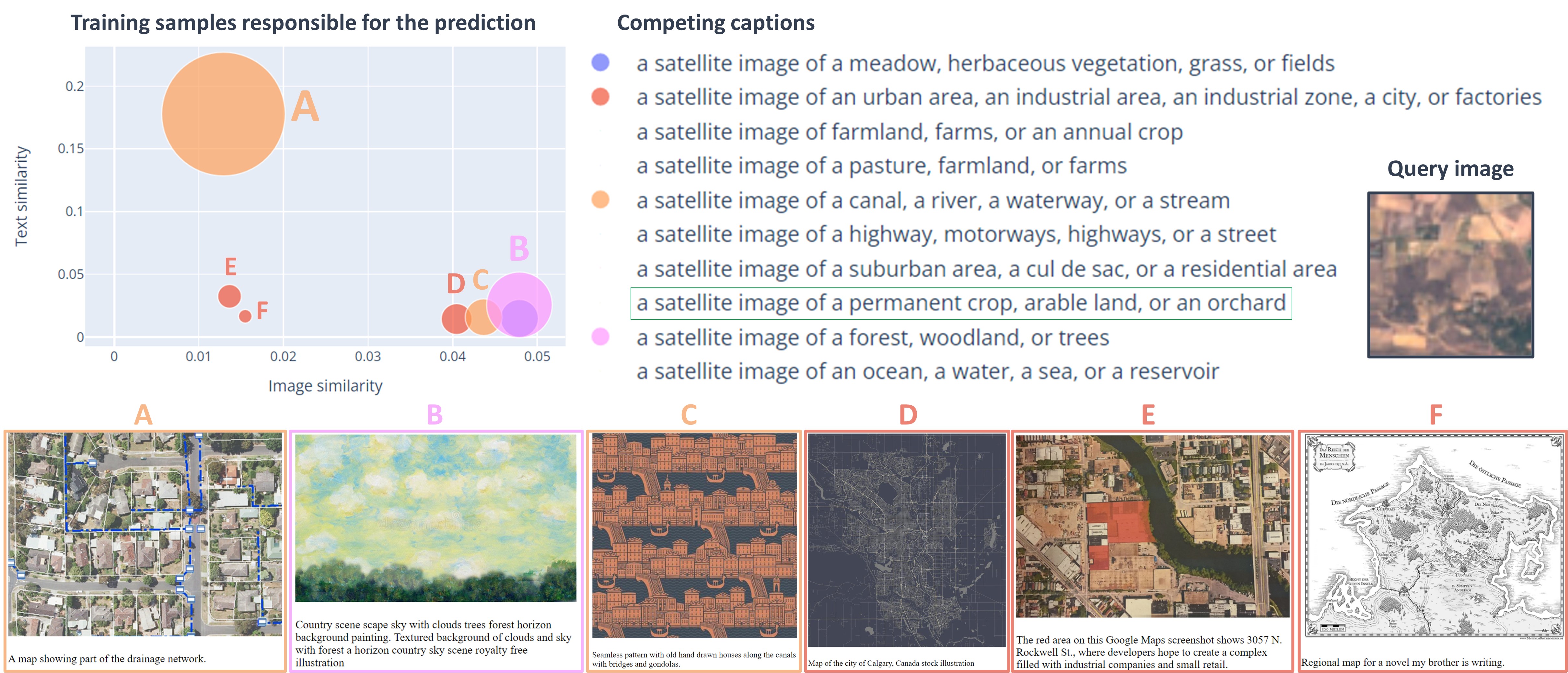}

            \caption[Interpretability of EuroSAT classifications through ASIF]{\textsc{Interpretability of EuroSAT classifications through ASIF.} Analysis of the classification outcome of two EuroSAT query images using ASIF. The scatter plot shows the samples in the training set closer to the query image and the candidate caption of the corresponding color. Image and text similarity are computed through cosine similarity in the visual space of DINO and the text space of SentenceT. The size of the marks is proportional to the product of the image and text similarity. The class chosen is the one with the largest total area. Below are shown the corresponding pairs from the training dataset CC12M. We can notice the distance between the EuroSAT dataset and the 1.6M samples of CC12M we used, many of the closest images are not from satellite and even then may have misleading descriptions, as image \textit{A} in the second example. An interactive version of this plot for any ASIF classification can be obtained using our code demo attached in the supplementary material.} 
    \label{fig-eurosat}
        \end{figure}

\begin{figure*}[t]
    \centering
    \includegraphics[width=\linewidth]{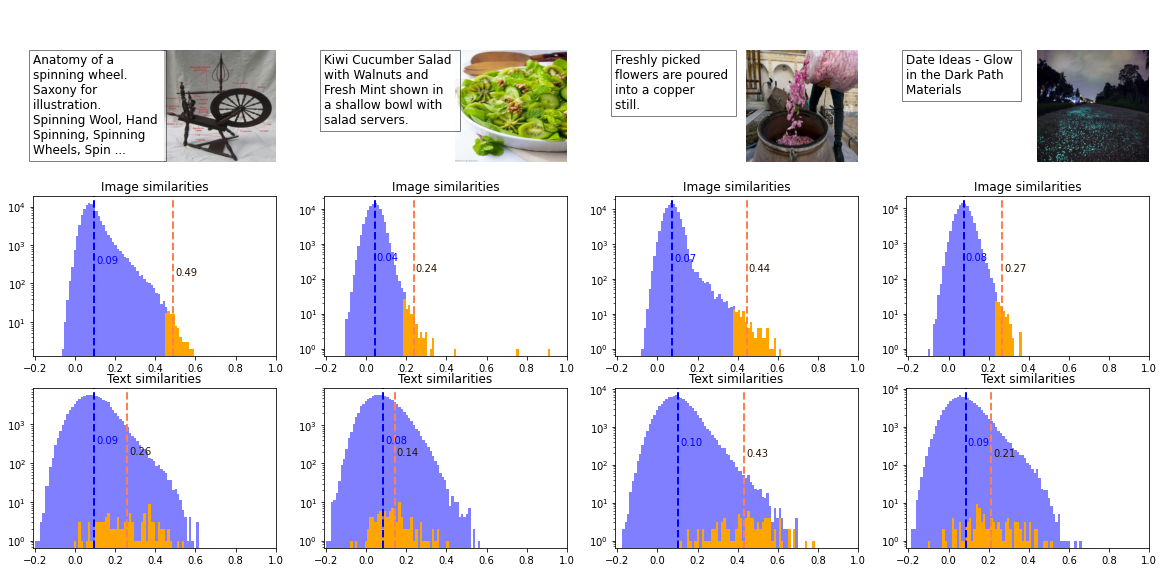}
    \includegraphics[width=\linewidth]{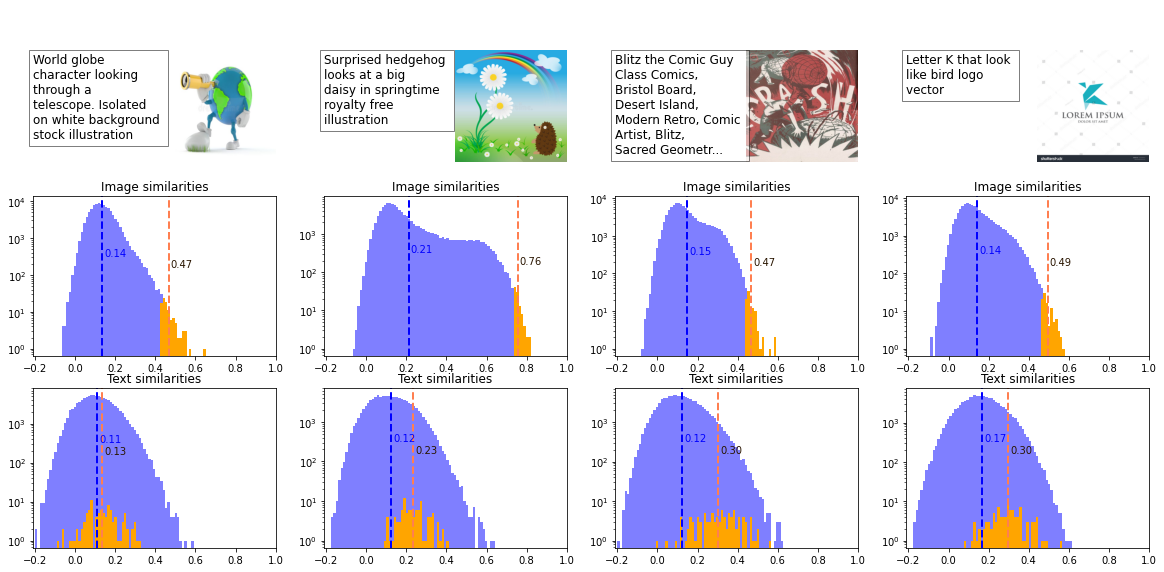}
    \caption[Captions of similar images are themselves similar]{\textsc{Caption of similar images are themselves similar.} For 8 image-text pairs, we show in the first row the distribution of the image similarities against $100k$ images in the train set in blue (CC12M), and highlight the $1000$ most similar in orange. The dashed lines indicate the mean of the two distributions. In the second row, we show the text similarities against the captions of the same $100k$ (blue) and $1000$ (orange) images.
    If captions of similar images are themselves similar, we expect the dashed orange line in the second row to be at the right of the blue dashed line, as we observe. The average gap between the orange and blue lines in the second row over 10,000 image-text couples from CC12M is $0.098 \pm 0.070$.}
    \label{fig:similarIm}
\end{figure*}

In this appendix, we further showcase the interpretability of ASIF models when used for classification in Figure \ref{fig-eurosat}. Then we provide additional details for the \textit{scaling laws} and \textit{EuroSAT} experiments presented in the main paper, and report additional results about the impact of the size of the encoders (Table \ref{tab-deit}), and of the image training dataset. 
  We also report further evidence that the ASIF construction is not overly sensitive to its hyperparameters.
Lastly, we discuss more in detail the idea that captions of similar images are alike in Figure \ref{fig:similarIm}.

\begin{figure}
    \centering
    \includegraphics[width=0.799\linewidth]{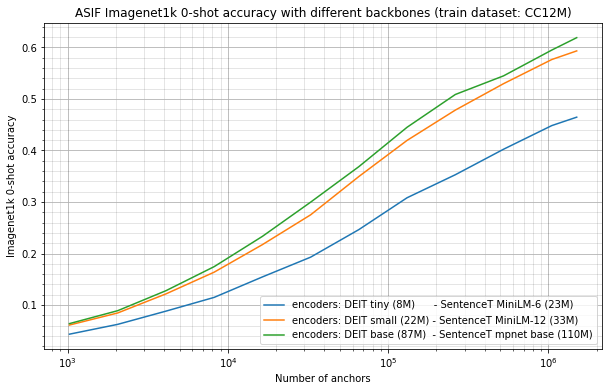}
    \includegraphics[width=0.799\linewidth]{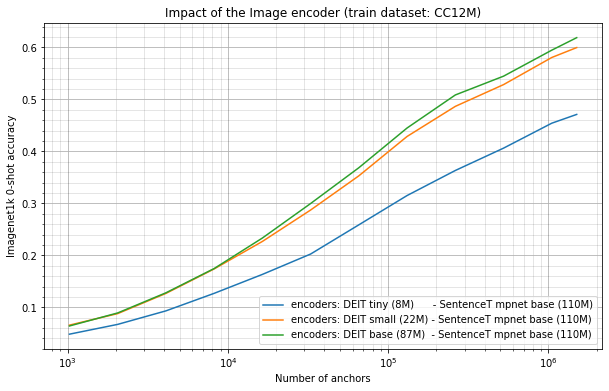}
    \includegraphics[width=0.799\linewidth]{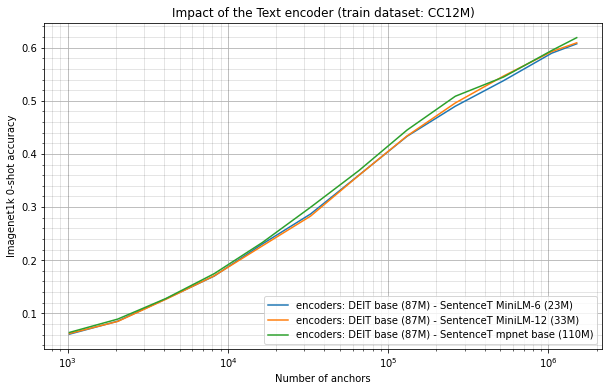}
    \caption[ASIF performance does not saturate earlier with smaller encoders]{\textsc{ASIF performance does not saturate earlier with smaller encoders.}  Classification accuracy keeps growing without saturating but is lower for smaller models. Furthermore, we observe that the quality of the vision encoder is more relevant than the quality of the text encoder with respect to zero-shot Imagenet classification.}
    \label{fig:impact-encoders}
\end{figure}

\section{Additional details on the scaling laws experiment}
\textsc{Models used in the scaling laws experiments.}
As discussed in the main paper, we tested ASIF with smaller image and text encoders to provide early evidence about ASIF scaling laws. We used three different instances of DEIT \cite{deit} vision transformers, the tiny (5.6M parameters, 192-dimensional embeddings), small (22M, 384), and base (87M, 768), and the original VITb16 vision transformer \cite{visualtransformers} (86M, 768). The DEIT models were pre-trained on a smaller dataset, the standard Imagenet1k training set \cite{deng2009imagenet}, while VITb16 was pretrained on Imagenet21k \cite{imagenet21k}. As text encoders, we used smaller versions of SentenceT \cite{sentencet}, with 23M and 33M parameters (both 384-dimensional embeddings), in contrast to the 110M parameters of the main model (768).

Figure \ref{fig:impact-encoders} shows that, with smaller encoders producing smaller embedddings, we do not observe a performance saturation within 1.6M image-text couples. Further experiments with larger datasets are left for future work.

\textsc{Impact of image pre-training data.} 
In Table \ref{tab-deit} we report the complete results of ASIF models using DEIT encoders \cite{deit}. We observe the expected positive correlation between the size of the encoders and the classification accuracy. Interestingly, ASIF with the largest instance of DEIT beats the one based on the standard VIT pretrained on Imagenet21k on three out of four of test datasets, while losing more than 10 points on CIFAR. These results may be interpreted in light of the similarity of the datasets we are using, with features useful to classify CIFAR images less overlapping with Imagenet1k features with respect to the other datasets.

\section{Additional details on the EuroSAT experiment.} EuroSAT, a renowned benchmark for satellite image classification, serves as a testing ground for out-of-distribution generalization in zero-shot and few-shot scenarios \cite{helber2019eurosat}. The dataset contains 27,000 images labeled under ten categories. Our ASIF model with a DINO visual backbone (denoted as 'ASIF unsup' in table \ref{tab-main}) achieved a zero-shot classification score of $29.4\%$. While significantly better than random chance, this modest performance is not surprising considering the scarce presence of satellite images in the CC12M dataset. 

As a further experiment, we randomly selected 100 images from the EuroSAT dataset and incorporated them into our ASIF training set, raising the total to 1,500,100 image-text pairs and leaving 26,900 images for testing. We created captions for the EuroSAT images using the template ``\textit{a satellite image of} [CLASS NAME]''. This way the ASIF model improves dramatically, reaching a classification accuracy of $82.5 \pm 2.8\%$ on EuroSAT (average $\pm$ standard deviation of 5 trials). 

Contrarily, CLIP \cite{clip}, while demonstrating better zero-shot accuracy at $54.1\%$, is trained on a private dataset comprising 400 million images. This dataset may contain a larger number of satellite images than our 1.6 million subset of CC12M. Given the substantial improvement observed when we added just 100 EuroSAT images, it's reasonable to speculate that CLIP's enhanced performance might stem from its larger database of satellite images. However, confirming this theory is impossible due to the private nature of CLIP's training set.

We can, nevertheless, examine the presence of satellite images in the CC12M dataset. Using ASIF models' unique interpretability property, we can trace the training samples behind each classification. Figure \ref{fig-eurosat} displays two EuroSAT samples, one classified correctly and the other not, along with the corresponding CC12M pairs responsible for the classifications. We note that our subset of CC12M is lacking in satellite images, and the few available often have misleading captions, such as a map of a drainage network tagged as "a satellite image of a canal, a river, a waterway, or a stream" instead of an urban area. 

The images shown are an adaptation of the interactive plot to analyze any ASIF image classification  we provided in the code demo attached in the supplementary material.

\begin{table*}[t]
\centering
\begin{tabular}{lcccc}

\toprule
\textsc{ASIF backbones} & \textsc{ImgNet} & \textsc{CIFAR} & \textsc{Oxford} & \textsc{ImgNet} \\
\textsc{(Params, pre-training data)} & & \textsc{100} & \textsc{Pets} & \textsc{v2} \\
\midrule

DEITtiny {\tiny( 5.6M, Im1k)} - STminiL6 {\tiny( 23M)} & 46.5 & 37.3 & 75.6 & 38.3 \\
DEITsmall {\tiny( 22M, Im1k)} - STminiL12 {\tiny( 33M)} & 59.3 & 46.0 & 80.4 & 50.3 \\
DEITbase {\tiny( 87M, Im1k)} - STbase {\tiny( 110M)} & 60.9 & 50.2 & 81.5 & 52.2 \\
VITb16 {\tiny( 86M, Im21k)} - STbase {\tiny( 110M)} & 55.4 & 63.3 & 71.5 & 45.6 \
\end{tabular}
\caption[Zero shot classification accuracy of ASIF models with different backbones]{\textsc{Zero shot classification accuracy of ASIF models with different backbones}. We observe that the ASIF procedure remains effective even with smaller encoders pre-trained on reduced visual datasets such as Imagenet1k.}
\label{tab-deit}
\end{table*}

\section{ASIF sensibility to its hyperparameters}
\begin{figure}[ht]
\centering
\includegraphics[width=0.69\linewidth]{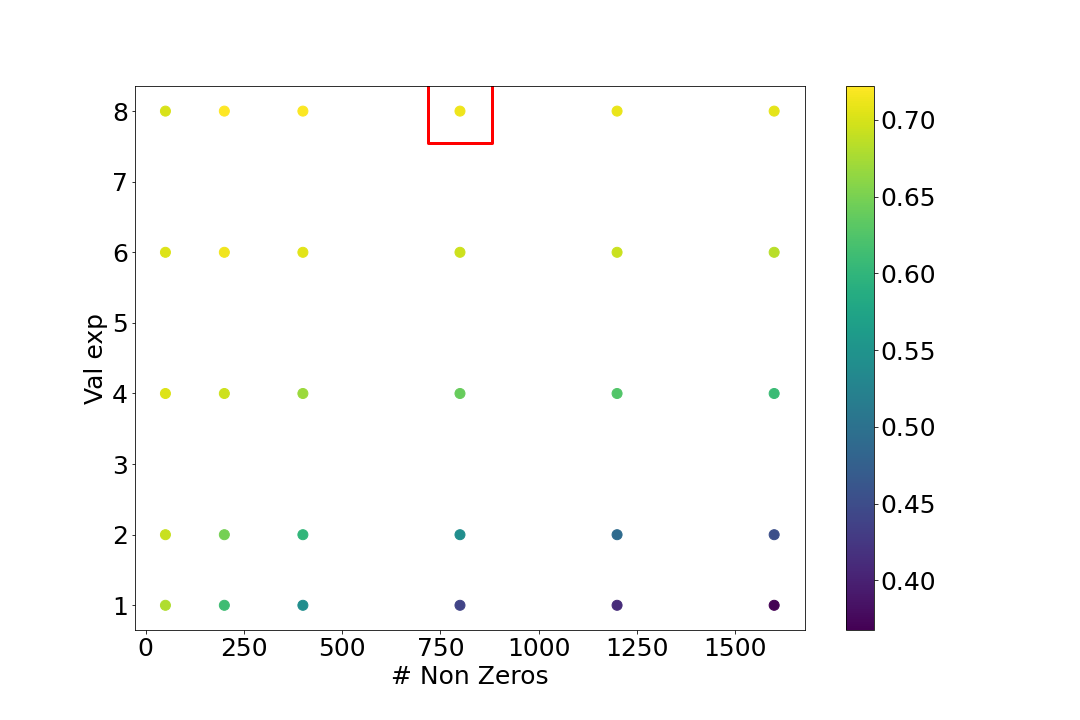}
\includegraphics[width=0.69\linewidth]{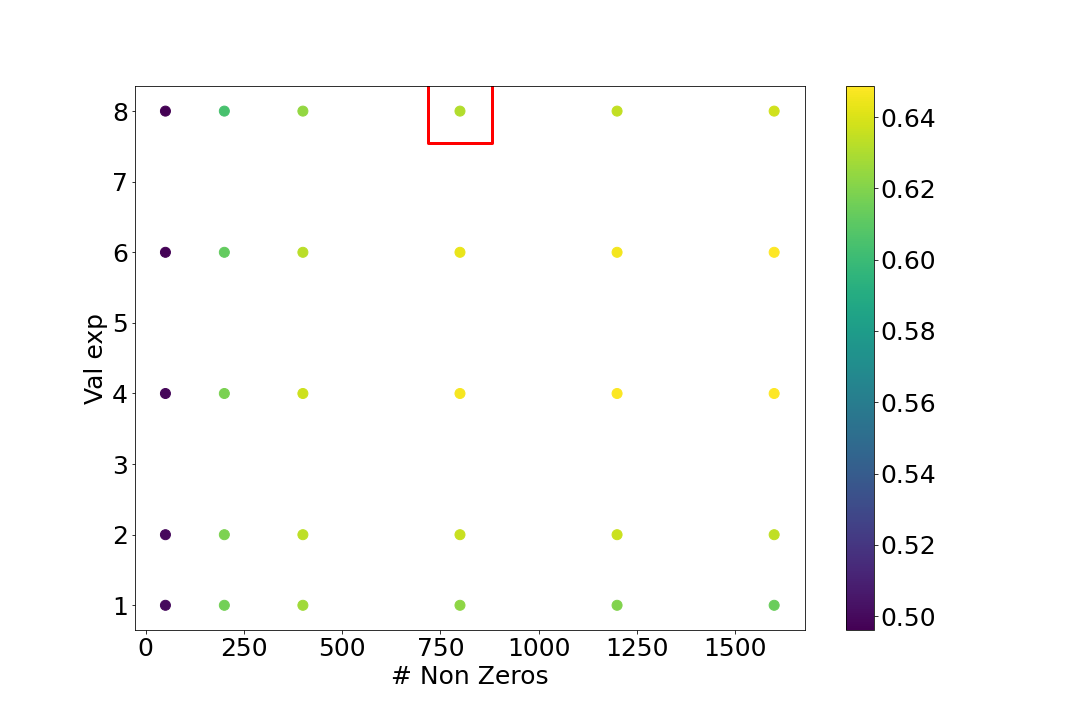}
    \caption[ASIF Hyperparameters search]{\textsc{ASIF Hyperparameters search} over Pets (up) and CIFAR100 (down). Highlighted in the red square the performance achieved tuning on Imagenet1K. 
    }
    \label{fig:hyperparamsearch}
\end{figure}

Finally, we present evidence about the sensitivity of the ASIF model to the hyperparameters $p$ and $k$. Specifically, we show the hyperparameter search for PETS and CIFAR100 in Figure \ref{fig:hyperparamsearch}. Table \ref{tab-hyperparam} with results on the parameters fine-tuned on the two datasets reveals marginal improvements over the standard choice of k=800 and p=8. This suggests that the ASIF model is relatively insensitive to the choice of these hyperparameters.

\begin{table}[h]
\centering
\begin{tabular}{lccc}
\textsc{Tuned on} & \textsc{Parameters $p$,$k$}      & \textsc{CIFAR} & \textsc{PETS}\\
\hline
PETS &(200,8)  & 60.9   & 72.3     \\  CIFAR & (1600,6) & 64.9     & 63      \\
ImageNet1K & (800,8) & 63.3    & 71.5  \\
\hline
\end{tabular}
\caption[ASIF Hyperparams search]{ASIF Hyperparams search: tuning on each dataset per row.}
\label{tab-hyperparam}
\end{table}

%% file: Bibliography.bib
@online{amodei2018ai,
 author = {Amodei, Dario and Hernandez, Danny},
 title = {{AI and Compute}},
 url = {https://openai.com/blog/ai-and-compute/},
 year = {2018}
}

@inproceedings{buro1995logistello,
 author = {Buro, Michael},
 booktitle = {19th Annual Conference Gesellschaft f{\"u}r Klassifikation eV},
organization = {Citeseer},
 title = {Logistello: A strong learning othello program},
 volume = {2},
 year = {1995}
}

@inproceedings{chang2018big,
 author = {Chang, Nai-Yuan and Chen, Chih-Hung and Lin, Shun-Shii and Nair, Surag},
 booktitle = {Proc. ICML},
 pages = {78--81},
 title = {The Big Win Strategy on Multi-Value Network: An Improvement over AlphaZero Approach for 6x6 Othello},
 year = {2018}
}

@article{chollet2019measure,
 author = {Chollet, Fran{\c{c}}ois},
 journal = {arXiv preprint arXiv:1911.01547},
 title = {The Measure of Intelligence},
 year = {2019}
}

@article{colaboratory_carneiro2018performance,
 author = {Carneiro, Tiago and Da N{\'o}brega, Raul Victor Medeiros and Nepomuceno, Thiago and Bian, Gui-Bin and De Albuquerque, Victor Hugo C and Reboucas Filho, Pedro Pedrosa and X, X},
 journal = {IEEE Access},
 pages = {61677--61685},
 publisher = {IEEE},
 title = {Performance analysis of google colaboratory as a tool for accelerating deep learning applications},
 volume = {6},
 year = {2018}
}

@online{connect4MediumLessonsAlphaZero,
 author = {Young, Anthony and Prasad, Aditya and Abrams, Ishaya},
 title = {{Lessons From Implementing Alpha Zero}},
 url = {https://link.medium.com/ylGDD6F7V9},
 urldate = {2019-05-27},
 year = {2018}
}

@online{Edax,
 author = {Delorme, Richard},
 title = {Edax},
 url = {https://github.com/abulmo/edax-reversi},
 urldate = {2020-09-19},
 year = {1998}
}

@article{ha2018world,
 author = {Ha, David and Schmidhuber, J{\"u}rgen},
 journal = {arXiv preprint arXiv:1803.10122},
 title = {World models},
 year = {2018}
}

@inproceedings{He2015DeepRecognition,
  author={He, Kaiming and Zhang, Xiangyu and Ren, Shaoqing and Sun, Jian},
  title={Deep residual learning for image recognition},
  booktitle={Proceedings of the IEEE conference on computer vision and pattern recognition},
  pages={770--778},
  year={2016}
}

@article{Kasparov2018ChessReasoning.,
 author = {Kasparov, Garry},
 journal = {Science (New York, N.Y.)},
 number = {6419},
 pages = {1087},
 publisher = {American Association for the Advancement of Science},
 title = {{Chess, a Drosophila of reasoning.}},
 volume = {362},
 year = {2018}
}

@article{katago-wu2019accelerating,
 author = {Wu, David J},
 journal = {arXiv preprint arXiv:1902.10565},
 title = {Accelerating Self-Play Learning in Go},
 year = {2019}
}

@article{knuth1975analysis,
 author = {Knuth, Donald E and Moore, Ronald W},
 journal = {Artificial intelligence},
 number = {4},
 pages = {293--326},
 publisher = {Elsevier},
 title = {An analysis of alpha-beta pruning},
 volume = {6},
 year = {1975}
}

@article{LeCun2015DeepLearning,
  title={Deep learning},
  author={LeCun, Yann and Bengio, Yoshua and Hinton, Geoffrey},
  journal={nature},
  volume={521},
  number={7553},
  pages={436--444},
  year={2015},
  publisher={Nature Publishing Group UK London}
}

@inproceedings{leeminigo,
 author = {Lee, Brian and Jackson, Andrew and Madams, Tom and Troisi, Seth and Jones, Derek and X, X and Y, Y},
 booktitle = {RML@ICLR},
 title = {Minigo: A Case Study in Reproducing Reinforcement Learning Research},
 year = {2019}
}

@article{liskowski2018learning,
 author = {Liskowski, Pawe{\l} and Ja{\'s}kowski, Wojciech and Krawiec, Krzysztof},
 journal = {IEEE Transactions on Games},
 number = {4},
 pages = {354--364},
 publisher = {IEEE},
 title = {Learning to play othello with deep neural networks},
 volume = {10},
 year = {2018}
}

@article{mcts-browne2012survey,
 author = {Browne, Cameron B and Powley, Edward and Whitehouse, Daniel and Lucas, Simon M and Cowling, Peter I and Rohlfshagen, Philipp and Tavener, Stephen and Perez, Diego and Samothrakis, Spyridon and Colton, Simon},
 journal = {IEEE Transactions on Computational Intelligence and AI in games},
 number = {1},
 pages = {1--43},
 publisher = {IEEE},
 title = {{A survey of Monte Carlo Tree Search Methods}},
 volume = {4},
 year = {2012}
}

@article{mullins2007checkers,
  title={Checkers is solved},
  author={Schaeffer, Jonathan and Burch, Neil and Bj{\"o}rnsson, Yngvi and Kishimoto, Akihiro and M{\"u}ller, Martin and Lake, Robert and Lu, Paul and Sutphen, Steve},
  journal={science},
  volume={317},
  number={5844},
  pages={1518--1522},
  year={2007},
  publisher={American Association for the Advancement of Science}
}

@article{muzero-schrittwieser2019mastering,
  title={Mastering atari, go, chess and shogi by planning with a learned model},
  author={Schrittwieser, Julian and Antonoglou, Ioannis and Hubert, Thomas and Simonyan, Karen and Sifre, Laurent and Schmitt, Simon and Guez, Arthur and Lockhart, Edward and Hassabis, Demis and Graepel, Thore and others},
  journal={Nature},
  volume={588},
  number={7839},
  pages={604--609},
  year={2020},
  publisher={Nature Publishing Group}
}

@article{neumann1928theorie,
 author = {Neumann, John von},
 journal = {Mathematische annalen},
 number = {1},
 pages = {295--320},
 publisher = {Springer},
 title = {Zur theorie der gesellschaftsspiele (German)},
 volume = {100},
 year = {1928}
}

@article{pachockiopenai,
 author = {Pachocki, Jakub and Brockman, Greg and Raiman, Jonathan and Zhang, Susan and Pond{\'e}, Henrique and Tang, Jie and Wolski, Filip and Dennison, Christy and Jozefowicz, Rafal and Debiak, Przemyslaw and others},
 title = {{OpenAI Five}},
 url = {https://openai.com/blog/openai-five/},
 year = {2018}
}

@phdthesis{RomanoBenedetto2009SAIO:Dellothello,
 author = {Romano, Benedetto},
 school = {University of Naples Federico II},
 title = {{SAIO: Un sistema esperto per il gioco dell'othello (Italian)}},
 url = {http://www.romanobenedetto.it/tesi.pdf},
 year = {2009}
}

@inproceedings{segal2010scalability,
 author = {Segal, Richard B},
 booktitle = {International Conference on Computers and Games},
 organization = {Springer},
 pages = {36--47},
 title = {On the scalability of parallel UCT},
 year = {2010}
}

@article{Silver2017MasteringKnowledge,
 author = {Silver, David and Schrittwieser, Julian and Simonyan, Karen and Antonoglou, Ioannis and Huang, Aja and Guez, Arthur and Hubert, Thomas and Baker, Lucas and Lai, Matthew and Bolton, Adrian and Chen, Yutian and Lillicrap, Timothy and Hui, Fan and Sifre, Laurent and van den Driessche, George and Graepel, Thore and Hassabis, Demis},
 journal = {Nature},
 number = {7676},
 pages = {354--359},
 publisher = {Nature Publishing Group},
 title = {{Mastering the game of Go without human knowledge}},
 volume = {550},
 year = {2017}
}

@article{Silver2018ASelf-play.,
 author = {Silver, David and Hubert, Thomas and Schrittwieser, Julian and Antonoglou, Ioannis and Lai, Matthew and Guez, Arthur and Lanctot, Marc and Sifre, Laurent and Kumaran, Dharshan and Graepel, Thore and Lillicrap, Timothy and Simonyan, Karen and Hassabis, Demis},
 journal = {Science},
 number = {6419},
 pages = {1140--1144},
 publisher = {American Association for the Advancement of Science},
 title = {{A general reinforcement learning algorithm that masters chess, shogi, and Go through self-play.}},
 volume = {362},
 year = {2018}
}

@inproceedings{tian2019elf,
  title={Elf opengo: An analysis and open reimplementation of alphazero},
  author={Tian, Yuandong and Ma, Jerry and Gong, Qucheng and Sengupta, Shubho and Chen, Zhuoyuan and Pinkerton, James and Zitnick, Larry},
  booktitle={International Conference on Machine Learning},
  pages={6244--6253},
  year={2019},
  organization={PMLR}
}

@article{vinyals2019grandmaster,
 author = {Vinyals, Oriol and Babuschkin, Igor and Czarnecki, Wojciech M and Mathieu, Micha{\"e}l and Dudzik, Andrew and Chung, Junyoung and Choi, David H and Powell, Richard and Ewalds, Timo and Georgiev, Petko and others},
 journal = {Nature},
 number = {7782},
 pages = {350--354},
 publisher = {Nature Publishing Group},
 title = {{Grandmaster level in StarCraft II using multi-agent reinforcement learning}},
 volume = {575},
 year = {2019}
}

@article{wang2019hyper,
 author = {Wang, Hui and Emmerich, Michael and Preuss, Mike and Plaat, Aske},
 journal = {arXiv preprint arXiv:1903.08129},
 title = {Hyper-Parameter Sweep on AlphaZero General},
 year = {2019}
}

@article{wu2020population-based-accelerating,
 author = {Wu, Ti-Rong and Wei, Ting-Han and Wu, I and others},
 journal = {arXiv:2003.06212},
 title = {Accelerating and Improving AlphaZero Using Population Based Training},
 year = {2020}
}

@online{zebra,
 author = {Andersson, Gunnar},
 title = {Zebra},
 url = {https://web.archive.org/web/20200226073330/http://radagast.se/othello/zebra.html},
 urldate = {2020-02-26},
 year = {1997}
}

@misc{zendo,
author={Kory Heath},
title={Zendo},
year={2001},
url={http://www.koryheath.com/zendo/},
}

@inproceedings{zero,
author = {Blum, Manuel and Feldman, Paul and Micali, Silvio},
title = {Non-Interactive Zero-Knowledge and Its Applications},
year = {1988},
isbn = {0897912640},
publisher = {Association for Computing Machinery},
address = {New York, NY, USA},
url = {https://doi.org/10.1145/62212.62222},
doi = {10.1145/62212.62222},
booktitle = {Proceedings of the Twentieth Annual ACM Symposium on Theory of Computing},
pages = {103–112},
numpages = {10},
location = {Chicago, Illinois, USA},
series = {STOC '88}
}

@ARTICLE {hospedales,
author = {T. M. Hospedales and A. Antoniou and P. Micaelli and A. J. Storkey},
journal = {IEEE Transactions on Pattern Analysis \& Machine Intelligence},
title = {Meta-Learning in Neural Networks: A Survey},
year = {2020},
issn = {1939-3539},
doi = {10.1109/TPAMI.2021.3079209},
publisher = {IEEE Computer Society},
address = {Los Alamitos, CA, USA},
month = {may}
}

@inproceedings{clip,
  title={Learning Transferable Visual Models From Natural Language Supervision},
  author={Radford, Alec and Kim, Jong Wook and Hallacy, Chris and Ramesh, Aditya and Goh, Gabriel and Agarwal, Sandhini and Sastry, Girish and Askell, Amanda and Mishkin, Pamela and Clark, Jack and others},
  booktitle={Proc. ICML},
  year={2021},
}

@inproceedings{clevr,
    title = {{CLEVR: A Diagnostic Dataset for Compositional Language and Elementary Visual Reasoning}},
    year = {2017},
    booktitle = {Proc. CVPR},
    author = {Johnson, Justin and Hariharan, Bharath and van der Maaten, Laurens and Fei-Fei, Li and Zitnick, C. Lawrence and Girshick, Ross},
    pages = {1988--1997},
}

@article{Angluin,
author = {Angluin, Dana},
title = {Learning Regular Sets from Queries and Counterexamples},
year = {1987},
issue_date = {November 1, 1987},
publisher = {Academic Press, Inc.},
address = {USA},
volume = {75},
number = {2},
issn = {0890-5401},
url = {https://doi.org/10.1016/0890-5401(87)90052-6},
doi = {10.1016/0890-5401(87)90052-6},
journal = {Inf. Comput.},
month = nov,
pages = {87–106},
numpages = {20}
}

@article{Taylor16389,
	author = {Taylor, Alex H. and Miller, Rachael and Gray, Russell D.},
	title = {New Caledonian crows reason about hidden causal agents},
	volume = {109},
	number = {40},
	pages = {16389--16391},
	year = {2012},
	journal = {PNAS}
}

@article{mcconnell1962memory,
  title={Memory transfer through cannibalism in planarians},
  author={McConnell, James},
  journal={J. Neuropsychiat.},
  volume={3},
  pages={542--548},
  year={1962}
}

@article{microscope,
  author = {Goh, Gabriel and Cammarata, Nick and Voss, Chelsea and Carter, Shan and Petrov, Michael and Schubert, Ludwig and Radford, Alec and Olah, Chris},
  title = {Multimodal Neurons in Artificial Neural Networks},
  journal = {Distill},
  year = {2021},
  note = {https://distill.pub/2021/multimodal-neurons},
  doi = {10.23915/distill.00030}
}

@inproceedings{hendricks2016generating,
  title={Generating visual explanations},
  author={Hendricks, Lisa Anne and Akata, Zeynep and Rohrbach, Marcus and Donahue, Jeff and Schiele, Bernt and Darrell, Trevor},
  booktitle={European conference on computer vision},
  pages={3--19},
  year={2016},
  organization={Springer}
}

@article{meinke2019towards,
  title={Towards neural networks that provably know when they don't know},
  author={Meinke, Alexander and Hein, Matthias},
  journal={arXiv preprint arXiv:1909.12180},
  year={2019}
}

@article{aaronson2013philosophers,
  title={Why philosophers should care about computational complexity},
  author={Aaronson, Scott},
  journal={Computability: Turing, G{\"o}del, Church, and Beyond},
  volume={261},
  pages={327},
  year={2013},
  publisher={Cambridge, MA: The MIT Press.[Aaronson 2013 preprint available online]}
}

@article{santoro2021symbolic,
  author={Santoro, Adam and Lampinen, Andrew and Mathewson, Kory and Lillicrap, Timothy and Raposo, David},
  title={Symbolic behaviour in artificial intelligence},
  journal={arXiv preprint arXiv:2102.03406},
  year={2021}
}

@book{shapiro1981inductive,
  title={Inductive inference of theories from facts},
  author={Shapiro, Ehud Y},
  year={1981},
  publisher={Yale University, Department of Computer Science}
}

@book{deutsch2011beginning,
  title={The beginning of infinity: Explanations that transform the world},
  author={Deutsch, David},
  year={2011},
  publisher={Penguin UK}
}

@inproceedings{Vaswani2017,
author = {Vaswani, Ashish and Shazeer, Noam and Parmar, Niki and Uszkoreit, Jakob and Jones, Llion and Gomez, Aidan N. and Kaiser, \L{}ukasz and Polosukhin, Illia},
title = {Attention is All You Need},
year = {2017},
isbn = {9781510860964},
publisher = {Curran Associates Inc.},
address = {Red Hook, NY, USA},
booktitle = {Proceedings of the 31st International Conference on Neural Information Processing Systems},
pages = {6000–6010},
numpages = {11},
location = {Long Beach, California, USA},
series = {NIPS'17}
}

@article{chinese-room, 
title={Minds, brains, and programs}, 
volume={3}, 
DOI={10.1017/S0140525X00005756}, 
number={3}, 
journal={Behavioral and Brain Sciences}, 
publisher={Cambridge University Press}, 
author={Searle, John R.}, 
year={1980}, 
pages={417–424}
}

@book{popper1935logic,
  title={The Logic of Scientific Discovery},
  author={Popper, Karl},
  year={1935},
  publisher={Julius Springer, Hutchinson \& Co}
}

@inproceedings{balog2019deepcoder,
  title={DeepCoder: Learning to write programs},
  author={Balog, M and Gaunt, AL and Brockschmidt, M and Nowozin, S and Tarlow, D},
  booktitle={5th International Conference on Learning Representations, ICLR 2017-Conference Track Proceedings},
  year={2017}
}

@article{ellis2020dreamcoder,
  title={Dreamcoder: Growing generalizable, interpretable knowledge with wake-sleep bayesian program learning},
  author={Ellis, Kevin and Wong, Catherine and Nye, Maxwell and Sable-Meyer, Mathias and Cary, Luc and Morales, Lucas and Hewitt, Luke and Solar-Lezama, Armando and Tenenbaum, Joshua B},
  journal={arXiv preprint arXiv:2006.08381},
  year={2020}
}

@article{Rota-pernicious,
	publisher = {Springer},
	title = {The Pernicious Influence of Mathematics Upon Philosophy},
	author = {Gian{-}Carlo Rota},
	year = {1991},
	doi = {10.1007/BF00567744},
	pages = {165--178},
	volume = {88},
	number = {2},
	journal = {Synthese}
}

@article{shen2021generate,
  title={Generate \& Rank: A Multi-task Framework for Math Word Problems},
  author={Shen, Jianhao and Yin, Yichun and Li, Lin and Shang, Lifeng and Jiang, Xin and Zhang, Ming and Liu, Qun},
  journal={arXiv preprint arXiv:2109.03034},
  year={2021}
}

@article{mitchell2021abstraction,
  title={Abstraction and analogy-making in artificial intelligence},
  author={Mitchell, Melanie},
  journal={arXiv preprint arXiv:2102.10717},
  year={2021}
}

@book{hofstadter1979godel,
  title={G{\"o}del, Escher, Bach: an eternal golden braid},
  author={Douglas Hofstadter},
  volume={13},
  year={1979},
  publisher={Basic books New York}
}

@misc{alex2017shapeworld,
    title={ShapeWorld - A new test methodology for multimodal language understanding},
    author={Alexander Kuhnle and Ann Copestake},
    year={2017},
    eprint={1704.04517},
    archivePrefix={arXiv},
    primaryClass={cs.CL}
}

@article{andreas2017learning,
  title={Learning with latent language},
  author={Andreas, Jacob and Klein, Dan and Levine, Sergey},
  journal={arXiv preprint arXiv:1711.00482},
  year={2017}
}

@article{pearl2021radical,
  title={Radical empiricism and machine learning research},
  author={Pearl, Judea},
  journal={Journal of Causal Inference},
  volume={9},
  number={1},
  pages={78--82},
  year={2021},
  publisher={De Gruyter}
}

@misc{epistemologyLecun,
  author={Yann LeCun},
  title = {The Epistemology of Deep Learning},
  howpublished = {\emph{Institute for Advanced Studies} \url{https://www.ias.edu/sites/default/files/video/lecun-ias-20190222.pdf}; \url{https://youtu.be/gG5NCkMerHU}},
  year = {2019},
  note = {Accessed: {2021–10-04}}
}

@book{mitchell1980need,
  title={The need for biases in learning generalizations},
  author={Mitchell, Tom M},
  year={1980},
  publisher={Department of Computer Science, Laboratory for Computer Science Research}
}

@inproceedings{bender-koller-2020-climbing,
    title = "Climbing towards {NLU}: {On} Meaning, Form, and Understanding in the Age of Data",
    author = "Bender, Emily M.  and
      Koller, Alexander",
    booktitle = "Proceedings of the 58th Annual Meeting of the Association for Computational Linguistics",
    month = jul,
    year = "2020",
    address = "Online",
    publisher = "Association for Computational Linguistics",
    url = "https://aclanthology.org/2020.acl-main.463",
    doi = "10.18653/v1/2020.acl-main.463",
    pages = "5185--5198",
}

@article{schulz2007preschool,
    title = {{Preschool children learn about causal structure from conditional interventions}},
    year = {2007},
    journal = {Developmental science},
    author = {Schulz, Laura E and Gopnik, Alison and Glymour, Clark},
    number = {3},
    pages = {322--332},
    volume = {10},
    publisher = {Wiley Online Library}
}

@inproceedings{hind2019ted,
  title={TED: Teaching AI to explain its decisions},
  author={Hind, Michael and Wei, Dennis and Campbell, Murray and Codella, Noel CF and Dhurandhar, Amit and Mojsilovi{\'c}, Aleksandra and Natesan Ramamurthy, Karthikeyan and Varshney, Kush R},
  booktitle={Proceedings of the 2019 AAAI/ACM Conference on AI, Ethics, and Society},
  pages={123--129},
  year={2019}
}

@book{galilei1610sidereus,
  title={Sidereus Nuncius, Or The Sidereal Messenger},
  author={Galilei, Galileo},
  year={2016},
  publisher = "University of Chicago Press",
    url = "http://people.reed.edu/~wieting/mathematics537/SideriusNuncius.pdf",
}

@article{peirce-clear,
  title={How to make our ideas clear},
  author={Peirce, Charles Sanders},
  journal={Popular Science Monthly},
  volume={12},
  pages={286--302},
  year={1878}
}

@article{rudin2019stop,
  title={Stop explaining black box machine learning models for high stakes decisions and use interpretable models instead},
  author={Rudin, Cynthia},
  journal={Nature Machine Intelligence},
  volume={1},
  number={5},
  pages={206--215},
  year={2019},
  publisher={Nature Publishing Group}
}

@book{eco-kant,
  title={Kant and the platypus: Essays on language and cognition},
  author={Eco, Umberto},
  year={2000},
  publisher={HMH}
}

@inproceedings{finn2017model,
  title={Model-agnostic meta-learning for fast adaptation of deep networks},
  author={Finn, Chelsea and Abbeel, Pieter and Levine, Sergey},
  booktitle={International Conference on Machine Learning},
  pages={1126--1135},
  year={2017},
  organization={PMLR}
}

@InProceedings{Lee_2019_CVPR,
author = {Lee, Kwonjoon and Maji, Subhransu and Ravichandran, Avinash and Soatto, Stefano},
title = {Meta-Learning With Differentiable Convex Optimization},
booktitle = {Proceedings of the IEEE/CVF Conference on Computer Vision and Pattern Recognition (CVPR)},
month = {June},
year = {2019}
}

@inproceedings{sun2018neural,
  title={Neural program synthesis from diverse demonstration videos},
  author={Sun, Shao-Hua and Noh, Hyeonwoo and Somasundaram, Sriram and Lim, Joseph},
  booktitle={International Conference on Machine Learning},
  pages={4790--4799},
  year={2018},
  organization={PMLR}
}

@article{weng2018meta,
  title={Meta-learning: Learning to learn fast},
  author={Weng, Lilian},
  journal={Lil’Log https://lilianweng. github. io/lil-log/2018/11/30/meta-learning. html},
  year={2018}
}

@inproceedings{li2018learning,
  title={Learning to generalize: Meta-learning for domain generalization},
  author={Li, Da and Yang, Yongxin and Song, Yi-Zhe and Hospedales, Timothy M},
  booktitle={Thirty-Second AAAI Conference on Artificial Intelligence},
  year={2018}
}

@inproceedings{gulrajani2020search,
  title={In Search of Lost Domain Generalization},
  author={Gulrajani, Ishaan and Lopez-Paz, David},
  booktitle={International Conference on Learning Representations},
  year={2020}
}

@inproceedings{align,
  title={Scaling up visual and vision-language representation learning with noisy text supervision},
  author={Jia, Chao and Yang, Yinfei and Xia, Ye and Chen, Yi-Ting and Parekh, Zarana and Pham, Hieu and Le, Quoc and Sung, Yun-Hsuan and Li, Zhen and Duerig, Tom},
  booktitle={International Conference on Machine Learning},
  pages={4904--4916},
  year={2021},
  organization={PMLR}
}

@inproceedings{golatkar2021mixed,
  title={Mixed-privacy forgetting in deep networks},
  author={Golatkar, Aditya and Achille, Alessandro and Ravichandran, Avinash and Polito, Marzia and Soatto, Stefano},
  booktitle={Proceedings of the IEEE/CVF Conference on Computer Vision and Pattern Recognition},
  pages={792--801},
  year={2021}
}

@InProceedings{Achille_2021_CVPR,
    author    = {Achille, Alessandro and Golatkar, Aditya and Ravichandran, Avinash and Polito, Marzia and Soatto, Stefano},
    title     = {LQF: Linear Quadratic Fine-Tuning},
    booktitle = {Proceedings of the IEEE/CVF Conference on Computer Vision and Pattern Recognition (CVPR)},
    month     = {June},
    year      = {2021},
    pages     = {15729-15739}
}

@inproceedings{lit,
  title={Lit: Zero-shot transfer with locked-image text tuning},
  author={Zhai, Xiaohua and Wang, Xiao and Mustafa, Basil and Steiner, Andreas and Keysers, Daniel and Kolesnikov, Alexander and Beyer, Lucas},
  booktitle={Proceedings of the IEEE/CVF Conference on Computer Vision and Pattern Recognition},
  pages={18123--18133},
  year={2022}
}

@article{explanatory,
  title={Explanatory Learning: Beyond Empiricism in Neural Networks},
  author={Norelli, Antonio and Mariani, Giorgio and Moschella, Luca and Santilli, Andrea and Parascandolo, Giambattista and Melzi, Simone and Rodol{\`a}, Emanuele},
  journal={arXiv preprint arXiv:2201.10222},
  year={2022}
}

@article{dalle2,
  title   = {Hierarchical Text-Conditional Image Generation with CLIP Latents},
  author  = {Aditya Ramesh and Prafulla Dhariwal and Alex Nichol and Casey Chu and Mark Chen},
  year    = {2022},
  journal = {arXiv preprint arXiv: Arxiv-2204.06125}
}

@article{product-quantization,
  title={Product quantization for nearest neighbor search},
  author={Jegou, Herve and Douze, Matthijs and Schmid, Cordelia},
  journal={IEEE transactions on pattern analysis and machine intelligence},
  volume={33},
  number={1},
  pages={117--128},
  year={2010},
  publisher={IEEE}
}

@INPROCEEDINGS{inverse-indexing,
  author={Sivic and Zisserman},
  booktitle={Proceedings Ninth IEEE International Conference on Computer Vision}, 
  title={Video Google: a text retrieval approach to object matching in videos}, 
  year={2003},
  volume={},
  number={},
  pages={1470-1477 vol.2},
  doi={10.1109/ICCV.2003.1238663}}

@article{faiss,
  title={Billion-scale similarity search with {GPUs}},
  author={Johnson, Jeff and Douze, Matthijs and J{\'e}gou, Herv{\'e}},
  journal={IEEE Transactions on Big Data},
  volume={7},
  number={3},
  pages={535--547},
  year={2019},
  publisher={IEEE}
}

@inproceedings{dino,
  title={Emerging Properties in Self-Supervised Vision Transformers},
  author={Caron, Mathilde and Touvron, Hugo and Misra, Ishan and J\'egou, Herv\'e  and Mairal, Julien and Bojanowski, Piotr and Joulin, Armand},
  booktitle={Proceedings of the International Conference on Computer Vision (ICCV)},
  year={2021}
}

@inproceedings{sentencet,
  title = "Sentence-BERT: Sentence Embeddings using Siamese BERT-Networks",
  author = "Reimers, Nils and Gurevych, Iryna",
  booktitle = "Proceedings of the 2019 Conference on Empirical Methods in Natural Language Processing",
  month = "11",
  year = "2019",
  publisher = "Association for Computational Linguistics",
}

@article{ginart2019making,
  title={Making ai forget you: Data deletion in machine learning},
  author={Ginart, Antonio and Guan, Melody and Valiant, Gregory and Zou, James Y},
  journal={Advances in neural information processing systems},
  volume={32},
  year={2019}
}

@article{trauble2022discrete,
  title={Discrete Key-Value Bottleneck},
  author={Tr{\"a}uble, Frederik and Goyal, Anirudh and Rahaman, Nasim and Mozer, Michael and Kawaguchi, Kenji and Bengio, Yoshua and Sch{\"o}lkopf, Bernhard},
  journal={arXiv preprint arXiv:2207.11240},
  year={2022}
}

@article{bommasani2021opportunities,
  title={On the opportunities and risks of foundation models},
  author={Bommasani, Rishi and Hudson, Drew A and Adeli, Ehsan and Altman, Russ and Arora, Simran and von Arx, Sydney and Bernstein, Michael S and Bohg, Jeannette and Bosselut, Antoine and Brunskill, Emma and others},
  journal={arXiv preprint arXiv:2108.07258},
  year={2021}
}

@inproceedings{basu2020influence,
  title={Influence Functions in Deep Learning Are Fragile},
  author={Basu, Samyadeep and Pope, Phil and Feizi, Soheil},
  booktitle={International Conference on Learning Representations},
  year={2021}
}

@inproceedings{koh2017understanding,
  title={Understanding black-box predictions via influence functions},
  author={Koh, Pang Wei and Liang, Percy},
  booktitle={International conference on machine learning},
  pages={1885--1894},
  year={2017},
  organization={PMLR}
}

@inproceedings{guo2020certified,
  title={Certified Data Removal from Machine Learning Models},
  author={Guo, Chuan and Goldstein, Tom and Hannun, Awni and Van Der Maaten, Laurens},
  booktitle={International Conference on Machine Learning},
  pages={3832--3842},
  year={2020},
  organization={PMLR}
}

@inproceedings{golatkar2020eternal,
  title={Eternal sunshine of the spotless net: Selective forgetting in deep networks},
  author={Golatkar, Aditya and Achille, Alessandro and Soatto, Stefano},
  booktitle={Proceedings of the IEEE/CVF Conference on Computer Vision and Pattern Recognition},
  pages={9304--9312},
  year={2020}
}

@inproceedings{golatkar2020forgetting,
  title={Forgetting outside the box: Scrubbing deep networks of information accessible from input-output observations},
  author={Golatkar, Aditya and Achille, Alessandro and Soatto, Stefano},
  booktitle={European Conference on Computer Vision},
  pages={383--398},
  year={2020},
  organization={Springer}
}

@inproceedings{proto,
  author    = {Jake Snell and Kevin Swersky and Richard S. Zemel},
  editor    = {Isabelle Guyon and Ulrike von Luxburg and Samy Bengio and Hanna M. Wallach and Rob Fergus and S. V. N. Vishwanathan and Roman Garnett},
  title     = {Prototypical Networks for Few-shot Learning},
  booktitle = {Advances in Neural Information Processing Systems 30: Annual Conference on Neural Information Processing Systems 2017, December 4-9, 2017, Long Beach, CA, {USA}},
  pages     = {4077-4087},
  year      = {2017},
  bibsource = {dblp computer science bibliography, https://dblp.org}
}

@inproceedings{cc12m,
  title = {{Conceptual 12M}: Pushing Web-Scale Image-Text Pre-Training To Recognize Long-Tail Visual Concepts},
  author = {Changpinyo, Soravit and Sharma, Piyush and Ding, Nan and Soricut, Radu},
  booktitle = {CVPR},
  year = {2021},
}

@inproceedings{deng2009imagenet,
  title={Imagenet: A large-scale hierarchical image database},
  author={Deng, Jia and Dong, Wei and Socher, Richard and Li, Li-Jia and Li, Kai and Fei-Fei, Li},
  booktitle={2009 IEEE conference on computer vision and pattern recognition},
  pages={248--255},
  year={2009},
  organization={Ieee}
}

@inproceedings{goyal2022retrieval,
  title={Retrieval-augmented reinforcement learning},
  author={Goyal, Anirudh and Friesen, Abram and Banino, Andrea and Weber, Theophane and Ke, Nan Rosemary and Badia, Adria Puigdomenech and Guez, Arthur and Mirza, Mehdi and Humphreys, Peter C and Konyushova, Ksenia and others},
  booktitle={International Conference on Machine Learning},
  pages={7740--7765},
  year={2022},
  organization={PMLR}
}

@article{watson1964smooth,
  title={Smooth regression analysis},
  author={Watson, Geoffrey S},
  journal={Sankhy{\=a}: The Indian Journal of Statistics, Series A},
  pages={359--372},
  year={1964},
  publisher={JSTOR}
}

@article{nadaraya1964estimating,
  title={On estimating regression},
  author={Nadaraya, Elizbar A},
  journal={Theory of Probability \& Its Applications},
  volume={9},
  number={1},
  pages={141--142},
  year={1964},
  publisher={SIAM}
}

@inproceedings{locatello2017unified,
  title={A unified optimization view on generalized matching pursuit and frank-wolfe},
  author={Locatello, Francesco and Khanna, Rajiv and Tschannen, Michael and Jaggi, Martin},
  booktitle={Artificial Intelligence and Statistics},
  pages={860--868},
  year={2017},
  organization={PMLR}
}

@inproceedings{locatello2018matching,
  title={On matching pursuit and coordinate descent},
  author={Locatello, Francesco and Raj, Anant and Karimireddy, Sai Praneeth and R{\"a}tsch, Gunnar and Sch{\"o}lkopf, Bernhard and Stich, Sebastian and Jaggi, Martin},
  booktitle={International Conference on Machine Learning},
  pages={3198--3207},
  year={2018},
  organization={PMLR}
}

@article{mallat1993matching,
  title={Matching pursuits with time-frequency dictionaries},
  author={Mallat, St{\'e}phane G and Zhang, Zhifeng},
  journal={IEEE Transactions on signal processing},
  volume={41},
  number={12},
  pages={3397--3415},
  year={1993},
  publisher={IEEE}
}

@article{wang2012generalized,
  title={Generalized orthogonal matching pursuit},
  author={Wang, Jian and Kwon, Seokbeop and Shim, Byonghyo},
  journal={IEEE Transactions on signal processing},
  volume={60},
  number={12},
  pages={6202--6216},
  year={2012},
  publisher={IEEE}
}

@inproceedings{sun2017revisiting,
  title={Revisiting unreasonable effectiveness of data in deep learning era},
  author={Sun, Chen and Shrivastava, Abhinav and Singh, Saurabh and Gupta, Abhinav},
  booktitle={Proceedings of the IEEE international conference on computer vision},
  pages={843--852},
  year={2017}
}

@article{thomee2016yfcc100m,
  title={YFCC100M: The new data in multimedia research},
  author={Thomee, Bart and Shamma, David A and Friedland, Gerald and Elizalde, Benjamin and Ni, Karl and Poland, Douglas and Borth, Damian and Li, Li-Jia},
  journal={Communications of the ACM},
  volume={59},
  number={2},
  pages={64--73},
  year={2016},
  publisher={ACM New York, NY, USA}
}

@inproceedings{imagenet21k,
  author    = {Tal Ridnik and Emanuel Ben Baruch and Asaf Noy and Lihi Zelnik},
  editor    = {Joaquin Vanschoren and Sai{-}Kit Yeung},
  title     = {ImageNet-21K Pretraining for the Masses},
  booktitle = {Proceedings of the Neural Information Processing Systems Track on Datasets and Benchmarks 1, NeurIPS Datasets and Benchmarks 2021, December 2021, virtual},
  year      = {2021},
  bibsource = {dblp computer science bibliography, https://dblp.org}
}

@book{mitchell1997machine,
  title={Machine learning},
  author={Mitchell, Tom},
  volume={1, 9},
  year={1997},
  publisher={McGraw-hill New York},
  isbn={0070428077}
}

@article{datacentric,
  title={Unbiggen ai},
  author={Ng, Andrew},
  journal={IEEE Spectrum},
  url = {https://spectrum.ieee.org/andrew-ng-data-centric-ai},
  year={2022}
}

@inproceedings{visualtransformers,
  author    = {Alexey Dosovitskiy and Lucas Beyer and Alexander Kolesnikov and Dirk Weissenborn and Xiaohua Zhai and Thomas Unterthiner and Mostafa Dehghani and Matthias Minderer and Georg Heigold and Sylvain Gelly and Jakob Uszkoreit and Neil Houlsby},
  title     = {An Image is Worth 16x16 Words: Transformers for Image Recognition at Scale},
  booktitle = {9th International Conference on Learning Representations, {ICLR} 2021, Virtual Event, Austria, May 3-7, 2021},
  year      = {2021},
  bibsource = {dblp computer science bibliography, https://dblp.org}
}

@article{earlyfrome2013devise,
  title={Devise: A deep visual-semantic embedding model},
  author={Frome, Andrea and Corrado, Greg S and Shlens, Jon and Bengio, Samy and Dean, Jeff and Ranzato, Marc'Aurelio and Mikolov, Tomas},
  journal={Advances in neural information processing systems},
  volume={26},
  year={2013}
}

@inproceedings{earlykarpathy2015deep,
  title={Deep visual-semantic alignments for generating image descriptions},
  author={Karpathy, Andrej and Fei-Fei, Li},
  booktitle={Proceedings of the IEEE conference on computer vision and pattern recognition},
  pages={3128--3137},
  year={2015}
}

@article{coca,
  title={Coca: Contrastive captioners are image-text foundation models},
  author={Yu, Jiahui and Wang, Zirui and Vasudevan, Vijay and Yeung, Legg and Seyedhosseini, Mojtaba and Wu, Yonghui},
  journal={arXiv preprint arXiv:2205.01917},
  year={2022}
}

@inproceedings{retro,
  author    = {Sebastian Borgeaud and Arthur Mensch and Jordan Hoffmann and Trevor Cai and Eliza Rutherford and Katie Millican and George van den Driessche and Jean{-}Baptiste Lespiau and Bogdan Damoc and Aidan Clark and Diego de Las Casas and Aurelia Guy and Jacob Menick and Roman Ring and Tom Hennigan and Saffron Huang and Loren Maggiore and Chris Jones and Albin Cassirer and Andy Brock and Michela Paganini and Geoffrey Irving and Oriol Vinyals and Simon Osindero and Karen Simonyan and Jack W. Rae and Erich Elsen and Laurent Sifre},
  title     = {Improving Language Models by Retrieving from Trillions of Tokens},
  booktitle = {International Conference on Machine Learning, {ICML} 2022, 17-23 July 2022, Baltimore, Maryland, {USA}},
  series    = {Proceedings of Machine Learning Research},
  volume    = {162},
  pages     = {2206-2240},
  publisher = {{PMLR}},
  year      = {2022},
  bibsource = {dblp computer science bibliography, https://dblp.org}
}

@article{atlas,
  title   = {Few-shot Learning with Retrieval Augmented Language Models},
  author  = {Gautier Izacard and Patrick Lewis and Maria Lomeli and Lucas Hosseini and Fabio Petroni and Timo Schick and Jane Dwivedi-Yu and Armand Joulin and Sebastian Riedel and Edouard Grave},
  year    = {2022},
  journal = {arXiv preprint arXiv: Arxiv-2208.03299}
}

@inproceedings{liu2021retrieval,
  title={Retrieval-Augmented Generation for Code Summarization via Hybrid GNN},
  author={Liu, Shangqing and Chen, Yu and Xie, Xiaofei and Siow, Jing Kai and Liu, Yang},
  booktitle={International Conference on Learning Representations},
  year={2021}
}

@article{alias2021neural,
  title={Neural production systems},
  author={Anirudh Goyal and Didolkar, Aniket and Ke, Nan Rosemary and Blundell, Charles and Beaudoin, Philippe and Heess, Nicolas and Mozer, Michael C and Bengio, Yoshua},
  journal={Advances in Neural Information Processing Systems},
  volume={34},
  pages={25673--25687},
  year={2021}
}

@article{rota,
  title={The barrier of meaning},
  author={Rota, Gian-Carlo},
  journal={Letters in Mathematical Physics},
  volume={10},
  number={2},
  pages={97--99},
  year={1985},
  publisher={Springer}
}

@article{hofstadter2001analogy,
  title={Analogy as the core of cognition},
  author={Hofstadter, Douglas},
  journal={The analogical mind: Perspectives from cognitive science},
  pages={499--538},
  year={2001}
}

@article{hofmann2008kernel,
  title={Kernel methods in machine learning},
  author={Hofmann, Thomas and Sch{\"o}lkopf, Bernhard and Smola, Alexander J},
  journal={The annals of statistics},
  volume={36},
  number={3},
  pages={1171--1220},
  year={2008},
  publisher={Institute of Mathematical Statistics}
}

@article{lecun1989backpropagation,
  title={Backpropagation applied to handwritten zip code recognition},
  author={LeCun, Yann and Boser, Bernhard and Denker, John S and Henderson, Donnie and Howard, Richard E and Hubbard, Wayne and Jackel, Lawrence D},
  journal={Neural computation},
  volume={1},
  number={4},
  pages={541--551},
  year={1989},
  publisher={MIT Press}
}

@article{moschella2022relative,
  title={Relative representations enable zero-shot latent space communication},
  author={Moschella, Luca and Maiorca, Valentino and Fumero, Marco and Norelli, Antonio and Locatello, Francesco and Rodol{\`a}, Emanuele},
  journal = {11th International Conference on Learning Representations, {ICLR} 2023},
  publisher = {OpenReview.net},
  year={2022}
}

@inproceedings{deit,
  title={Training data-efficient image transformers \& distillation through attention},
  author={Touvron, Hugo and Cord, Matthieu and Douze, Matthijs and Massa, Francisco and Sablayrolles, Alexandre and J{\'e}gou, Herv{\'e}},
  booktitle={International conference on machine learning},
  pages={10347--10357},
  year={2021},
  organization={PMLR}
}

@article{helber2019eurosat,
  title     = {Eurosat: A novel dataset and deep learning benchmark for land use and land cover classification},
  author    = {Helber, Patrick and Bischke, Benjamin and Dengel, Andreas and Borth, Damian},
  journal   = {IEEE Journal of Selected Topics in Applied Earth Observations and Remote Sensing},
  volume    = {12},
  number    = {7},
  pages     = {2217-2226},
  year      = {2019},
  publisher = {IEEE}
}

@mastersthesis{cifar,
  title={Learning multiple layers of features from tiny images},
  author={Krizhevsky, Alex and Hinton, Geoffrey and others},
  year={2009},
  school={University of Toronto, ON, Canada}
}

@INPROCEEDINGS{pets,
  author={Parkhi, Omkar M and Vedaldi, Andrea and Zisserman, Andrew and Jawahar, C. V.},
  booktitle={2012 IEEE Conference on Computer Vision and Pattern Recognition}, 
  title={Cats and dogs}, 
  year={2012},
  volume={},
  number={},
  pages={3498-3505},
  doi={10.1109/CVPR.2012.6248092}}

@article{imagenetv2,
  title     = {Do ImageNet Classifiers Generalize to ImageNet?},
  author    = {B. Recht and R. Roelofs and Ludwig Schmidt and Vaishaal Shankar},
  journal   = {International Conference On Machine Learning},
  year      = {2019},
  bibSource = {Semantic Scholar https://www.semanticscholar.org/paper/4e0bb8c1c683b43357c5d5216f6b74ff2cb32434}
}

@inproceedings{wang2023understanding,
  title={Understanding Shared Speech-Text Representations},
  author={Wang, Gary and Kastner, Kyle and Bapna, Ankur and Chen, Zhehuai and Rosenberg, Andrew and Ramabhadran, Bhuvana and Zhang, Yu},
  booktitle={ICASSP 2023-2023 IEEE International Conference on Acoustics, Speech and Signal Processing (ICASSP)},
  pages={1--5},
  year={2023},
  organization={IEEE}
}

@article{gpt2020,
  title   = {Language models are few-shot learners},
  author  = {Brown, Tom and Mann, Benjamin and Ryder, Nick and Subbiah, Melanie and Kaplan, Jared D and Dhariwal, Prafulla and Neelakantan, Arvind and Shyam, Pranav and Sastry, Girish and Askell, Amanda and others},
  journal = {Advances in neural information processing systems},
  volume  = {33},
  pages   = {1877-1901},
  year    = {2020}
}

@online{claude2023,
  title={Introducing Claude},
  author={Anthropic},
  year={2023},
  url={https://www.anthropic.com/index/introducing-claude},
  note={Accessed: October 29, 2023}
}

@article{touvron2023llama,
  title   = {LLaMA: Open and Efficient Foundation Language Models},
  author  = {Hugo Touvron and Thibaut Lavril and Gautier Izacard and Xavier Martinet and Marie-Anne Lachaux and Timothée Lacroix and Baptiste Rozière and Naman Goyal and Eric Hambro and Faisal Azhar and Aurelien Rodriguez and Armand Joulin and Edouard Grave and Guillaume Lample},
  year    = {2023},
  journal = {ARXIV}
}

@article{chowdhery2022palm,
  title     = {PaLM: Scaling Language Modeling with Pathways},
  author    = {Aakanksha Chowdhery and Sharan Narang and Jacob Devlin and Maarten Bosma and Gaurav Mishra and Adam Roberts and Paul Barham and Hyung Won Chung and Charles Sutton and Sebastian Gehrmann and Parker Schuh and Kensen Shi and Sasha Tsvyashchenko and Joshua Maynez and Abhishek Rao and Parker Barnes and Yi Tay and Noam M. Shazeer and Vinodkumar Prabhakaran and Emily Reif and Nan Du and B. Hutchinson and Reiner Pope and James Bradbury and Jacob Austin and M. Isard and Guy Gur-Ari and Pengcheng Yin and Toju Duke and Anselm Levskaya and S. Ghemawat and Sunipa Dev and H. Michalewski and Xavier García and Vedant Misra and Kevin Robinson and L. Fedus and Denny Zhou and Daphne Ippolito and D. Luan and Hyeontaek Lim and Barret Zoph and A. Spiridonov and Ryan Sepassi and David Dohan and Shivani Agrawal and Mark Omernick and Andrew M. Dai and T. S. Pillai and Marie Pellat and Aitor Lewkowycz and Erica Moreira and Rewon Child and Oleksandr Polozov and Katherine Lee and Zongwei Zhou and Xuezhi Wang and Brennan Saeta and Mark Díaz and Orhan Firat and Michele Catasta and Jason Wei and K. Meier-Hellstern and D. Eck and J. Dean and Slav Petrov and Noah Fiedel},
  journal   = {Journal of machine learning research},
  year      = {2022},
  bibSource = {Semantic Scholar https://www.semanticscholar.org/paper/094ff971d6a8b8ff870946c9b3ce5aa173617bfb}
}

@article{bubeck2023sparks,
  title   = {Sparks of Artificial General Intelligence: Early experiments with GPT-4},
  author  = {Sébastien Bubeck and Varun Chandrasekaran and Ronen Eldan and Johannes Gehrke and Eric Horvitz and Ece Kamar and Peter Lee and Yin Tat Lee and Yuanzhi Li and Scott Lundberg and Harsha Nori and Hamid Palangi and Marco Tulio Ribeiro and Yi Zhang},
  year    = {2023},
  journal = {arXiv preprint arXiv: Arxiv-2303.12712}
}

@book{frankfurt2005bullshit,
  title={On bullshit},
  author={Frankfurt, Harry G},
  year={2005},
  publisher={Princeton University Press}
}

@article{openai2023gpt4,
  title   = {GPT-4 Technical Report},
  author  = {OpenAI},
  year    = {2023},
  journal = {PREPRINT}
}

@article{hallucination,
author = {Ji, Ziwei and Lee, Nayeon and Frieske, Rita and Yu, Tiezheng and Su, Dan and Xu, Yan and Ishii, Etsuko and Bang, Ye Jin and Madotto, Andrea and Fung, Pascale},
title = {Survey of Hallucination in Natural Language Generation},
year = {2023},
issue_date = {December 2023},
publisher = {Association for Computing Machinery},
address = {New York, NY, USA},
volume = {55},
number = {12},
issn = {0360-0300},
url = {https://doi.org/10.1145/3571730},
doi = {10.1145/3571730},
journal = {ACM Comput. Surv.},
month = {mar},
articleno = {248},
numpages = {38}
}

@inproceedings{parrots,
  title={On the dangers of stochastic parrots: Can language models be too big?},
  author={Bender, Emily M and Gebru, Timnit and McMillan-Major, Angelina and Shmitchell, Shmargaret},
  booktitle={Proceedings of the 2021 ACM conference on fairness, accountability, and transparency},
  pages={610--623},
  year={2021}
}

@article{bullshit,
  title={ChatGPT: Bullshit spewer or the end of traditional assessments in higher education?},
  author={Rudolph, J{\"u}rgen and Tan, Samson and Tan, Shannon},
  journal={Journal of Applied Learning and Teaching},
  volume={6},
  number={1},
  year={2023}
}

@online{heaven2022meta,
  title={Why Meta’s latest large language model survived only three days online},
  author={Heaven, Will Douglas},
  year={2022},
  url={https://www.technologyreview.com/2022/11/18/1063487/meta-large-language-model-ai-only-survived-three-days-gpt-3-science/},
  note={Accessed: October 29, 2023},
  organization={MIT Technology Review}
}

@article{kaplan2020scaling,
  title   = {Scaling Laws for Neural Language Models},
  author  = {Jared Kaplan and Sam McCandlish and Tom Henighan and Tom B. Brown and Benjamin Chess and Rewon Child and Scott Gray and Alec Radford and Jeffrey Wu and Dario Amodei},
  year    = {2020},
  journal = {arXiv preprint arXiv: Arxiv-2001.08361}
}

@online{lee2012trainingdata,
  author = {Lee, Timothy B.},
  title = {Twitter Post on LLMs Training Data},
  year = {2012},
  url = {https://twitter.com/binarybits/status/1691558467146776957},
  note = {Accessed: October 29, 2023},
}

@online{trask2023,
  author = {Trask, Andrew},
  title = {Twitter Post on non-skeptical LLMs},
  year = {2023},
  url = {https://twitter.com/iamtrask/status/1695798588641538360},
  note = {Accessed: October 29, 2023},
}

@online{hofstadter2023,
  author={Hofstadter, Douglas},
  title={Gödel, Escher, Bach, and AI},
  year={2023},
  url={https://www.theatlantic.com/ideas/archive/2023/07/godel-escher-bach-geb-ai/674589/},
  note={Accessed: October 29, 2023},
  organization={The Atlantic}
}

@online{merriamwebster2005,
  title = {Merriam-Webster's Online Dictionary, truth},
  author={Webster, Noah},
  year = {2005},
  url = {http://web.archive.org/web/20091229000000/http://www.merriam-webster.com/dictionary/truth},
  note = {Archived on 2009-12-29},
}

@article{li2023othello,
author = {Li, Kenneth},
title = {Do Large Language Models learn world models or just surface statistics?},
journal = {The Gradient},
year = {2023},
howpublished = {\url{https://thegradient.pub/othello}},
}

@article{peng2023kosmos2,
  title   = {Kosmos-2: Grounding Multimodal Large Language Models to the World},
  author  = {Zhiliang Peng and Wenhui Wang and Li Dong and Yaru Hao and Shaohan Huang and Shuming Ma and Furu Wei},
  year    = {2023},
  journal = {arXiv preprint arXiv: 2306.14824}
}

@article{alayrac2022flamingo,
  title   = {Flamingo: a visual language model for few-shot learning},
  author  = {Alayrac, Jean-Baptiste and Donahue, Jeff and Luc, Pauline and Miech, Antoine and Barr, Iain and Hasson, Yana and Lenc, Karel and Mensch, Arthur and Millican, Katherine and Reynolds, Malcolm and others},
  journal = {Advances in Neural Information Processing Systems},
  volume  = {35},
  pages   = {23716-23736},
  year    = {2022}
}

@article{lecun2023,
  title = {Twitter Post on future LLMs},
  author = {LeCun, Yann},
  year = {2023},
  url = {https://twitter.com/ylecun/status/1718263303485501784},
  note = {Accessed: October 29, 2023},
}

@article{srivastava2023beyond,
title={Beyond the Imitation Game: Quantifying and extrapolating the capabilities of language models},
author={Aarohi Srivastava and Abhinav Rastogi and Abhishek Rao and Abu Awal Md Shoeb and Abubakar Abid and Adam Fisch and Adam R. Brown and Adam Santoro and Aditya Gupta and Adri{\`a} Garriga-Alonso and Agnieszka Kluska and Aitor Lewkowycz and Akshat Agarwal and Alethea Power and Alex Ray and Alex Warstadt and Alexander W. Kocurek and Ali Safaya and Ali Tazarv and Alice Xiang and Alicia Parrish and Allen Nie and Aman Hussain and Amanda Askell and Amanda Dsouza and Ambrose Slone and Ameet Rahane and Anantharaman S. Iyer and Anders Johan Andreassen and Andrea Madotto and Andrea Santilli and Andreas Stuhlm{\"u}ller and Andrew M. Dai and Andrew La and Andrew Lampinen and Andy Zou and Angela Jiang and Angelica Chen and Anh Vuong and Animesh Gupta and Anna Gottardi and Antonio Norelli and Anu Venkatesh and Arash Gholamidavoodi and Arfa Tabassum and Arul Menezes and Arun Kirubarajan and Asher Mullokandov and Ashish Sabharwal and Austin Herrick and Avia Efrat and Aykut Erdem and Ayla Karaka{\c{s}} and B. Ryan Roberts and Bao Sheng Loe and Barret Zoph and Bart{\l}omiej Bojanowski and Batuhan {\"O}zyurt and Behnam Hedayatnia and Behnam Neyshabur and Benjamin Inden and Benno Stein and Berk Ekmekci and Bill Yuchen Lin and Blake Howald and Bryan Orinion and Cameron Diao and Cameron Dour and Catherine Stinson and Cedrick Argueta and Cesar Ferri and Chandan Singh and Charles Rathkopf and Chenlin Meng and Chitta Baral and Chiyu Wu and Chris Callison-Burch and Christopher Waites and Christian Voigt and Christopher D Manning and Christopher Potts and Cindy Ramirez and Clara E. Rivera and Clemencia Siro and Colin Raffel and Courtney Ashcraft and Cristina Garbacea and Damien Sileo and Dan Garrette and Dan Hendrycks and Dan Kilman and Dan Roth and C. Daniel Freeman and Daniel Khashabi and Daniel Levy and Daniel Mosegu{\'\i} Gonz{\'a}lez and Danielle Perszyk and Danny Hernandez and Danqi Chen and Daphne Ippolito and Dar Gilboa and David Dohan and David Drakard and David Jurgens and Debajyoti Datta and Deep Ganguli and Denis Emelin and Denis Kleyko and Deniz Yuret and Derek Chen and Derek Tam and Dieuwke Hupkes and Diganta Misra and Dilyar Buzan and Dimitri Coelho Mollo and Diyi Yang and Dong-Ho Lee and Dylan Schrader and Ekaterina Shutova and Ekin Dogus Cubuk and Elad Segal and Eleanor Hagerman and Elizabeth Barnes and Elizabeth Donoway and Ellie Pavlick and Emanuele Rodol{\`a} and Emma Lam and Eric Chu and Eric Tang and Erkut Erdem and Ernie Chang and Ethan A Chi and Ethan Dyer and Ethan Jerzak and Ethan Kim and Eunice Engefu Manyasi and Evgenii Zheltonozhskii and Fanyue Xia and Fatemeh Siar and Fernando Mart{\'\i}nez-Plumed and Francesca Happ{\'e} and Francois Chollet and Frieda Rong and Gaurav Mishra and Genta Indra Winata and Gerard de Melo and Germ{\'a}n Kruszewski and Giambattista Parascandolo and Giorgio Mariani and Gloria Xinyue Wang and Gonzalo Jaimovitch-Lopez and Gregor Betz and Guy Gur-Ari and Hana Galijasevic and Hannah Kim and Hannah Rashkin and Hannaneh Hajishirzi and Harsh Mehta and Hayden Bogar and Henry Francis Anthony Shevlin and Hinrich Schuetze and Hiromu Yakura and Hongming Zhang and Hugh Mee Wong and Ian Ng and Isaac Noble and Jaap Jumelet and Jack Geissinger and Jackson Kernion and Jacob Hilton and Jaehoon Lee and Jaime Fern{\'a}ndez Fisac and James B Simon and James Koppel and James Zheng and James Zou and Jan Kocon and Jana Thompson and Janelle Wingfield and Jared Kaplan and Jarema Radom and Jascha Sohl-Dickstein and Jason Phang and Jason Wei and Jason Yosinski and Jekaterina Novikova and Jelle Bosscher and Jennifer Marsh and Jeremy Kim and Jeroen Taal and Jesse Engel and Jesujoba Alabi and Jiacheng Xu and Jiaming Song and Jillian Tang and Joan Waweru and John Burden and John Miller and John U. Balis and Jonathan Batchelder and Jonathan Berant and J{\"o}rg Frohberg and Jos Rozen and Jose Hernandez-Orallo and Joseph Boudeman and Joseph Guerr and Joseph Jones and Joshua B. Tenenbaum and Joshua S. Rule and Joyce Chua and Kamil Kanclerz and Karen Livescu and Karl Krauth and Karthik Gopalakrishnan and Katerina Ignatyeva and Katja Markert and Kaustubh Dhole and Kevin Gimpel and Kevin Omondi and Kory Wallace Mathewson and Kristen Chiafullo and Ksenia Shkaruta and Kumar Shridhar and Kyle McDonell and Kyle Richardson and Laria Reynolds and Leo Gao and Li Zhang and Liam Dugan and Lianhui Qin and Lidia Contreras-Ochando and Louis-Philippe Morency and Luca Moschella and Lucas Lam and Lucy Noble and Ludwig Schmidt and Luheng He and Luis Oliveros-Col{\'o}n and Luke Metz and L{\"u}tfi Kerem Senel and Maarten Bosma and Maarten Sap and Maartje Ter Hoeve and Maheen Farooqi and Manaal Faruqui and Mantas Mazeika and Marco Baturan and Marco Marelli and Marco Maru and Maria Jose Ramirez-Quintana and Marie Tolkiehn and Mario Giulianelli and Martha Lewis and Martin Potthast and Matthew L Leavitt and Matthias Hagen and M{\'a}ty{\'a}s Schubert and Medina Orduna Baitemirova and Melody Arnaud and Melvin McElrath and Michael Andrew Yee and Michael Cohen and Michael Gu and Michael Ivanitskiy and Michael Starritt and Michael Strube and Micha{\l} Sw{\k{e}}drowski and Michele Bevilacqua and Michihiro Yasunaga and Mihir Kale and Mike Cain and Mimee Xu and Mirac Suzgun and Mitch Walker and Mo Tiwari and Mohit Bansal and Moin Aminnaseri and Mor Geva and Mozhdeh Gheini and Mukund Varma T and Nanyun Peng and Nathan Andrew Chi and Nayeon Lee and Neta Gur-Ari Krakover and Nicholas Cameron and Nicholas Roberts and Nick Doiron and Nicole Martinez and Nikita Nangia and Niklas Deckers and Niklas Muennighoff and Nitish Shirish Keskar and Niveditha S. Iyer and Noah Constant and Noah Fiedel and Nuan Wen and Oliver Zhang and Omar Agha and Omar Elbaghdadi and Omer Levy and Owain Evans and Pablo Antonio Moreno Casares and Parth Doshi and Pascale Fung and Paul Pu Liang and Paul Vicol and Pegah Alipoormolabashi and Peiyuan Liao and Percy Liang and Peter W Chang and Peter Eckersley and Phu Mon Htut and Pinyu Hwang and Piotr Mi{\l}kowski and Piyush Patil and Pouya Pezeshkpour and Priti Oli and Qiaozhu Mei and Qing Lyu and Qinlang Chen and Rabin Banjade and Rachel Etta Rudolph and Raefer Gabriel and Rahel Habacker and Ramon Risco and Rapha{\"e}l Milli{\`e}re and Rhythm Garg and Richard Barnes and Rif A. Saurous and Riku Arakawa and Robbe Raymaekers and Robert Frank and Rohan Sikand and Roman Novak and Roman Sitelew and Ronan Le Bras and Rosanne Liu and Rowan Jacobs and Rui Zhang and Russ Salakhutdinov and Ryan Andrew Chi and Seungjae Ryan Lee and Ryan Stovall and Ryan Teehan and Rylan Yang and Sahib Singh and Saif M. Mohammad and Sajant Anand and Sam Dillavou and Sam Shleifer and Sam Wiseman and Samuel Gruetter and Samuel R. Bowman and Samuel Stern Schoenholz and Sanghyun Han and Sanjeev Kwatra and Sarah A. Rous and Sarik Ghazarian and Sayan Ghosh and Sean Casey and Sebastian Bischoff and Sebastian Gehrmann and Sebastian Schuster and Sepideh Sadeghi and Shadi Hamdan and Sharon Zhou and Shashank Srivastava and Sherry Shi and Shikhar Singh and Shima Asaadi and Shixiang Shane Gu and Shubh Pachchigar and Shubham Toshniwal and Shyam Upadhyay and Shyamolima Shammie Debnath and Siamak Shakeri and Simon Thormeyer and Simone Melzi and Siva Reddy and Sneha Priscilla Makini and Soo-Hwan Lee and Spencer Torene and Sriharsha Hatwar and Stanislas Dehaene and Stefan Divic and Stefano Ermon and Stella Biderman and Stephanie Lin and Stephen Prasad and Steven Piantadosi and Stuart Shieber and Summer Misherghi and Svetlana Kiritchenko and Swaroop Mishra and Tal Linzen and Tal Schuster and Tao Li and Tao Yu and Tariq Ali and Tatsunori Hashimoto and Te-Lin Wu and Th{\'e}o Desbordes and Theodore Rothschild and Thomas Phan and Tianle Wang and Tiberius Nkinyili and Timo Schick and Timofei Kornev and Titus Tunduny and Tobias Gerstenberg and Trenton Chang and Trishala Neeraj and Tushar Khot and Tyler Shultz and Uri Shaham and Vedant Misra and Vera Demberg and Victoria Nyamai and Vikas Raunak and Vinay Venkatesh Ramasesh and vinay uday prabhu and Vishakh Padmakumar and Vivek Srikumar and William Fedus and William Saunders and William Zhang and Wout Vossen and Xiang Ren and Xiaoyu Tong and Xinran Zhao and Xinyi Wu and Xudong Shen and Yadollah Yaghoobzadeh and Yair Lakretz and Yangqiu Song and Yasaman Bahri and Yejin Choi and Yichi Yang and Yiding Hao and Yifu Chen and Yonatan Belinkov and Yu Hou and Yufang Hou and Yuntao Bai and Zachary Seid and Zhuoye Zhao and Zijian Wang and Zijie J. Wang and Zirui Wang and Ziyi Wu},
journal={Transactions on Machine Learning Research},
issn={2835-8856},
year={2023},
url={https://openreview.net/forum?id=uyTL5Bvosj},
note={}
}

@inproceedings{cosmo2020limp,
  title={Limp: Learning latent shape representations with metric preservation priors},
  author={Cosmo, Luca and Norelli, Antonio and Halimi, Oshri and Kimmel, Ron and Rodola, Emanuele},
  booktitle={Computer Vision--ECCV 2020},
  volume={3},
  pages={9--35},
  year={2020}
}

@article{norelli2022olivaw,
  title={Olivaw: Mastering othello without human knowledge, nor a penny},
  author={Norelli, Antonio and Panconesi, Alessandro},
  journal={IEEE Transactions on Games},
  year={2022},
  publisher={IEEE}
}

@inproceedings{lauletta2022errare,
  title={Errare humanum est? a pilot study to evaluate the human-likeness of a AI othello playing agent},
  author={Lauletta, Enrico and Biancardi, Beatrice and Norelli, Antonio and Mancini, Maurizio and Panconesi, Alessandro},
  booktitle={Proceedings of the 22nd ACM International Conference on Intelligent Virtual Agents},
  pages={1--3},
  year={2022}
}

@article{norelli2022asif,
  title={{ASIF}: Coupled Data Turns Unimodal Models to Multimodal
Without Training},
  author={Norelli, Antonio and Fumero, Marco and Maiorca, Valentino and Moschella, Luca and Rodola, Emanuele and Locatello, Francesco},
 booktitle={Advances in Neural Information Processing Systems 37: Annual Conference
    on Neural Information Processing Systems 2023, NeurIPS 2023, December
    10-16, 2023},
    year={2023},
    url={https://neurips.cc/virtual/2023/poster/71339}
    }

@article{cannistraci2023bootstrapping,
  title={Bootstrapping Parallel Anchors for Relative Representations},
  author={Cannistraci, Irene and Moschella, Luca and Maiorca, Valentino and Fumero, Marco and Norelli, Antonio and Rodol{\`a}, Emanuele},
  journal={Tiny paper at ICLR 2023},
  year={2023}
}

@article{shokry2023learning,
  title={Learning Rotation-Agnostic Representations via Group Equivariant VAEs},
  author={Shokry, Ahmedeo and Norelli, Antonio},
  journal={Tiny paper at ICLR 2023},
  year={2023}
}

@article{maiorca2023latent,
  title={Latent Space Translation via Semantic Alignment},
  author={Maiorca, Valentino and Moschella, Luca and Norelli, Antonio and Fumero, Marco and Locatello, Francesco and Rodolà, Emanuele },
 booktitle={Advances in Neural Information Processing Systems 37: Annual Conference
    on Neural Information Processing Systems 2023, NeurIPS 2023, December
    10-16, 2023},
    year={2023},
    url={https://neurips.cc/virtual/2023/poster/71339}
    }

@article{chiang2000catching,
  title={Catching crumbs from the table},
  author={Chiang, Ted},
  journal={Nature},
  volume={405},
  number={6786},
  pages={517--517},
  year={2000},
  publisher={Nature Publishing Group UK London}
}

@inproceedings{zeng2023socratic,
  title     = {Socratic Models: Composing Zero-Shot Multimodal Reasoning with Language},
  author    = {Andy Zeng and Maria Attarian and brian ichter and Krzysztof Marcin Choromanski and Adrian Wong and Stefan Welker and Federico Tombari and Aveek Purohit and Michael S Ryoo and Vikas Sindhwani and Johnny Lee and Vincent Vanhoucke and Pete Florence},
  booktitle = {The Eleventh International Conference on Learning Representations},
  year      = {2023},
  url       = {https://openreview.net/forum?id=G2Q2Mh3avow}
}

@article{mu2019shaping,
  title   = {Shaping Visual Representations with Language for Few-shot Classification},
  author  = {Jesse Mu and Percy Liang and Noah Goodman},
  year    = {2020},
  journal = {ACL}
}

@inproceedings{VinyalsBLKW16,
  author    = {Oriol Vinyals and Charles Blundell and Tim Lillicrap and Koray Kavukcuoglu and Daan Wierstra},
  editor    = {Daniel D. Lee and Masashi Sugiyama and Ulrike von Luxburg and Isabelle Guyon and Roman Garnett},
  title     = {Matching Networks for One Shot Learning},
  booktitle = {Advances in Neural Information Processing Systems 29: Annual Conference on Neural Information Processing Systems 2016, December 5-10, 2016, Barcelona, Spain},
  pages     = {3630-3638},
  year      = {2016},
  url       = {https://proceedings.neurips.cc/paper/2016/hash/90e1357833654983612fb05e3ec9148c-Abstract.html},
  bibsource = {dblp computer science bibliography, https://dblp.org}
}

@inproceedings{Santoro16Meta,
author = {Santoro, Adam and Bartunov, Sergey and Botvinick, Matthew and Wierstra, Daan and Lillicrap, Timothy},
title = {Meta-learning with memory-augmented neural networks},
year = {2016},
publisher = {JMLR.org},
booktitle = {Proceedings of the 33rd International Conference on International Conference on Machine Learning - Volume 48},
pages = {1842–1850},
numpages = {9},
location = {New York, NY, USA},
series = {ICML'16}
}

@article{wang2016learning,
  title   = {Learning to reinforcement learn},
  author  = {Jane X Wang and Zeb Kurth-Nelson and Dhruva Tirumala and Hubert Soyer and Joel Z Leibo and Remi Munos and Charles Blundell and Dharshan Kumaran and Matt Botvinick},
  year    = {2016},
  journal = {arXiv preprint arXiv: 1611.05763}
}

@book{bongard1967pattern,
  author       = {Bongard, M. M.},
  title        = {Pattern Recognition},
  year         = {1967},
  publisher    = {Hayden Book Co., Spartan Books},
  address      = {Rochelle Park, N.J.},
  note         = {Original publication: \textit{Recognition Problem}, Nauka Press, Moscow, 1967}
}

@article{oitzman2021openai,
  title={OpenAI abandons robotics research},
  author={Oitzman, Mike},
  journal={The Robot Report},
  year={2021},
  url={https://www.therobotreport.com/openai-abandons-robotics-research/},
  note={Accessed: 2023-10-24}
}

@article{google2016ai,
  title={How Google Plans to Solve Artificial Intelligence},
  author={Simonite, Tom},
  journal={MIT Technology Review},
  year={2016},
  url={https://www.technologyreview.com/2016/03/31/161234/how-google-plans-to-solve-artificial-intelligence/},
  note={Accessed: 2023-10-24}
}

@article{kaur2021ask,
  title   = {Ask \& Explore: Grounded Question Answering for Curiosity-Driven Exploration},
  author  = {Jivat Neet Kaur and Yiding Jiang and Paul Pu Liang},
  year    = {2021},
  journal = {arXiv preprint arXiv: 2104.11902}
}

@article{pathak2017curiositydriven,
  title   = {Curiosity-driven Exploration by Self-supervised Prediction},
  author  = {Deepak Pathak and Pulkit Agrawal and Alexei A. Efros and Trevor Darrell},
  year    = {2017},
  journal = {ICML}
}
